\documentclass[11pt, a4paper]{lumia}

\usepackage[sort&compress]{natbib}
\usepackage{xspace}

\theoremstyle{plain}
\newtheorem*{proposition*}{Proposition}

\theoremstyle{definition}

\theoremstyle{definition}

\def\eqref#1{equation~\ref{#1}}
\usepackage{graphicx}           % graphics support
\usepackage{tikz}               % TikZ graphics
\usetikzlibrary{arrows.meta,positioning,fit,calc}
\usepackage{pgfplots}           % publication-quality sensitivity plots
\usepgfplotslibrary{groupplots}
\pgfplotsset{compat=1.18}
\usepackage[edges]{forest}      % forest trees

\usepackage{url}                % simple URL typesetting
\usepackage{xurl}               % extended URL breaking

\usepackage{array}              % extended array and tabular
\usepackage{longtable}          % multi-page tables
\usepackage{multirow}           % multirow cells
\usepackage{makecell}           % enhanced cell formatting
\usepackage{ragged2e}           % ragged text alignment

\usepackage{mathtools}          % additional math tools
\usepackage{nicefrac}           % compact symbols for 1/2, etc.

\usepackage{algorithm}          % algorithm environment
\usepackage{algorithmicx}       % algorithmic pseudocode
\usepackage{algpseudocode}      % pseudocode formatting
\usepackage{listings}           % code listings

\usepackage{subcaption}         % subfigures and subcaptions
\usepackage{wrapfig}            % text wrapping around figures
\usepackage[export]{adjustbox}  % adjustable boxes

\usepackage{changes}            % track changes
\usepackage{xspace}             % intelligent spacing
\usepackage[normalem]{ulem}     % underlining without affecting emphasis
\usepackage{CJKutf8}            % CJK character support

\usepackage[tikz]{bclogo}       % logo boxes
\usepackage[framemethod=tikz]{mdframed} % framed text

\usepackage{lipsum}             % lorem ipsum text
\usepackage{tocloft}            % table of contents formatting
\usepackage{afterpage}          % page break control
\usepackage{bbding}             % additional symbols
\usepackage{epigraph}           % epigraphs
\usepackage{minitoc}            % mini table of contents
\usepackage{multicol}           % multiple columns
\usepackage{textgreek}          % Greek text

\newcolumntype{P}[1]{>{\RaggedRight\arraybackslash}p{#1}}

\definecolor{uclablue}{RGB}{39, 116, 174}
\definecolor{bigaired}{RGB}{156, 0, 0}
\definecolor{myblue}{HTML}{598BE7}
\definecolor{mildblue}{RGB}{31,119,180}
\definecolor{sectionblue}{RGB}{70, 130, 180}
\definecolor{methodblue}{RGB}{0, 150, 136}
\definecolor{bgblue}{RGB}{245,243,253}
\definecolor{ttblue}{RGB}{91,194,224}
\definecolor{mygreen}{rgb}{0.64, 0.56, 0.88}
\definecolor{myyellow}{rgb}{0.68, 0.6, 0.1}
\definecolor{fancygreen}{rgb}{0.33, 0.68, 0.20}
\definecolor{salmon}{rgb}{0.94, 0.52, 0.49}
\definecolor{tablegreen}{rgb}{0.82, 0.94, 0.75}
\definecolor{tableblue}{rgb}{0.81, 0.90, 0.94}
\definecolor{tablered}{rgb}{0.97, 0.85, 0.85}
\definecolor{tableorange}{rgb}{0.96, 0.85, 0.81}
\definecolor{myorange}{rgb}{1.0, 0.49, 0.0}
\definecolor{tlgreen}{rgb}{0.33, 0.68, 0.20}
\definecolor{darkgreen}{RGB}{0,100,0}
\definecolor{darkred}{RGB}{200, 0, 0}
\definecolor{customyellow}{HTML}{FFFACD}
\definecolor{refinegreen}{RGB}{0, 128, 75}
\definecolor{scoregreen}{RGB}{34, 139, 34}
\definecolor{hidden-blue}{RGB}{194,232,247}
\definecolor{hidden-black}{RGB}{20,68,106}
\definecolor{yes}{HTML}{C6EFCE}
\definecolor{no}{HTML}{FFC7CE}
\definecolor{partial}{HTML}{FFEB9C}
\definecolor{external}{HTML}{D9E1F2}
\definecolor{hdr}{HTML}{F2F2F2}
\definecolor{GRPOrow}{gray}{0.96}
\definecolor{FlowRLrow}{RGB}{225,236,255}
\definecolor{FlowBlue}{RGB}{80,120,210}
\definecolor{GRPOGray}{gray}{0.35}

\hypersetup{
    colorlinks=true,
    pdfpagelayout=OneColumn,
    citecolor=uclablue,
    linkcolor=bigaired,
    urlcolor=darkblue
}

\setlist[itemize]{leftmargin=20pt, noitemsep, topsep=0pt}

\NewDocumentCommand{\kaiyan}{mO{}}{\textcolor{purple}{\textsuperscript{\textit{kaiyan}}\textsf{\textbf{\small[#1]}}}}
\NewDocumentCommand{\yuxin}{mO{}}{\textcolor{cyan}{\textsuperscript{\textit{yuxin}}\textsf{\textbf{\small[#1]}}}}
\NewDocumentCommand{\bx}{mO{}}{\textcolor{green}{\textsuperscript{\textit{bx}}\textsf{\textbf{\small[#1]}}}}
\NewDocumentCommand{\at}{mO{}}{\textcolor{red}{\textsuperscript{\textit{AT}}\textsf{\textbf{\small[#1]}}}}
\NewDocumentCommand{\re}{mO{}}{\textcolor{blue}{\textsuperscript{\textit{RE}}\textsf{\textbf{\small[#1]}}}}
\NewDocumentCommand{\ybsun}{mO{}}{\textcolor{magenta}{\textsuperscript{\textit{youbang}}\textsf{\textbf{\small[#1]}}}}
\NewDocumentCommand{\runze}{mO{}}{\textcolor{orange}{\textsuperscript{\textit{runze}}\textsf{\textbf{\small[#1]}}}}
\NewDocumentCommand{\add}{mO{}}{\textcolor{darkgreen}{\textsuperscript{\textit{Maybe Consider Discuss}}\textsf{\textbf{[#1]}}}}

\newcommand{\cmark}{\textcolor{darkgreen}{\boldmath$\checkmark$}}
\newcommand{\xmark}{\textcolor{darkred}{\boldmath$\times$}}

\newenvironment{itemize*}%
 {\leftmargini=10pt\begin{itemize}%
  \setlength{\itemsep}{0pt}%
  \setlength{\parskip}{0pt}%
  }%
 {\end{itemize}}

\newenvironment{enumerate*}%
 {\begin{enumerate}%
  \setlength{\itemsep}{0pt}%
  \setlength{\parskip}{0pt}}%
 {\end{enumerate}}

\newcommand{\cellstatus}[1]{%
  \begingroup
  \StrTrim{#1}[\statusval]%
  \IfStrEq{\statusval}{Yes}{\cellcolor{yes}\cmark}{}%
  \IfStrEq{\statusval}{No}{\cellcolor{no}\xmark}{}%
  \IfBeginWith{\statusval}{Yes (}{\cellcolor{yes}\cmark~\textit{\statusval\unskip}}{}%
  \IfStrEq{\statusval}{Partial}{\cellcolor{partial}\textbf{Partial}}{}%
  \IfStrEq{\statusval}{External}{\cellcolor{external}\textbf{External}}{}%
  \endgroup
}

\newtcolorbox{myboxi}[1][]{
  breakable,
  title=#1,
  colback=red!5,
  colbacktitle=red!5,
  coltitle=black,
  fonttitle=\bfseries,
  bottomrule=0pt,
  toprule=0pt,
  leftrule=2pt,
  rightrule=2pt,
  titlerule=0pt,
  arc=0pt,
  outer arc=0pt,
  colframe=red,
}

\newtcolorbox{myboxnote}[1][]{
  breakable,
  title=#1,
  colback=orange!0,
  colbacktitle=orange!0,
  coltitle=black,
  fonttitle=\bfseries,
  bottomrule=0pt,
  toprule=0pt,
  leftrule=2pt,
  rightrule=2pt,
  titlerule=0pt,
  arc=0pt,
  outer arc=0pt,
  colframe=orange,
}

\newtcolorbox{myboxii}[1][]{
  breakable,
  freelance,
  title=#1,
  colback=white,
  colbacktitle=white,
  coltitle=black,
  fonttitle=\bfseries,
  bottomrule=0pt,
  boxrule=0pt,
  colframe=white,
  overlay unbroken and first={
  \draw[red!75!black,line width=3pt]
    ([xshift=5pt]frame.north west) --
    (frame.north west) --
    (frame.south west);
  \draw[red!75!black,line width=3pt]
    ([xshift=-5pt]frame.north east) --
    (frame.north east) --
    (frame.south east);
  },
  overlay unbroken app={
  \draw[red!75!black,line width=3pt,line cap=rect]
    (frame.south west) --
    ([xshift=5pt]frame.south west);
  \draw[red!75!black,line width=3pt,line cap=rect]
    (frame.south east) --
    ([xshift=-5pt]frame.south east);
  },
  overlay middle and last={
  \draw[red!75!black,line width=3pt]
    (frame.north west) --
    (frame.south west);
  \draw[red!75!black,line width=3pt]
    (frame.north east) --
    (frame.south east);
  },
  overlay last app={
  \draw[red!75!black,line width=3pt,line cap=rect]
    (frame.south west) --
    ([xshift=5pt]frame.south west);
  \draw[red!75!black,line width=3pt,line cap=rect]
    (frame.south east) --
    ([xshift=-5pt]frame.south east);
  },
}

\tcbset{
  takeawaysbox/.style={
    title=Takeaways,
    colback=lightblue!80,
    colframe=black,
    fonttitle=\bfseries\small,
    coltitle=white,
    colbacktitle=black,
    enhanced,
    attach boxed title to top left={xshift=2.5mm,yshift=-2.5mm},
    boxed title style={rounded corners, size=small, colframe=black, colback=black},
    width=\linewidth,
    arc=3.5mm
  }
}

\mdfdefinestyle{mystyle}{%
  rightline=true,
  innerleftmargin=10,
  innerrightmargin=10,
  outerlinewidth=3pt,
  topline=false,
  rightline=true,
  bottomline=false,
  skipabove=\topsep,
  skipbelow=\topsep
}

\tikzset{%
    every node/.style={font=\tiny},
    parent/.style =          {align=center,text width=2cm,rounded corners=3pt, line width=0.3mm, fill=gray!10,draw=gray!80},
    child/.style =           {align=center,text width=2.0cm,rounded corners=3pt, fill=blue!10,draw=blue!80,line width=0.3mm},
    grandchild/.style =      {align=center,text width=2cm,rounded corners=3pt},
    greatgrandchild/.style = {align=center,text width=1.5cm,rounded corners=3pt},
    greatgrandchild2/.style = {align=center,text width=1.5cm,rounded corners=3pt},
    referenceblock/.style =  {align=center,text width=1.5cm,rounded corners=2pt},
    pretrain/.style =           {align=center,text width=2.0cm,rounded corners=3pt, fill=blue!10,draw=blue!80,line width=0.3mm},
    pretrain_work/.style =           {align=center, text width=8.5cm,rounded corners=3pt, fill=blue!10,draw=blue!0,line width=0.3mm},
    template/.style =           {align=center,text width=2.0cm,rounded corners=3pt, fill=red!10,draw=red!80,line width=0.3mm},
    template_work/.style =           {align=center,text width=8.5cm,rounded corners=3pt, fill=red!10,draw=red!0,line width=0.3mm},
    answer/.style =           {align=center,text width=2.0cm,rounded corners=3pt, fill= cyan!10,draw= cyan!80,line width=0.3mm},
    answer_work/.style =           {align=center,text width=8.5cm,rounded corners=3pt, fill= cyan!10,draw= cyan!0,line width=0.3mm},
    multiple/.style =           {align=center,text width=2.0cm,rounded corners=3pt, fill= orange!10,draw= orange!80,line width=0.3mm},
    multiple_work/.style =           {align=center,text width=8.5cm,rounded corners=3pt, fill= orange!10,draw= orange!0,line width=0.3mm},
    tuning/.style =           {align=center,text width=2.0cm,rounded corners=3pt, fill= magenta!10,draw= magenta!80,line width=0.3mm},
    tuning_work/.style =           {align=center,text width=8.5cm,rounded corners=3pt, fill= magenta!10,draw= magenta!0,line width=0.3mm},
}

\newcommand{\lstbg}[3][0pt]{{\fboxsep#1\colorbox{#2}{\strut #3}}}

\lstdefinelanguage{diff}{
  basicstyle=\ttfamily\small,
  morecomment=[f][\lstbg{red!20}]-,
  morecomment=[f][\lstbg{green!20}]+,
}

\lstdefinelanguage{diffpython}{
  language=diff,
  morekeywords={def, if, else, for, while, return, import, from, as, class, with, try, except, finally, raise, lambda, and, or, not, in, is, None, True, False},
  morecomment=[l]{\#},
  morestring=[b]",
  morestring=[b]',
}

\setheadertext{}

\correspondingemail{\emailicon\ weirubinn@gmail.com, lin.zhouhan@gmail.com \quad $^\ddagger$ Corresponding Author.}

\githublink{https://github.com/LUMIA-Group/MemoryDecoder-at-Scale}
\githubtext{LUMIA-Group/MemoryDecoder-at-Scale}

\huggingfacelink{https://huggingface.co/collections/Rubin-Wei/memorydecoder-at-scale}
\huggingfacetext{Rubin-wei/MemoryDecoder-at-Scale}

\projectpagelink{https://rubin-wei.github.io/memory-decoder-at-scale/}
\projectpagetext{MemoryDecoder-at-Scale}

\title{Memory Decoder at Scale:\\
A Pretrained, Parametric Long-Term Memory}
\setheadertitle{Memory Decoder at Scale: A Pretrained, Parametric Long-Term Memory}

\author{%
  Rubin Wei$^{1,2}$, Jiaqi Cao$^{1}$, Jiarui Wang$^{1,2}$, Junming Zhang$^{1}$, Qipeng Guo$^{2}$, Bowen Zhou$^{2,3}$, Zhouhan Lin$^{1,2\ddagger}$\\
  $^1$ LUMIA Lab, School of Artificial Intelligence, Shanghai Jiao Tong University\\
  $^2$ Shanghai Artificial Intelligence Laboratory\\
  $^3$ Electronic Engineering, Tsinghua University
}

\begin{document}

% ====================
% ABSTRACT
% ====================
\begin{abstract}
Decoder-only language models entangle long-term memory and reasoning in a single parameter set, making it difficult to scale memory capacity independently.
Memory Decoder \citep{cao2026memory} introduces a parametric long-term memory module but only studies it at a relatively small scale.
In this work, we present \textbf{\emph{Memory Decoder at Scale}}, scaling memory models up to 6.9B parameters and pretraining them on 300B tokens.
At this data scale, the combined cost of indexing and search makes a standard Faiss pipeline infeasible.
% We address this bottleneck with a distributed Faiss pipeline using embedding compression, index sharding, and parallel search, together with sparse storage and distributed batch-wise loading of \(k\)NN distributions for memory pretraining.
We address this bottleneck with a distributed pipeline for Faiss indexing and retrieval, together with sparse, batch-wise loading of \(k\)NN distributions.
% Experiments across base model and memory scales reveal a consistent advantage for pairing smaller base models with larger memory models. This asymmetric parameter allocation can be more efficient than scaling the base model alone.
Across model scales, we find that allocating more parameters to memory yields a better parameter-performance tradeoff than scaling the base model alone.
On 17 benchmarks, pairing a 6.9B general memory with Pythia-410M raises its average score from 29.86 to 37.34, surpassing Pythia-12B (37.24) with 39\% fewer total parameters.
For Qwen3 Base models ranging from 0.6B to 14B, 1.7B domain memories improve the average score across the three domains by more than 9 points at every scale.
% Overall, our results point to independently scalable memory as a promising architectural direction for disentangling long-term memory and reasoning in language models.
Overall, our results demonstrate that independently scaling pretrained memory offers a more parameter efficient path to improving language model performance.
\end{abstract}

% Generate title with LUMIA style formatting
\maketitle

\begingroup
\setlength{\intextsep}{4pt}
\setlength{\abovecaptionskip}{5pt}
\begin{figure}[H]
\centering
\includegraphics[width=0.95\textwidth]{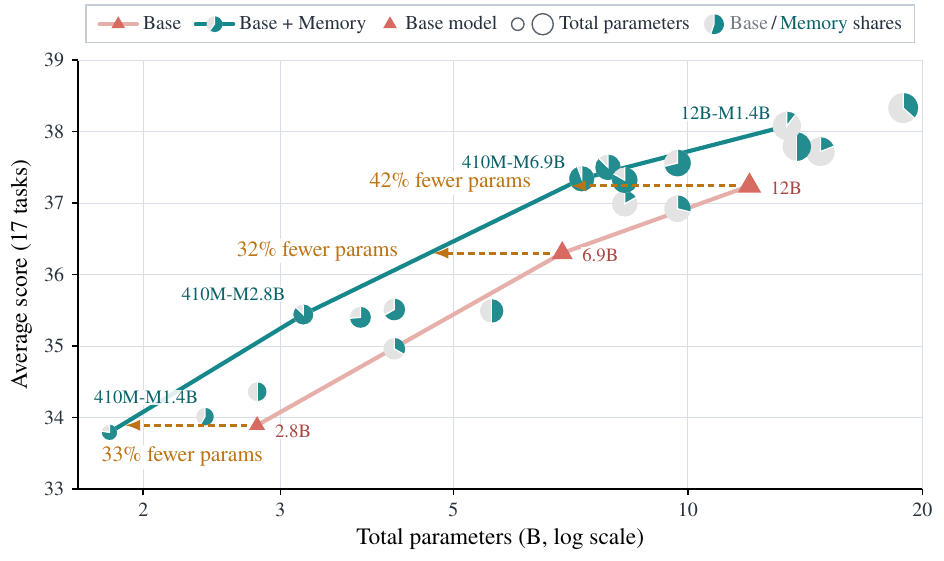}
\caption{Memory scaling is more parameter-efficient than backbone scaling. Across 17 tasks,
Pythia-410M + Mem-6.9B reaches \(37.34\), surpassing Pythia-12B at \(37.24\)
with \(39\%\) fewer total parameters. Across scales, pairing larger memories
with smaller backbones outperforms backbone-only scaling.}
\label{fig:general-memory-avg-selected-path-scatter-log}
\end{figure}
\endgroup

% ====================
% PAPER CONTENT
% ====================
\section{Introduction}
\label{sec:introduction}

\begin{figure}[t]
    \centering
    \includegraphics[width=\linewidth]{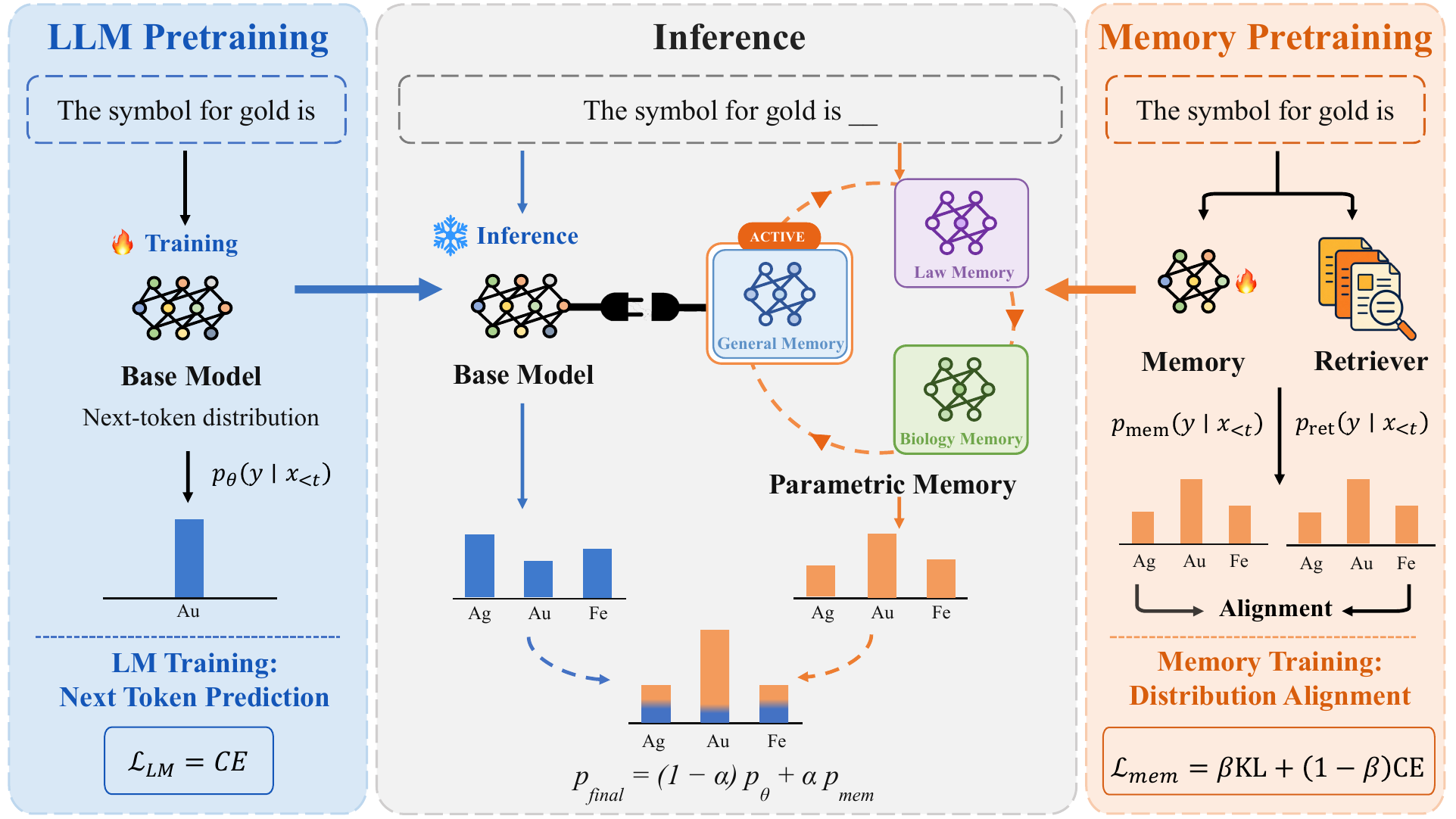}
    \caption{Overview of parametric long-term memory pretraining.
    \textbf{Left}, a standard language model is pretrained with next-token
    prediction and then frozen for inference. \textbf{Right}, a parametric
    memory is pretrained to align its output distribution with a retriever
    while retaining a next-token prediction objective. \textbf{Middle}, the
    frozen base model and a swappable memory process the same context in
    parallel, and their next-token distributions are interpolated to produce
    the final prediction.}
    \label{fig:overview}
\end{figure}

% The human brain is organized into specialized functional networks that
% interact to support cognition
% \citep{fair2009functional}. Likewise, human memory is not
% a unitary faculty but is composed of multiple systems with different
% operating principles and neuroanatomy
% \citep{squire2009memory}. Short-term memory maintains information over brief
% intervals, whereas long-term memory retains information over extended periods.
% Neuropsychological evidence indicates that the two can be dissociated
% \citep{baddeley1970amnesia}. Complementary learning
% systems theory further distinguishes a hippocampal system that rapidly learns
% the specifics of individual experiences from a neocortical system that
% gradually acquires structured knowledge representations
% \citep{mcclelland1995there}. Consistent with this functional
% specialization, studies of London taxi drivers and trainees show that
% acquiring an internal spatial representation of London is associated with
% regionally specific structural changes in the posterior hippocampus
% \citep{maguire2000navigation,woollett2011acquiring}. Human memory is therefore
% organized into systems that differ in their timescales and functional roles.

The human brain is organized into specialized functional systems that interact to support cognition \citep{fair2009functional}. Memory and reasoning likewise rely on partially distinct neural systems, allowing memory storage and cognitive computation to be functionally dissociated \citep{squire2009memory,baddeley1970amnesia}. In contrast, standard decoder-only language models
\citep{xu2026deepseek,team2026qwen3,zeng2026glm,singh2025openai} entangle
long-term memory and reasoning within a single set of parameters. Long-term
memory cannot be pretrained
or scaled independently, and increasing memory size requires a
corresponding increase in the total parameter count of the model. Under this
shared parameterization, domain adaptation through continued pretraining or
full fine-tuning requires optimization of the entire parameter set, incurs
substantial training cost, and risks catastrophic
forgetting~\citep{kirkpatrick2017overcoming} or other unintended degradation
of previously acquired capabilities. Decoder-only language models also lack a
standalone memory that can be swapped for different domains or reused across
models. These limitations motivate a modular architecture in which long-term
memory is pretrained and scaled independently of the base model, allowing
domain memories to be swapped at inference and reused across models while the
base model remains frozen.

\Needspace{3\baselineskip}
Existing work on memory in language models has not fully addressed this
entanglement. One line of work focuses on short-term memory within the input
context. Segment-level recurrence
\citep{dai2019transformer}, bounded KV caches
\citep{xiao2024efficient}, retrieval from distant context
\citep{xiao2024infllm}, and RoPE rescaling
\citep{ding2024longrope} improve access to contextual
information during inference, but do not address the entanglement between long-term memory and
reasoning. Another line of work focuses on long-term memory. RAG
\citep{lewis2020retrieval} retrieves passages from an external corpus as
additional context, whereas $k$NN language models
\citep{khandelwal2019generalization} interpolate model predictions with a
distribution constructed from nearest neighbors in an external datastore. Both
require external retrieval during inference, with additional context processing
for RAG and substantial storage and nearest-neighbor search costs for $k$NN
language models. Titans
\citep{behrouz2024titans} instead uses a neural long-term memory that encodes
information from historical context at test time rather than knowledge
acquired from a pretraining corpus. Memory Decoder \citep{cao2026memory} and
MLP Memory \citep{wei2025mlp} pretrain parametric memory modules to imitate the
output distributions of non-parametric $k$NN retrievers. 
In particular, Memory Decoder \citep{cao2026memory} demonstrates the effectiveness of pretrained parametric memory with models scaling up to 1B parameters, using WikiText-103 and domain-specific corpora containing only millions of tokens. However, its scaling behavior on substantially larger models and diverse pretraining corpora remains unexplored. Open questions remain as to whether memory pretraining continues to scale with model and data size, whether a general-purpose memory can acquire broad knowledge, and how the parameter budget should be allocated between the base model and memory.

In this work, we present \textbf{\emph{Memory Decoder at Scale}}, scaling parametric
memory pretraining to language model pretraining scale. Specifically, we
pretrain general memories with up to 6.9B parameter on 300B tokens. At this
data scale, constructing \(k\)NN distributions over the deduplicated Pile, which
contains 207B tokens, is constrained by the combined cost of indexing and
search, making a standard Faiss pipeline infeasible. We therefore develop a
distributed Faiss pipeline that addresses these bottlenecks through embedding
compression, index sharding, and parallel search. General memory pretraining
additionally uses sparse \(k\)NN distribution storage to reduce space
requirements and distributed streaming to read only the entries required by
each batch. Experiments across base model and memory scales
show a clear advantage when smaller base models are paired with larger
memories. Scaling memory can therefore be more parameter-efficient than scaling
the base model alone. We evaluate both general and domain memory. For general
memory, pairing the frozen Pythia-410M backbone with a 6.9B memory raises the
average score across 17 benchmarks from 29.86 to 37.34. This configuration
surpasses the 37.24 score of a frozen
Pythia-12B backbone while using 39\% fewer total parameters. Domain memories
with 1.7B parameters improve the average over BioInst, LawBench, and FinEval by
more than 9 points across Qwen3 backbones from 0.6B to 14B. On
Qwen3-14B-Base, they improve the three domains by 17.96, 8.97, and 3.04 points,
respectively. After transfer across vocabularies using only 20\% of the
standard memory training budget, these memories improve the domain averages of
OLMo-2-7B and OLMo-3-7B by 4.26 and 7.77 points, respectively. Further analyses
show that memory remains effective under few-shot prompting, benefits
knowledge-intensive tasks most, and improves with memory size and training budget.

\begingroup
\setlength{\parindent}{0pt}

\section{Preliminary: Memory Decoder}
\label{sec:method-rewrite-preliminaries}

Memory Decoder \citep{cao2026memory} is a standalone parametric memory trained
to imitate the behavior of a non-parametric retriever. For a sequence
\(x=(x_1,\ldots,x_T)\) over vocabulary \(\mathcal{V}\), each position \(t\)
has context \(c_t=x_{<t}\) and observed next token \(x_t\). A frozen base
language model \(M_\theta\) encodes this context as a query. Each datastore key
is a corpus context representation from \(M_\theta\), and its value is the
observed next token. The retriever searches the datastore with the query and
forms the \(k\)NN distribution
\(p_{\mathrm{ret}}(\cdot\mid c_t)\) from the values of the \(K\) nearest keys
and their normalized distance weights.

Unlike language modeling with a single observed target \(x_t\),
\(p_{\mathrm{ret}}\) offers richer supervision by capturing the diversity of
plausible continuations. A transformer decoder \(M_\psi\) predicts
\(p_\psi(\cdot\mid c_t)\) directly from the original context without taking
retrieved text as input. \(M_\psi\) and \(M_\theta\) use the same tokenizer and
output vocabulary. Retrieval is used only to construct offline supervision
for memory pretraining.

Memory Decoder combines distribution alignment with the corpus language
modeling objective \citep{bengio2003neural}. The loss for each position is
\begin{equation}
    \mathcal{L}_{\mathrm{mem}}
    =
    \beta
    D_{\mathrm{KL}}\!\left(
        p_{\mathrm{ret}}(\cdot\mid c_t)
        \,\Vert\,
        p_\psi(\cdot\mid c_t)
    \right)
    +(1-\beta)
    \bigl[-\log p_\psi(x_t\mid c_t)\bigr],
    \label{eq:method-rewrite-memory-loss}
\end{equation}
where \(D_{\mathrm{KL}}\) denotes the KL divergence
\citep{van2014renyi} and \(\beta\in[0,1]\) controls the contribution of
retrieval supervision. The KL term trains \(M_\psi\) to imitate the \(k\)NN
distribution. The language modeling term prevents excessive deviation from
the underlying corpus distribution.

At inference, \(M_\theta\) and \(M_\psi\) process the same context in parallel.
Let \(p_\theta(\cdot\mid c_t)\) and \(p_\psi(\cdot\mid c_t)\) denote their
output distributions. The predictions are combined as
\begin{equation}
    p_{\mathrm{final}}(y\mid c_t)
    =
    (1-\alpha)p_\theta(y\mid c_t)
    +\alpha p_\psi(y\mid c_t),
    \qquad y\in\mathcal{V},
    \label{eq:memory_interpolation}
\end{equation}
where \(\alpha\in[0,1]\) controls the memory contribution. Through this
interpolation, the final prediction incorporates the retrieval behavior
learned by \(M_\psi\) during memory pretraining.

\section{Memory Decoder at Scale}
\label{sec:method-rewrite-at-scale}

Constructing \(k\)NN distributions always requires an index over contextual
token representations and one search per training context. At \(207\)B
entries, storage and search costs make a standard Faiss pipeline
infeasible.
\hyperref[sec:method-rewrite-data-construction]{Section~\ref*{sec:method-rewrite-data-construction}}
develops a distributed pipeline for \(k\)NN distribution construction at this
scale.
\hyperref[sec:method-rewrite-general-memory]{Section~\ref*{sec:method-rewrite-general-memory}}
supports general memory pretraining through sparse distribution storage and
distributed loading.
\hyperref[sec:method-rewrite-domain-memory]{Section~\ref*{sec:method-rewrite-domain-memory}}
extends the same procedure to domain corpora up to \(4.4\)B tokens.

\subsection{Large-Scale \texorpdfstring{\(k\)NN}{kNN} Distribution Construction}
\label{sec:method-rewrite-data-construction}

\begin{figure}[t]
    \centering
    \includegraphics[width=\linewidth]{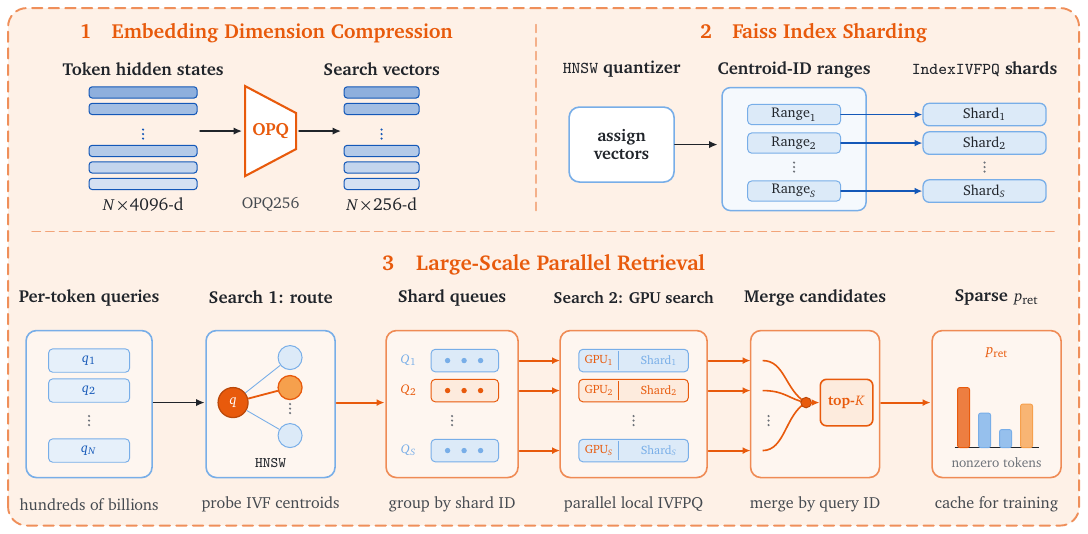}
\caption{Distributed Faiss pipeline for constructing \(k\)NN distributions
from \(207\)B contextual token representations.
    (1) OPQ compresses \(4096\)-dimensional hidden states into
    \(256\)-dimensional search vectors. (2) An IVF index with an HNSW quantizer
    partitions them into \texttt{IndexIVFPQ} shards over contiguous centroid
    ranges. (3) HNSW routes queries to parallel GPU shard search, and writer
    workers merge candidates into top \(K\) neighbors and cache
    \(p_{\mathrm{ret}}\) in sparse form for memory pretraining.}
    \label{fig:method-rewrite-pretraining-scale-construction}
\end{figure}

To construct \(k\)NN distributions at pretraining scale, let
\(\mathcal{P}_{\mathrm{mem}}\) denote the set of context-target pairs
extracted from the training corpus \(\mathcal{D}_{\mathrm{mem}}\). Each
position \(t\) defines a context \(c_t=x_{<t}\) and a target token \(x_t\).
We derive retrieval keys from the frozen base language model \(M_\theta\).
For each context \(c_t\), \(\phi_\theta(c_t)\) denotes the hidden state from
the final layer of \(M_\theta\) at the final position of \(c_t\). We set
\(k_t=\phi_\theta(c_t)\), yielding the key-value datastore
\begin{equation}
    \mathcal{R}_{\mathrm{mem}}
    =
    \{(k_t,x_t)\mid
    k_t=\phi_\theta(c_t),\
    (c_t,x_t)\in\mathcal{P}_{\mathrm{mem}}\}.
    \label{eq:method-rewrite-memory-datastore}
\end{equation}
For a query context \(c_t=x_{<t}\), let \(q_t=\phi_\theta(c_t)\). We retrieve
its \(K\) nearest key-value pairs
\(\mathcal{N}_K(q_t)\subset\mathcal{R}_{\mathrm{mem}}\) under distance \(d\),
discarding the top-ranked match if its key equals \(q_t\) to avoid trivial
self-retrieval. Each retrieved pair \((k_i,x_i)\) is assigned a normalized
weight
\begin{equation}
    \lambda_i(q_t)
    =
    \frac{\exp\!\left(-d(q_t,k_i)/\tau\right)}
    {\sum_{(k_j,x_j)\in\mathcal{N}_K(q_t)}
    \exp\!\left(-d(q_t,k_j)/\tau\right)},
    \qquad
    (k_i,x_i)\in\mathcal{N}_K(q_t),
    \label{eq:method-rewrite-retrieval-weight}
\end{equation}
where \(\tau\) is a temperature. The \(k\)NN distribution is a weighted vote
over the retrieved next tokens
\begin{equation}
    p_{\mathrm{ret}}(y\mid c_t)
    =
    \sum_{(k_i,x_i)\in\mathcal{N}_K(q_t)}
    \lambda_i(q_t)\mathbf{1}[x_i=y],
    \qquad y\in\mathcal{V}.
    \label{eq:method-rewrite-retrieval-distribution}
\end{equation}
The distribution is sparse over \(\mathcal{V}\), with neighbors that share
the same target token adding their weights to its probability.
This sparse distribution is distilled into \(M_\psi\) as the non-parametric
memory signal. At small scale, \(p_{\mathrm{ret}}\) can be constructed with a
standard Faiss index \citep{douze2025faiss}.

\paragraph{Construction at Pretraining Scale.}
At pretraining scale, this procedure creates a joint indexing and search
bottleneck. A corpus with \(N\) tokens yields \(N\) keys and \(N\) training
queries, requiring \(N\) searches over \(N\) keys and therefore quadratic
construction. At Pile scale, each key is a \(4096\)-dimensional hidden state
from the final layer of Pythia-6.9B \citep{biderman2023pythia}, increasing both
storage and distance computation. Together, quadratic construction and search
in a high-dimensional space make a standard Faiss pipeline infeasible.
Figure~\ref{fig:method-rewrite-pretraining-scale-construction} summarizes our
distributed Faiss pipeline, which addresses these bottlenecks through
embedding compression, index sharding, and parallel search.

First, we train an Optimized Product Quantization transform
\citep{ge2013optimized}, denoted OPQ256, which maps each \(4096\)-dimensional
key to a \(256\)-dimensional search representation before indexing. This
compression reduces storage and the cost of distance computation while
preserving the retrieval signal used to form \(p_{\mathrm{ret}}\). We then
learn an IVF partition over the compressed space, with \texttt{IndexHNSW}
\citep{malkov2018efficient} serving as the quantizer for assigning vectors and
routing queries to centroids. Each shard stores a contiguous range of
centroids as an \texttt{IndexIVFPQ} \citep{jegou2011product}. This partition
replaces one oversized index with smaller indices that support independent
GPU search.

Search proceeds through routing and local retrieval. During routing, the
\texttt{IndexHNSW} quantizer identifies the probed centroids and the shards
that contain them. Queries are then grouped by shard using this routing
information. During local retrieval, each assigned shard performs
\texttt{IndexIVFPQ} search for its query batch without random access across
shards. Writer workers collect the shard outputs, merge candidates with the
same query identifier into the final set of \(K\) nearest neighbors, and
cache \(p_{\mathrm{ret}}\) in sparse form. This design converts one
search over \(N\) queries against an index with \(N\) entries into a compressed
and sharded retrieval process with data parallelism.

\subsection{General Memory Pretraining}
\label{sec:method-rewrite-general-memory}

We pretrain general memories with \(1.4\)B, \(2.8\)B, and \(6.9\)B parameters
on \(300\)B tokens. Training at this scale is enabled by sparse \(k\)NN
distribution storage, which reduces space requirements, and distributed
streaming, which reads only the entries required by each batch.

\paragraph{Sparse \(k\)NN Distribution Storage.}
Dense \(k\)NN distributions are impractical at this scale because their
storage grows with the vocabulary size. The support of
\(p_{\mathrm{ret}}(\cdot\mid c_t)\) contains at most \(K\) token identifiers
and becomes smaller when multiple neighbors share the same target token. We
aggregate duplicate targets in FP32 and remove negligible probabilities using
a small threshold \(\epsilon\), which gives
\begin{equation}
    \mathcal{S}_t
    =
    \{y\in\mathcal{V}\mid
    p_{\mathrm{ret}}(y\mid c_t)>\epsilon\}.
    \label{eq:method-rewrite-sparse-support}
\end{equation}
We store only the retained token identifiers and probabilities in sharded
flat arrays, with offsets marking the boundary of each distribution row. For
\(N\) distribution rows and \(M\) retained token and probability pairs,
storage therefore scales as \(\mathcal{O}(N+M)\), rather than
\(\mathcal{O}(N|\mathcal{V}|)\) for dense distributions. On the Pile, each row
retains \(64.95\) token-probability pairs on average, yielding an approximately
\(250\times\) dense-to-sparse storage ratio for the \(50{,}304\)-token
vocabulary after accounting for INT64 offsets, INT32 token identifiers and
labels, and FP16 probabilities. It requires no padding and preserves direct
access to every distribution row, while the memory model still predicts over
the complete vocabulary.

\paragraph{Distributed Streaming of \(k\)NN Distributions.}
Preprocessing attaches the aligned distribution row range to each packed
language model example. Dataset shuffling therefore preserves the
correspondence between tokens and retrieval supervision. Distributed workers
use the stored offsets and read-only memory-mapped arrays to load the required
contiguous slices directly from the relevant shards. The collator retains the
standard language
modeling labels and represents \(p_{\mathrm{ret}}\) as sparse coordinate
triplets, allowing the same batching and training interface used for language
model pretraining. On the GPU, the retained probabilities are renormalized in
FP32 before computing the KL divergence between the \(k\)NN and model
distributions.

\subsection{Domain Memory Pretraining}
\label{sec:method-rewrite-domain-memory}

Domain memories follow the same architecture, objective, and sparse training
pipeline as general memories. For each domain corpus
\(\mathcal{D}_{\mathrm{mem}}\), we first adapt the model used for datastore
construction through continued pretraining and then construct the corresponding
\(k\)NN distributions from its contextual representations. These distributions
supervise a separate memory \(M_\psi\) for each domain. We apply this procedure
to biology, law, and finance corpora containing up to \(4.4\)B tokens.

\endgroup

\section{Experimental Setup}
\label{sec:experiments_setup}

\paragraph{Overview.}
We evaluate parametric memory as a plug-and-play long-term memory component in four settings. Specifically, we test (1) whether general memories trained on broad pretraining data improve diverse downstream tasks, especially knowledge-intensive tasks, across matched backbone and memory scales (\hyperlink{overview-general-memory}{\S\ref*{sec:results_general_memory}}); (2) whether pretrained general memories transfer across backbone scales (\hyperlink{overview-general-memory-transfer}{\S\ref*{sec:cross_backbone_memory_transfer}}); (3) whether the same training recipe yields effective domain memories for biology, law, and finance across backbone scales (\hyperlink{overview-specialist-memory}{\S\ref*{sec:results_specialist_qwen}}); and (4) whether domain memories transfer across model families and vocabularies (\hyperlink{overview-specialist-memory-transfer}{\S\ref*{sec:results_specialist_cross_vocab}}). Throughout our experiments, the base model remains frozen, and memory predictions are combined with the backbone distribution using the interpolation rule in Eq.~\ref{eq:memory_interpolation}.

\paragraph{Datasets.}
The general memory is trained on the deduplicated Pile \citep{gao2020pile}, which contains 207B tokens. We evaluate it across three task categories: general tasks from the Pythia suite \citep{biderman2023pythia}, including ARC-Easy, ARC-Challenge \citep{clark2018think}, LAMBADA \citep{paperno2016lambada}, LogiQA \citep{liu2020logiqa}, PIQA \citep{bisk2020piqa}, SciQ \citep{welbl2017crowdsourcing}, and WinoGrande \citep{sakaguchi2021winogrande}; knowledge-intensive tasks, including MMLU \citep{hendrycks2020measuring}, Natural Questions \citep{kwiatkowski2019natural}, TriviaQA \citep{joshi2017triviaqa}, PopQA \citep{mallen2023not}, 2WikiMultiHopQA \citep{ho2020constructing}, Bamboogle \citep{press2023measuring}, HotpotQA \citep{yang2018hotpotqa}, and GPQA-main \citep{rein2023gpqa}; and factuality and hallucination robustness on TruthfulQA \citep{lin2022truthfulqa} and HaluEval \citep{li2023halueval}. We evaluate domain memories in biology, law, and finance. For biology, we train the memory on Biology-Instructions \citep{he2024biology} after reformatting the dataset as unsupervised text and evaluate it on the Biology-Instructions benchmark \citep{he2024biology}. For law, we use the same procedure with DISC-Law-SFT \citep{yue2023disc} for training and LawBench \citep{fei2024lawbench} for evaluation. For finance, we train the memory on the book and finance subsets of FinTrain \citep{ke2025demystifying} and evaluate it on FinEval \citep{ke2025demystifying}.

\paragraph{Backbones and Baselines.}
We evaluate parametric memory across diverse model families and scales. Throughout the paper, we use deduplicated Pythia checkpoints \citep{biderman2023pythia}. For general memory experiments, we evaluate frozen backbones from 410M to 12B parameters and train 1.4B, 2.8B, and 6.9B memories initialized from the corresponding checkpoints. Domain memory experiments use Qwen3~Base backbones ranging from 0.6B to 14B parameters \citep{yang2025qwen3}. To evaluate transfer across vocabularies, we use OLMo-2-1124-7B \citep{olmo20242} and OLMo-3-1025-7B \citep{olmo2025olmo} backbones. For general memory, we compare Base + Memory configurations against decoder-only Pythia models under matched total parameter counts and training budgets. For domain memory, we compare against three baselines. CPT \citep{gururangan2020don} trains the backbone on the corresponding corpus using the same training budget as memory. LoRA \citep{hu2022lora} is applied to the query, key, value, and MLP layers, with its rank adjusted to match the parameter count of our memory modules. RAG \citep{lewis2020retrieval} employs Qwen3-Embedding-0.6B \citep{zhang2025qwen3} as the retrieval model and retrieves the top-5 documents.

\paragraph{Implementation Details.}
All experiments are conducted on \(256\) NVIDIA A800 80GB GPUs with Megatron-LM \citep{shoeybi2019megatron}. For general memory, we construct the datastore from the deduplicated Pile using Pythia-6.9B and train 1.4B, 2.8B, and 6.9B memories for \(300\)B tokens by default, matching the Pythia pretraining budget and corresponding to approximately 1.5 epochs over the datastore. The 1.4B, 2.8B, and 6.9B memories use peak learning rates of \(3\times10^{-4}\), \(2.5\times10^{-4}\), and \(2\times10^{-4}\), respectively. All runs use AdamW \citep{loshchilov2017decoupled} with \(\beta_1=0.9\), \(\beta_2=0.95\), weight decay \(0.01\), cosine decay to \(10\%\) of the peak learning rate, and \(2{,}000\) warmup steps. For domain memory, we perform one epoch of continued pretraining with Qwen3-4B-Base on each domain corpus and use the model to construct the datastore. Each memory is initialized from Qwen3-1.7B-Base and trained with a peak learning rate of \(3\times10^{-4}\) under a standard budget matching the compute of one Qwen3-8B-Base CPT epoch. For transfer across vocabularies, a trained Qwen3 memory is adapted to OLMo-2-1124-7B and OLMo-3-1025-7B by replacing the embedding layer and language model head, followed by continued training on an OLMo-3-1025-7B datastore using \(20\%\) of the standard memory training budget. We follow the standard metric for each benchmark. We use exact match for NQ-Open, TriviaQA, 2WikiMultiHopQA, Bamboogle, and HotpotQA, and standard accuracy metrics for the remaining general benchmarks. For domain evaluation, we report the mean over the metrics defined for BioInst and LawBench, while FinEval combines exact match, Matthews correlation coefficient (MCC), and ROUGE-1 across tasks. When validation data is available, the interpolation coefficient \(\alpha\) is tuned on the validation split and then fixed for test evaluation.

% Required color/style macros.
\providecommand{\metric}[1]{{\tiny(#1)}}
\providecommand{\gain}[1]{\textcolor{green!60!black}{\tiny(+#1)}}
\providecommand{\drop}[1]{\textcolor{red!80!}{\tiny(#1)}}
\providecommand{\nogain}{\textcolor{black!55}{\tiny(+0.00)}}
\providecommand{\avggain}[1]{\textcolor{green!60!black}{(+#1)}}

\begin{table}[t]
\centering
\caption{General memory scaling across Pythia backbones on 17 tasks. Scores are percentages. Base columns report frozen backbones, while +Mem-x.xB columns add an x.xB memory to the corresponding backbone. Each memory is trained on 300B tokens, matching the training budget of its backbone. \(\dagger\) marks LAMBADA (OpenAI), and GPQA-main scores average 10 random seeds.}
\label{tab:general-memory-main-scale}
\vspace{-1mm}
\resizebox{\linewidth}{!}{%
\begin{tabular}{@{}lccccccc@{}}
\toprule
\multirow{2}{*}{\raisebox{-0.85ex}{\textbf{Benchmark} \metric{Metric}}} &
\multicolumn{2}{c}{\textbf{1.4B}} &
\multicolumn{2}{c}{\textbf{2.8B}} &
\multicolumn{2}{c}{\textbf{6.9B}} &
\multicolumn{1}{c}{\textbf{12B}} \\
\cmidrule(lr){2-3}\cmidrule(lr){4-5}\cmidrule(lr){6-7}\cmidrule(l){8-8}
& \multicolumn{1}{c}{\textbf{Base}} & \multicolumn{1}{c}{\textbf{+Mem-1.4B}}
& \multicolumn{1}{c}{\textbf{Base}} & \multicolumn{1}{c}{\textbf{+Mem-2.8B}}
& \multicolumn{1}{c}{\textbf{Base}} & \multicolumn{1}{c}{\textbf{+Mem-6.9B}}
& \multicolumn{1}{c}{\textbf{Base}} \\
\midrule
\rowcolor{gray!8}\multicolumn{8}{@{}l@{}}{\textbf{General Tasks}}\\
ARC-Easy \metric{Acc.}      & 61.83 & 62.42 & 63.64 & 65.82 & 68.39 & 70.12 & 70.58 \\
ARC-Challenge \metric{Acc.} & 27.39 & 27.56 & 30.03 & 30.29 & 33.11 & 35.67 & 33.45 \\
LAMBADA\textsuperscript{\(\dagger\)} \metric{Acc.} & 61.91 & 63.30 & 65.09 & 65.75 & 68.76 & 69.94 & 70.99 \\
LogiQA \metric{Acc.}        & 22.43 & 22.73 & 21.66 & 21.97 & 23.04 & 23.04 & 21.97 \\
PIQA \metric{Acc.}          & 72.20 & 73.23 & 74.05 & 74.92 & 76.01 & 76.17 & 76.28 \\
SciQ \metric{Acc.}          & 86.30 & 88.00 & 88.10 & 90.20 & 90.90 & 91.50 & 92.70 \\
WinoGrande \metric{Acc.}    & 56.12 & 59.43 & 58.56 & 61.25 & 62.98 & 64.72 & 65.59 \\
\midrule
\rowcolor{gray!8}\multicolumn{8}{@{}l@{}}{\textbf{Knowledge}}\\
MMLU \metric{Acc.}          & 23.76 & 24.75 & 24.72 & 24.87 & 26.26 & 26.95 & 25.67 \\
NQ-Open \metric{EM}         & 2.80 & 3.88 & 3.82 & 5.15 & 4.68 & 6.32 & 6.32 \\
TriviaQA \metric{EM}        & 8.49 & 10.08 & 8.30 & 17.11 & 22.21 & 25.58 & 26.80 \\
PopQA \metric{Acc.}         & 10.65 & 11.90 & 11.45 & 12.83 & 14.00 & 15.13 & 14.95 \\
2WikiMultiHopQA \metric{EM} & 16.89 & 22.57 & 18.96 & 21.64 & 20.60 & 21.11 & 20.57 \\
Bamboogle \metric{EM}       & 1.60 & 2.40 & 0.80 & 2.40 & 4.00 & 5.60 & 2.40 \\
HotpotQA \metric{EM}        & 6.75 & 8.44 & 8.83 & 9.49 & 8.97 & 11.78 & 11.11 \\
GPQA-main \metric{Acc.}     & 24.53 & 27.90 & 24.91 & 25.92 & 24.11 & 26.36 & 25.49 \\
\midrule
\rowcolor{gray!8}\multicolumn{8}{@{}l@{}}{\textbf{Hallucination}}\\
TruthfulQA \metric{Acc.}    & 30.63 & 30.46 & 28.47 & 27.92 & 28.49 & 27.81 & 26.91 \\
HaluEval \metric{Acc.}      & 42.67 & 45.13 & 44.68 & 45.83 & 40.53 & 44.59 & 41.23 \\
\midrule
\rowcolor{SkyBlue!20}
\textbf{AVG}                & 32.76 & \textbf{34.36}\,\avggain{1.60} & 33.89 & \textbf{35.49}\,\avggain{1.60} & 36.30 & \textbf{37.79}\,\avggain{1.49} & 37.24 \\
\bottomrule
\end{tabular}%
}
\end{table}

\section{Experimental Results}
\label{sec:experiments_results}

\makeatletter
\DeclareRobustCommand{\memorysubsectiontarget}[1]{%
  \Hy@raisedlink{\pdfdest name{#1} fith\relax}%
}
\makeatother

\subsection{\texorpdfstring{\protect\memorysubsectiontarget{overview-general-memory}General Memory at Scale}{General Memory at Scale}}
\label{sec:results_general_memory}

Table~\ref{tab:general-memory-main-scale} evaluates general memory on frozen
Pythia backbones. Across 17 benchmarks, attaching a memory of equal size raises
AVG at every scale. The score increases from \(32.76\) to \(34.36\) at 1.4B,
from \(33.89\) to \(35.49\) at 2.8B, and from \(36.30\) to \(37.79\) at 6.9B.
More importantly, the 1.4B backbone paired with a 1.4B memory reaches \(34.36\),
while the 2.8B backbone reaches \(33.89\). These configurations use the same
total parameter count and training budget, yet the memory configuration
outperforms the larger backbone. The 6.9B backbone paired with a 6.9B memory
also reaches \(37.79\) and surpasses the frozen 12B backbone at \(37.24\).
The largest gains concentrate on knowledge
tasks, including TriviaQA (\(8.30\!\rightarrow\!17.11\)) with the 2.8B backbone,
2WikiMultiHopQA (\(16.89\!\rightarrow\!22.57\)) with the 1.4B backbone, and
HotpotQA (\(8.97\!\rightarrow\!11.78\)) with the 6.9B backbone. These results
establish parametric memory as an effective knowledge component for frozen models.

The improvements are broad rather than driven by a few outliers. Across the
\(51\) combinations of tasks and scales, memory improves \(47\) and matches the
base in one more. The gains also extend beyond knowledge tasks.
A 1.4B backbone with 1.4B memory improves WinoGrande and GPQA-main by \(3.31\)
and \(3.37\) points, while 6.9B counterparts improve HaluEval by \(4.06\). Together,
these results show that general memory provides substantial and consistent
improvements across diverse tasks and model scales while preserving the general
and reasoning performance of frozen backbones.

\subsection{\texorpdfstring{\protect\memorysubsectiontarget{overview-general-memory-transfer}General Memory Transfer Across Backbones}{General Memory Transfer Across Backbones}}
\label{sec:cross_backbone_memory_transfer}

A key architectural advantage is that a trained memory can seamlessly
augment different backbones without retraining. We therefore study general
memory across Pythia backbone scales.
Figure~\ref{fig:general-memory-avg-selected-path-scatter-log} plots zero-shot
AVG against total parameter count. All 18 configurations outperform their
backbone alone, and many lie above the scaling trend of the
base models. This benefit does not require the backbone and memory to have
matched capacity. The largest gains emerge when larger memories augment smaller
backbones. Pairing a 410M backbone with a 6.9B memory raises AVG from
\(29.86\) to \(37.34\), surpassing the \(37.24\) score of the 12B base model
with \(39\%\) fewer total parameters. These results show that long-term memory
need not be entangled with reasoning in one parameter set. A small backbone
with a larger standalone memory is more parameter-efficient than backbone
scaling.

All memories use the same training budget, keeping training exposure fixed
across parameter allocations. At matched AVG, configurations along the Base +
Memory curve use \(33\%\), \(32\%\), and \(42\%\) fewer total parameters than
the 2.8B, 6.9B, and 12B base models, respectively. This makes reusable memory an
efficient alternative to scaling the backbone alone. At the same total
parameter count, a Base + Memory configuration can offer lower inference
latency than a single backbone because the two components can run in parallel.
Appendix~\ref{sec:appendix_general_memory_details} reports results for
individual tasks.

\subsection{\texorpdfstring{\protect\memorysubsectiontarget{overview-specialist-memory}Domain Memory Across Backbones}{Domain Memory Across Backbones}}
\label{sec:results_specialist_qwen}

Our memory is effective at scale. We next examine whether memory
supports domain specialization.
Table~\ref{tab:specialist-memory-qwen} evaluates 1.7B domain memories for
biology, law, and finance with Qwen3 backbones from 0.6B to 14B. Domain
memory achieves the highest average across these domains at every scale. The
average gains over the frozen backbones are \(9.88\), \(9.64\), \(10.00\),
\(9.09\), and \(9.99\) points
for the 0.6B, 1.7B, 4B, 8B, and 14B backbones, respectively. Memory outperforms
the strongest baseline at each scale by at least \(4.05\) points, with the
largest margin of \(8.53\) points for the 0.6B backbone.
These results further establish domain memory as a reusable component for
domain knowledge. The biology, law, and finance memories improve every frozen
backbone, covering all \(15\) evaluations. With the 14B backbone, the respective
gains on BioInst, LawBench, and FinEval are \(17.96\), \(8.97\), and \(3.04\)
points. Switching domains only requires activating the corresponding domain
memory from the memory bank at inference, while the backbone remains frozen and
requires no continued pretraining.
Appendix~\ref{sec:appendix_specialist_memory_details} reports results for the
0.6B memories.

\providecommand{\spgain}[1]{}
\providecommand{\spdrop}[1]{}
\renewcommand{\spgain}[1]{\,{\raisebox{-0.35ex}{\scriptsize\textcolor{green!60!black}{+#1}}}}
\renewcommand{\spdrop}[1]{\,{\raisebox{-0.35ex}{\scriptsize\textcolor{red!80!}{-#1}}}}
\providecommand{\avgain}[1]{\textcolor{green!60!black}{(+#1)}}
\providecommand{\avdrop}[1]{\textcolor{red!80!}{(-#1)}}
\providecommand{\spfamilyrow}[1]{\rowcolor{gray!8}\multicolumn{5}{c}{\textit{#1}}\\}
\providecommand{\spmemrow}{\rowcolor{SkyBlue!20}}
\providecommand{\spdomainhead}{%
\toprule
\textbf{Model} &
\textbf{BioInst} &
\textbf{LawBench} &
\textbf{FinEval} &
\textbf{Avg} \\
\midrule
}
\providecommand{\spdomaintail}{%
\bottomrule%
}

\begin{table}[!t]
\centering
\caption{Domain memory results with Qwen3 backbones under zero-shot domain evaluation. BioInst, LawBench, and FinEval use their respective benchmark metrics. Score changes relative to the corresponding frozen backbone are shown in smaller type. \(+\)Mem-1.7B denotes a 1.7B domain memory attached to a frozen Qwen3 backbone.}
\label{tab:specialist-memory-qwen}
\begingroup
\small
\setlength{\tabcolsep}{5.5pt}
\renewcommand{\arraystretch}{1.02}
\begin{tabular}{@{}lcccc@{}}
\spdomainhead
\spfamilyrow{Qwen3 Family}
\textbf{Qwen3-0.6B-Base}    & 5.78 & 17.87 & 23.31 & 15.65 \\
\textit{+CPT}               & 7.94\spgain{2.16} & 20.39\spgain{2.52} & 14.90\spdrop{8.41} & 14.41\avdrop{1.24} \\
\textit{+LoRA}              & 10.36\spgain{4.58} & 21.10\spgain{3.23} & 19.54\spdrop{3.77} & 17.00\avgain{1.35} \\
\textit{+RAG}               & 2.96\spdrop{2.82} & 17.82\spdrop{0.05} & 21.37\spdrop{1.94} & 14.05\avdrop{1.60} \\
\spmemrow\textit{+Mem-1.7B} & \textbf{19.94}\spgain{14.16} & \textbf{26.53}\spgain{8.66} & \textbf{30.12}\spgain{6.81} & \textbf{25.53}\avgain{9.88} \\
\midrule
\textbf{Qwen3-1.7B-Base}    & 5.39 & 26.86 & 27.26 & 19.84 \\
\textit{+CPT}               & 4.84\spdrop{0.55} & 28.86\spgain{2.00} & 29.02\spgain{1.76} & 20.91\avgain{1.07} \\
\textit{+LoRA}              & 9.01\spgain{3.62} & 28.53\spgain{1.67} & \textbf{38.74}\spgain{11.48} & 25.43\avgain{5.59} \\
\textit{+RAG}               & 8.07\spgain{2.68} & 29.61\spgain{2.75} & 36.63\spgain{9.37} & 24.77\avgain{4.93} \\
\spmemrow\textit{+Mem-1.7B} & \textbf{23.21}\spgain{17.82} & \textbf{30.78}\spgain{3.92} & 34.45\spgain{7.19} & \textbf{29.48}\avgain{9.64} \\
\midrule
\textbf{Qwen3-4B-Base}      & 5.02 & 27.47 & 37.43 & 23.31 \\
\textit{+CPT}               & 9.14\spgain{4.12} & 36.10\spgain{8.63} & 30.22\spdrop{7.21} & 25.15\avgain{1.85} \\
\textit{+LoRA}              & 8.06\spgain{3.04} & 33.65\spgain{6.18} & 33.52\spdrop{3.91} & 25.08\avgain{1.77} \\
\textit{+RAG}               & 8.34\spgain{3.32} & 32.62\spgain{5.15} & 39.13\spgain{1.70} & 26.70\avgain{3.39} \\
\spmemrow\textit{+Mem-1.7B} & \textbf{23.48}\spgain{18.46} & \textbf{36.89}\spgain{9.42} & \textbf{39.54}\spgain{2.11} & \textbf{33.30}\avgain{10.00} \\
\midrule
\textbf{Qwen3-8B-Base}      & 4.82 & 32.67 & 43.51 & 27.00 \\
\textit{+CPT}               & 9.79\spgain{4.97} & 39.24\spgain{6.57} & 40.00\spdrop{3.51} & 29.68\avgain{2.68} \\
\textit{+LoRA}              & 5.32\spgain{0.50} & 33.55\spgain{0.88} & 42.43\spdrop{1.08} & 27.10\avgain{0.10} \\
\textit{+RAG}               & 7.78\spgain{2.96} & 40.67\spgain{8.00} & \textbf{46.97}\spgain{3.46} & 31.81\avgain{4.81} \\
\spmemrow\textit{+Mem-1.7B} & \textbf{20.02}\spgain{15.20} & \textbf{41.91}\spgain{9.24} & 46.33\spgain{2.82} & \textbf{36.09}\avgain{9.09} \\
\midrule
\textbf{Qwen3-14B-Base}     & 4.01 & 35.45 & 44.25 & 27.90 \\
\textit{+RAG}               & 7.70\spgain{3.69} & 44.38\spgain{8.93} & 42.81\spdrop{1.44} & 31.63\avgain{3.73} \\
\spmemrow\textit{+Mem-1.7B} & \textbf{21.97}\spgain{17.96} & \textbf{44.42}\spgain{8.97} & \textbf{47.29}\spgain{3.04} & \textbf{37.89}\avgain{9.99} \\
\spdomaintail
\end{tabular}
\endgroup
\end{table}

\subsection{\texorpdfstring{\protect\memorysubsectiontarget{overview-specialist-memory-transfer}Domain Memory Transfer Across Vocabularies}{Domain Memory Transfer Across Vocabularies}}
\label{sec:results_specialist_cross_vocab}

We further examine domain memory transfer across vocabularies. We adapt the
trained Qwen3 memories for biology, law, and finance from
Table~\ref{tab:specialist-memory-qwen} on datastores built with the OLMo-3
vocabulary. With only \(20\%\) of the standard memory training budget, the
transferred memories achieve the best average for both OLMo backbones in
Table~\ref{tab:specialist-memory-olmo}. Average scores rise from \(19.57\) to
\(23.83\) on OLMo-2-7B and from \(18.67\) to \(26.44\) on OLMo-3-7B, gains of
\(4.26\) and \(7.77\) points. Memory improves all six evaluations, gaining
\(4.09\) and \(1.77\) points on BioInst, \(8.52\) and \(9.21\) on LawBench, and
\(0.18\) and \(12.33\) on FinEval for OLMo-2 and OLMo-3, respectively. CPT leads
on BioInst for both backbones and LawBench for OLMo-2, while memory provides the
best overall average. The consistent gains demonstrate effective transfer
across vocabularies with limited training.

\begin{table}[!t]
\centering
\caption{Domain memory transfer across vocabularies to OLMo backbones.
Training uses \(20\%\) of the budget in
Table~\ref{tab:specialist-memory-qwen}, and evaluation follows the same setting.}
\label{tab:specialist-memory-olmo}
\begingroup
\small
\setlength{\tabcolsep}{5.5pt}
\renewcommand{\arraystretch}{1.05}
\begin{tabular}{@{}lcccc@{}}
\spdomainhead
\spfamilyrow{OLMo Family (cross-vocabulary transfer, 20\% training budget)}
\textbf{OLMo-2-7B}          & 3.90 & 6.37 & 48.43 & 19.57 \\
\textit{+CPT}               & \textbf{9.51}\spgain{5.61} & \textbf{22.95}\spgain{16.58} & 34.61\spdrop{13.82} & 22.36\avgain{2.79} \\
\textit{+LoRA}              & 9.44\spgain{5.54} & 13.21\spgain{6.84} & 32.34\spdrop{16.09} & 18.33\avdrop{1.24} \\
\textit{+RAG}               & 3.70\spdrop{0.20} & 12.11\spgain{5.74} & 40.88\spdrop{7.55} & 18.90\avdrop{0.67} \\
\spmemrow\textit{+Mem-1.7B} & 7.98\spgain{4.09} & 14.89\spgain{8.52} & \textbf{48.61}\spgain{0.18} & \textbf{23.83}\avgain{4.26} \\
\midrule
\textbf{OLMo-3-7B}          & 6.39 & 13.19 & 36.43 & 18.67 \\
\textit{+CPT}               & \textbf{13.69}\spgain{7.30} & 20.29\spgain{7.10} & 38.89\spgain{2.46} & 24.29\avgain{5.62} \\
\textit{+LoRA}              & 6.95\spgain{0.56} & 16.40\spgain{3.21} & 28.21\spdrop{8.22} & 17.19\avdrop{1.48} \\
\textit{+RAG}               & 5.95\spdrop{0.44} & 21.85\spgain{8.66} & 33.10\spdrop{3.33} & 20.30\avgain{1.63} \\
\spmemrow\textit{+Mem-1.7B} & 8.16\spgain{1.77} & \textbf{22.40}\spgain{9.21} & \textbf{48.76}\spgain{12.33} & \textbf{26.44}\avgain{7.77} \\
\spdomaintail
\end{tabular}
\endgroup
\end{table}

\clearpage
\section{Analysis}
\label{sec:analysis}

\subsection{Few-Shot Robustness and Knowledge Task Improvements}
\label{sec:analysis_fewshot_knowledge}

\begin{figure}[t]
\centering
\includegraphics[width=0.94\textwidth]{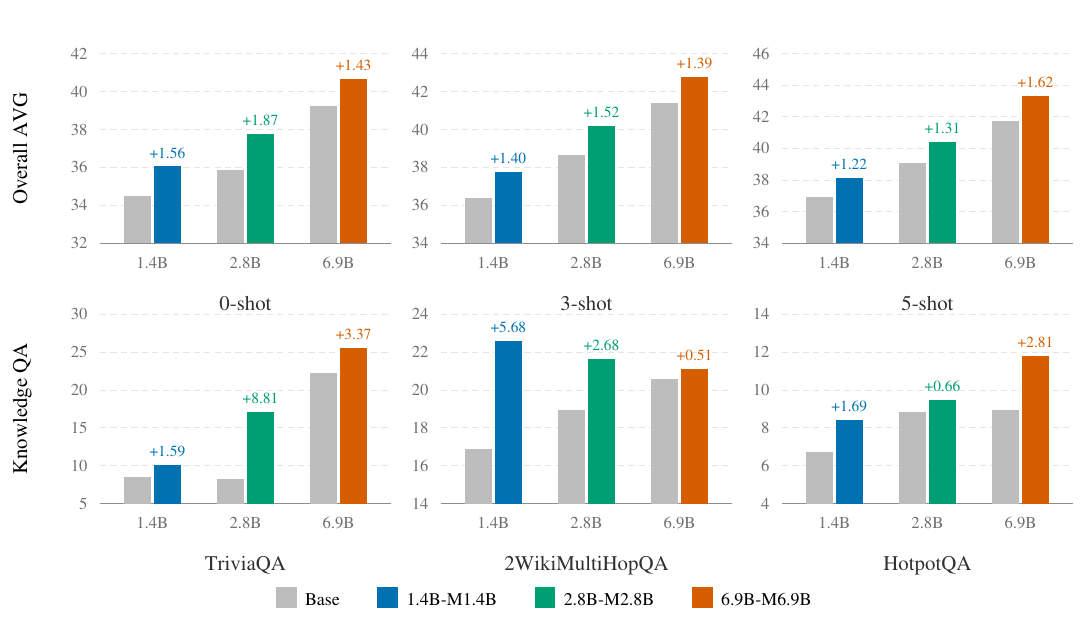}
\caption{\textbf{Top:} AVG over the same 13 tasks under zero-shot, three-shot,
and five-shot settings. \textbf{Bottom:} Zero-shot results on TriviaQA,
2WikiMultiHopQA, and HotpotQA. Gray bars denote frozen backbones. The notation
\(x.x\text{B}\)-M\(x.x\text{B}\) represents an \(x.x\)B backbone paired with
an \(x.x\)B memory.}
\label{fig:general-memory-summary-bars}
\end{figure}

Figure~\ref{fig:general-memory-summary-bars} tests whether general memory remains
useful when in-context examples are available. For a controlled comparison, all
three averages use the same 13 tasks. The set contains ARC-Easy,
ARC-Challenge, LAMBADA, LogiQA, MMLU, PIQA, SciQ, WinoGrande, NQ-Open,
TriviaQA, 2WikiMultiHopQA, Bamboogle, and HotpotQA. Bamboogle retains its
zero-shot protocol. Across three model scales, memory improves AVG by
\(1.43\)--\(1.87\) points in the zero-shot setting, \(1.39\)--\(1.52\) with
three shots, and \(1.22\)--\(1.62\) with five shots. The gain appears at every
model scale and prompt setting. This indicates that memory remains
complementary to in-context examples.
The bottom row reports zero-shot results on TriviaQA, 2WikiMultiHopQA, and
HotpotQA. In Table~\ref{tab:general-memory-main-scale}, memory improves all eight
zero-shot knowledge tasks at every model scale. Averaged over scales, the gains
on the three displayed tasks are
\(4.59\), \(2.96\), and \(1.72\) points, respectively. These improvements show
that the result is not driven by a single benchmark. The larger effects on open-domain and multi-hop QA suggest that memory supplies factual evidence that is not included in the prompt.

\subsection{Case Study: Memory Strengthens Factual Evidence}
\label{sec:case_study}

\begin{figure}[H]
    \centering
    \includegraphics[width=\linewidth]{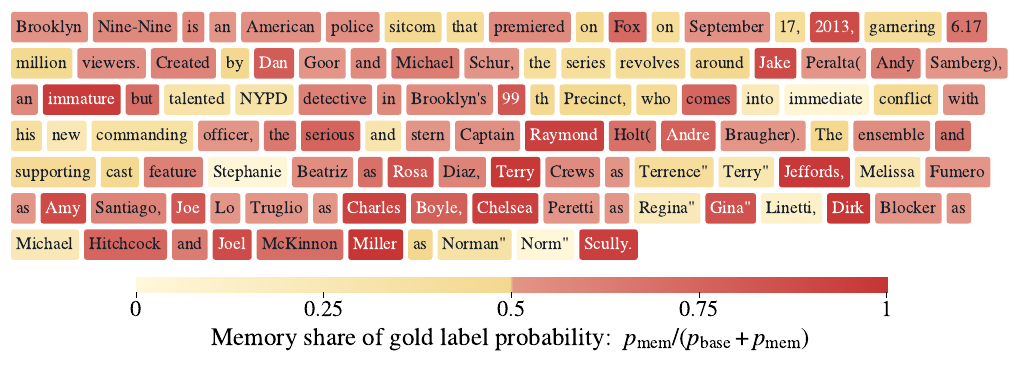}
    \caption{Memory probability share on the HotpotQA context titled
    \textit{Brooklyn Nine-Nine}. Colors encode
    $p_{\mathrm{mem}}/(p_{\mathrm{base}}+p_{\mathrm{mem}})$. Values below
    $0.5$ favor the frozen base model, while values above $0.5$ favor the
    memory model.}
    \label{fig:case-study-brooklyn-nine-nine}
\end{figure}

Figure~\ref{fig:case-study-brooklyn-nine-nine} compares gold label probabilities
at the word level within a supporting HotpotQA paragraph titled
\textit{Brooklyn Nine-Nine} \citep{yang2018hotpotqa}. The case study uses frozen
Pythia-1.4B as the base model and the 1.4B memory in
Table~\ref{tab:general-memory-main-scale}. For each displayed word $w_i$, let
$p_{\mathrm{base}}(w_i)$ and $p_{\mathrm{mem}}(w_i)$ denote the geometric mean
gold label probabilities of its constituent subword tokens. We visualize the
normalized memory share
\begin{equation}
\label{eq:normalized-memory-share}
s_i = \frac{p_{\mathrm{mem}}(w_i)}
           {p_{\mathrm{base}}(w_i)+p_{\mathrm{mem}}(w_i)}.
\end{equation}
A value of $s_i=0.5$ indicates equal support, while $s_i>0.5$ indicates greater
support from memory. The memory model assigns larger shares to many entity and
attribute words. This pattern is consistent with the aggregate QA gains and
suggests that memory reinforces factual evidence. Additional examples appear
in Appendix~\ref{sec:appendix_case_studies}.

\subsection{Memory Size}
\label{sec:analysis_memory_size}

\begin{figure}[t]
    \centering
    \includegraphics[width=0.95\textwidth]{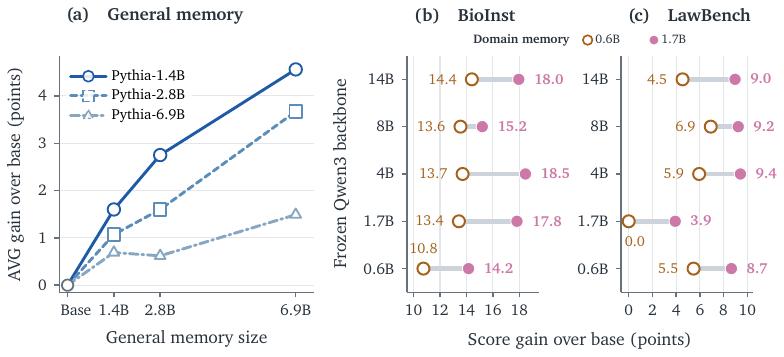}
    \caption{Effect of memory size.
    \textbf{(a)} AVG gains for three general memory sizes on frozen Pythia
    backbones. \textbf{(b,c)} Gains on BioInst and LawBench for frozen Qwen3
    backbones. A 0.6B or 1.7B marker denotes the backbone paired with a
    domain memory of that size.}
    \label{fig:memory-size-effect}
\end{figure}

Figure~\ref{fig:memory-size-effect} examines how performance changes as memory
capacity increases while keeping the backbone fixed. In the general setting,
every tested memory size improves AVG over the corresponding base across all
three Pythia backbones. For each fixed backbone, the gain also grows overall
with memory capacity. The 6.9B memory therefore yields the largest AVG gain in
every case: \(4.56\), \(3.67\), and \(1.49\) points for the 1.4B, 2.8B, and 6.9B
backbones, respectively. This consistent trend shows that general memory scales
across backbone sizes, with the largest gains on the smaller frozen backbones.

In the domain setting, the 1.7B memory outperforms the 0.6B memory for every
Qwen3 backbone on both BioInst and LawBench. Across backbones, adding the 1.7B
memory increases scores by as much as \(18.5\) points on BioInst and \(9.4\)
points on LawBench. Increasing memory capacity from 0.6B to 1.7B provides up to
an additional \(4.8\) points on BioInst and \(4.4\) points on LawBench. Taken
together, the general and domain results show a consistent scaling trend,
with larger memory capacities generally improving performance across backbone
sizes and domains.

\subsection{Memory Training Budget}
\label{sec:analysis_training_budget}

Table~\ref{tab:memory-training-budget} examines how the amount of memory
training affects performance. For general memory, extending training from 1.5
to 5 epochs raises the average across general tasks from \(56.67\) to \(57.33\)
and the knowledge average from \(13.99\) to \(14.17\). The domain memory
experiments follow the setting in Table~\ref{tab:specialist-memory-olmo} and
compare \(20\%\) of the standard training budget with the full budget. Using the
full budget improves Avg for every combination of OLMo backbone and memory size.
The gains are \(0.34\) and \(3.51\) points for the 0.6B and 1.7B memories on
OLMo-2, and \(0.94\) and \(1.75\) points on OLMo-3. Across both settings,
increasing the memory training budget consistently improves aggregate
performance, showing that memory continues to benefit from additional training.

\begin{table}[!t]
\centering
\caption{Effect of training budget for general and domain memory. The
domain evaluation setting follows Table~\ref{tab:specialist-memory-olmo}.
Bold marks the better result within each comparison.}
\label{tab:memory-training-budget}
\begingroup
\small
\renewcommand{\arraystretch}{1.06}
\setlength{\tabcolsep}{4.5pt}
\begin{tabular}{@{}lllcccc@{}}
\toprule
\rowcolor{gray!8}
\multicolumn{7}{c}{\textbf{General memory: Pythia-1.4B + Mem-1.4B}} \\
\midrule
\multicolumn{2}{l}{\textbf{Memory checkpoint}} & \textbf{Epochs} &
\multicolumn{2}{c}{\textbf{General Tasks}} & \multicolumn{2}{c}{\textbf{Knowledge}} \\
\midrule
\multicolumn{2}{l}{Default} & 1.5 & \multicolumn{2}{c}{56.67} & \multicolumn{2}{c}{13.99} \\
\multicolumn{2}{l}{Longer training} & 5 & \multicolumn{2}{c}{\textbf{57.33}} & \multicolumn{2}{c}{\textbf{14.17}} \\
\midrule
\rowcolor{gray!8}
\multicolumn{7}{c}{\textbf{Domain memory: OLMo transfer}} \\
\midrule
\textbf{Backbone} & \textbf{Memory} & \textbf{Budget} & \textbf{BioInst} & \textbf{LawBench} & \textbf{FinEval} & \textbf{Avg} \\
\midrule
\multirow{4}{*}{OLMo-2-7B}
& 0.6B & 20\% & 9.86 & 16.78 & \textbf{50.15} & 25.60 \\
& 0.6B & Full & \textbf{11.15} & \textbf{17.76} & 48.92 & \textbf{25.94} \\
& 1.7B & 20\% & 7.98 & 14.89 & 48.61 & 23.83 \\
& 1.7B & Full & \textbf{11.72} & \textbf{17.15} & \textbf{53.14} & \textbf{27.34} \\
\midrule
\multirow{4}{*}{OLMo-3-7B}
& 0.6B & 20\% & 9.28 & 22.34 & \textbf{46.66} & 26.09 \\
& 0.6B & Full & \textbf{10.79} & \textbf{24.02} & 46.29 & \textbf{27.03} \\
& 1.7B & 20\% & 8.16 & 22.40 & 48.76 & 26.44 \\
& 1.7B & Full & \textbf{11.64} & \textbf{23.74} & \textbf{49.20} & \textbf{28.19} \\
\bottomrule
\end{tabular}
\endgroup
\end{table}

\subsection{Memory Training Objective}
\label{sec:analysis_memory_objective}

To determine whether the gains come from memory training rather than the
interpolation interface, we replace our 1.7B memory with a Qwen3-1.7B-Base model
trained by standard CPT on the corresponding domain corpus. Within each
domain, the two modules match in training data, FLOPs, capacity, and inference
interface. We evaluate this control with frozen Qwen3-1.7B-Base and Qwen3-8B-Base backbones on
BioInst and LawBench.

Table~\ref{tab:memory-objective-control} shows memory
outperforms attached CPT. On BioInst, memory exceeds attached CPT by \(10.21\)
points with the 1.7B backbone and \(8.71\) points with the 8B backbone. On
LawBench, the corresponding margins are \(1.68\) and \(3.93\) points. The
consistent advantage in all four comparisons attributes the additional gains
to the memory objective rather than the interpolation interface.

\begin{table}[!t]
\centering
\caption{Attached module control on BioInst and LawBench. The 1.7B CPT and
memory modules use matched training and the same interpolation interface with
each frozen backbone.}
\label{tab:memory-objective-control}
\begingroup
\small
\setlength{\tabcolsep}{8pt}
\renewcommand{\arraystretch}{1.02}
\begin{tabular}{@{}llcc@{}}
\toprule
\textbf{Domain} & \textbf{Attached module} & \textbf{Qwen3 1.7B} & \textbf{Qwen3 8B} \\
\midrule
\multirow{3}{*}{BioInst}
& None & 5.39 & 4.82 \\
& CPT (1.7B) & 13.00 & 11.31 \\
& Memory (1.7B) & \textbf{23.21} & \textbf{20.02} \\
\midrule
\multirow{3}{*}{LawBench}
& None & 26.86 & 32.67 \\
& CPT (1.7B) & 29.10 & 37.98 \\
& Memory (1.7B) & \textbf{30.78} & \textbf{41.91} \\
\bottomrule
\end{tabular}
\endgroup
\end{table}

\subsection{Extractable Memorization Evaluation}
\label{sec:extractable_memorization}

\begin{table}[!t]
    \centering
    \small
    \setlength{\tabcolsep}{5pt}
    \caption{Extractable memorization on BioInst training examples. \(N\) denotes
    the number of evaluated training examples for each probe.}
    \label{tab:extractable-memorization}
    \begin{tabular}{@{}lrrr@{}}
        \toprule
        Probe & \(N\) & CPT model & memory model \\
        \midrule
        Verbatim continuation, \(\mathrm{EM@8,16}\) & 1,024 & 42.4\% & \textbf{49.7\%} \\
        Domain anchor completion & 1,024 & 22.6\% & \textbf{56.5\%} \\
        \bottomrule
    \end{tabular}
\end{table}

\begin{figure}[!t]
    \centering
    \begin{minipage}[t]{0.485\linewidth}
        \vspace{0pt}
        \begin{tcolorbox}[
            colback=blue!3,
            colframe=sectionblue,
            boxrule=0.55pt,
            arc=1.5pt,
            height=36mm,
            left=4pt,right=4pt,top=3pt,bottom=3pt,
            before skip=0pt,after skip=0pt
        ]
        \scriptsize\raggedright
        \textbf{Verbatim continuation (FunctionEC)}\\[-1pt]
        \textcolor{black!60}{Prefix}\quad
        \texttt{\ldots{} Output: EC3.4.24.-}\\
        \textcolor{black!60}{Gold}\quad
        \texttt{,EC3.4.24.69 identifies the enzyme's function}\\
        \textcolor{darkred}{\(\times\) CPT model}\quad
        \texttt{,EC3.4.24.69.}\\
        \textcolor{darkgreen}{\(\checkmark\) memory model}\quad
        \texttt{,EC3.4.24.69 identifies the enzyme's function}
        \end{tcolorbox}
    \end{minipage}
    \hfill
    \begin{minipage}[t]{0.485\linewidth}
        \vspace{0pt}
        \begin{tcolorbox}[
            colback=violet!3,
            colframe=violet!65!black,
            boxrule=0.55pt,
            arc=1.5pt,
            height=36mm,
            left=4pt,right=4pt,top=3pt,bottom=3pt,
            before skip=0pt,after skip=0pt
        ]
        \scriptsize\raggedright
        \textbf{Domain anchor completion (EMP)}\\[-1pt]
        \textcolor{black!60}{Prefix}\quad
        \texttt{\ldots{} Output: EMP, or Epigenetic Marks Prediction, aims to identify}\\
        \textcolor{black!60}{Gold}\quad
        \texttt{epigenetic changes, successfully confirmed}\\
        \textcolor{darkred}{\(\times\) CPT model}\quad
        \texttt{acetylation and methylation nucleosome occupancies, \ldots{}}\\
        \textcolor{darkgreen}{\(\checkmark\) memory model}\quad
        \texttt{epigenetic changes, successfully confirmed \ldots{}}
        \end{tcolorbox}
    \end{minipage}
    \caption{Model outputs for the two extractable memorization probes. Each panel
    shows the provided prefix, gold target, and continuations from the CPT and
    memory models. Ellipses abbreviate the task context and text beyond the scored span.}
    \label{fig:extractable-memorization-examples}
\end{figure}

We next ask whether the memory makes training content more
extractable, beyond its effect on downstream accuracy. We conduct this analysis
in the domain memory setting. Following the discoverable extraction protocol
that conditions on a prefix to recover its suffix
\citep{carlini2022quantifying,hayes2025measuring}, we evaluate the 1.7B BioInst
memory model against a Qwen3-1.7B-Base CPT model on the same 1,024 training prompts.
Both models are trained on the same BioInst training data under the same budget
and differ only in the training mechanism. \emph{Verbatim Continuation} provides
the task context and the first eight output tokens from an enzyme function
annotation task (\texttt{FunctionEC}). \(\mathrm{EM@8,16}\) is the percentage of examples for which
greedy decoding exactly reproduces the next 16 target tokens.
\emph{Domain Anchor Verbatim Completion} instead hides a four-word
span from an epigenetic mark prediction task (\texttt{EMP}) and tests whether
generation starts with that exact span.

Table~\ref{tab:extractable-memorization} shows that the memory model raises
strict \(\mathrm{EM@8,16}\) from \(42.4\%\) to \(49.7\%\). On domain anchor
completion, the gap widens from \(22.6\%\) to
\(56.5\%\). Figure~\ref{fig:extractable-memorization-examples} compares the
outputs of the memory and CPT models under the two probes. Together, these results
show that the memory provides stronger traceability to its training data,
suggesting that it retains content from the training domain in a form that can be more
reliably recovered from partial context.
Appendix~\ref{sec:appendix_extractable_memorization} gives the construction,
analysis by suffix frequency, paired counts, and additional examples.

\subsection{Memory Fidelity to Retrieval Supervision}
\label{sec:analysis_knn_distribution}

We evaluate how closely a parametric memory matches its retrieval-based training
targets. We use the 1.7B BioInst domain setting and draw 1,024 samples from
the training datastore, each containing a context and its next token. For sample
\(i\), \(P_i\) is the stored \(k\)NN target distribution and \(Q_i\) is the memory
distribution without datastore access. Their agreement measures how well the
memory internalizes retrieval supervision. We use four complementary metrics.
Top token match compares modes, while KL divergence and total variation (TV)
compare full distributions. Smaller values indicate closer agreement. KL is
sensitive when \(Q_i\) underweights tokens with high probability under \(P_i\).
TV is half the \(\ell_1\) distance and equals the probability mass that must be
reassigned for exact agreement. Pearson \(r\) measures whether the memory
probability for the token selected by \(k\)NN tracks the corresponding target
probability across samples.

\par\vspace{0.5em}
\noindent
\begin{minipage}[c]{0.43\linewidth}
    \centering
    \captionsetup{hypcap=false,font=small}
    \includegraphics[width=0.86\linewidth]{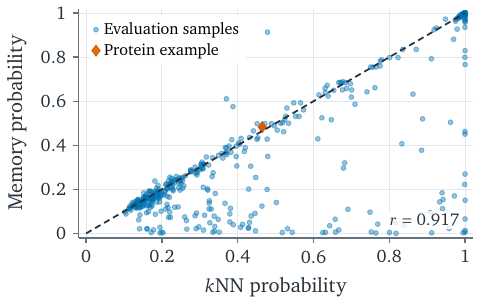}
    \captionof{figure}{\(k\)NN and memory probabilities for the mode of the
    \(k\)NN distribution. The
    dashed line marks equality.}
    \label{fig:knn-distribution-scatter}
\end{minipage}\hfill
\begin{minipage}[c]{0.53\linewidth}
    \centering
    \captionsetup{hypcap=false,font=small}
\captionof{table}{Agreement between \(k\)NN targets and memory distributions
over 1,024 BioInst samples. Top token match compares modes, KL and TV compare
full distributions, and Pearson \(r\) compares probabilities for the \(k\)NN mode.}
    \label{tab:knn-distribution-summary}
    \begingroup
\small
\setlength{\tabcolsep}{7pt}
\renewcommand{\arraystretch}{1.04}
\begin{tabular}{@{}lr@{}}
    \toprule
    Metric & All samples \\
    \midrule
    Top token match & 86.62\% \\
    Mean KL & 0.1820 \\
    Mean TV & 0.0823 \\
    Pearson \(r\) & 0.9174 \\
    \bottomrule
\end{tabular}
\endgroup

\end{minipage}

\par\vspace{0.5em}

\begin{figure}[H]
    \centering
    \includegraphics[width=0.86\linewidth]{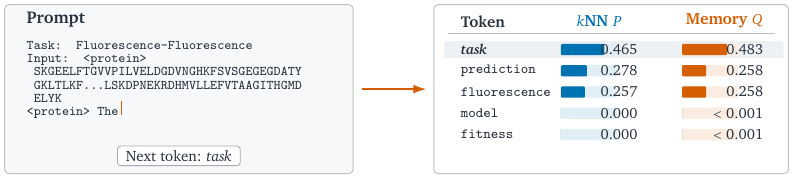}
    \caption{Protein example. The sequence is abbreviated. The target
    continuation is highlighted. The right panel compares \(k\)NN and memory
    probabilities for five candidate tokens.}
    \label{fig:knn-distribution-example}
\end{figure}

Figure~\ref{fig:knn-distribution-scatter} shows strong correspondence between
\(k\)NN and memory probabilities for the mode of the \(k\)NN
distribution. Most samples lie near equality, indicating that memory tracks the
target probability rather than merely selecting the same token.
Table~\ref{tab:knn-distribution-summary} confirms this agreement across 1,024
samples. The modes match in 86.62\% of cases, with Pearson \(r=0.9174\), mean KL
0.1820, and mean TV 0.0823. Together, these metrics show alignment in both token
choice and probability allocation. Additional results are provided in
Appendix~\ref{sec:appendix_knn_distribution}. In the protein example in
Figure~\ref{fig:knn-distribution-example}, both distributions select
\emph{task}, with KL 0.0013 and TV 0.0198. Overall, parametric memory learns the
structure of the retrieval distribution rather than only its argmax. Residual
errors arise mainly for diffuse targets, suggesting that matching uncertainty
is harder than identifying the mode. The learned signal can be
interpolated with the frozen base distribution at inference without an online
datastore lookup.

\section{Related Work}
\label{sec:related_work}

\paragraph{Memory-augmented pretraining.}
Most work on language model memory focuses on augmentation at inference time,
whereas few methods incorporate memory into pretraining.
RETRO \citep{borgeaud2022improving} trains an autoregressive model to condition on
chunks retrieved from a database containing trillions of tokens through a
retrieval encoder and cross-attention.
TRIME \citep{zhong2022training} trains with accessible in-batch examples, enabling
the model to adapt to local, long-term, and external memories at inference time.
Limited Memory Language Models (LMLMs) \citep{zhao2025pre} annotate the
pretraining corpus with database lookups and mask retrieved factual values from
the loss, encouraging factual knowledge to remain external to model parameters.
These methods incorporate memory during pretraining but continue to use
non-parametric memory at inference time. RETRO retrieves from an external
database, TRIME accesses non-parametric testing memories and uses nearest
neighbor search when the external memory is large, and LMLMs query an external
database for factual values. In contrast, our work retains the memory
signal learned during pretraining by distilling it into a standalone parametric
module. The resulting memory requires neither external memory construction nor
online retrieval at inference time.

\paragraph{Short-term memory.}
One line of work improves the ability of language models to retain and access
information over long input contexts. Transformer-XL
\citep{dai2019transformer} combines segment-level recurrence with relative
positional encoding, reusing hidden states from previous segments to capture
dependencies beyond a fixed-length context. StreamingLLM
\citep{xiao2024efficient} retains the KV states of initial tokens that act as
attention sinks along with those of the most recent tokens, enabling stable
streaming inference with a fixed cache size. InfLLM \citep{xiao2024infllm}
stores distant context as memory units and retrieves the units relevant to the
current query for attention, enabling a model to process longer contexts without
additional training. LongRoPE
\citep{ding2024longrope} uses non-uniform RoPE rescaling and progressive
positional interpolation to extend the context window while preserving
performance on short inputs. These studies focus primarily on memory within the
input context and thus align with short-term memory. Our work instead pretrains
a standalone parametric long-term memory on a large pretraining corpus.

\paragraph{Long-term memory.}
Long-term memory can be implemented non-parametrically or parametrically.
Non-parametric memory stores knowledge in an external datastore and retrieves
it at inference time. $k$NN-LM
\citep{khandelwal2019generalization} retrieves training contexts most similar to the
current context from a datastore, constructs a distribution over their next
tokens, and interpolates it with the model output distribution.
RAG \citep{lewis2020retrieval} retrieves passages and conditions generation on
the retrieved evidence. Maintaining these datastores incurs substantial storage
overhead, while querying them through nearest-neighbor search adds inference
latency. RAG further increases computation by processing retrieved passages as
additional context. Parametric memory instead encodes knowledge in learned
weights. Memory Decoder
\citep{cao2026memory} and MLP Memory \citep{wei2025mlp} distill
retrieval distributions into plug-and-play neural memories, while Titans
\citep{behrouz2024titans} learns to write and read memory at test time. We build
on this parametric direction but shift its target from memory learned at small
scale or updated online at test time to long-term memory learned at pretraining
scale. General and domain memories are trained
separately, attached to frozen backbones through the same interface, and
selected at inference time without modifying the base model.

\section{Conclusion}
\label{sec:conclusion}

In this paper, we present \emph{Memory Decoder at Scale}, scaling parametric
memory models up to 6.9B parameters and pretraining them for 300B tokens. To
enable memory pretraining at this scale, we developed a distributed Faiss
pipeline based on embedding compression, index sharding, and parallel search,
which addresses the indexing and search bottlenecks in constructing \(k\)NN
distributions over 207B corpus tokens. Sparse \(k\)NN distribution storage
further reduces space requirements, while distributed streaming loads only the
entries required by each batch, enabling retrieval supervision construction at
the scale of language model pretraining.

Experiments across base model and memory scales reveal a consistent advantage
for pairing small base models with large memory models. Scaling memory can
therefore be more parameter-efficient than scaling the base model alone.
A 6.9B general memory enables Pythia-410M to surpass Pythia-12B with 39\%
fewer total parameters. Across Qwen3 Base models from 0.6B to 14B, 1.7B domain
memories improve the average score across biology, law, and finance by more than
9 points at every scale. Further analyses show that memory remains effective
with few-shot examples, benefits knowledge-intensive tasks most, and continues
to improve with memory size and training budget.

Memory Decoder at Scale shows that long-term memory can be pretrained and
scaled independently rather than remaining entangled with reasoning in a single
parameter set. Overall, our results demonstrate that independently scaling pretrained memory offers a more parameter efficient path to improving language model performance.

%Overall, our results point to independently scalable memory as a promising architectural direction for disentangling long-term memory and reasoning in language models.

\section{Limitations}
\label{sec:limitations}

Although inference requires no external retrieval, constructing the $k$NN target
distributions still incurs indexing and retrieval overhead during memory
pretraining. This offline cost remains an additional preprocessing burden that
grows with corpus scale despite the compressed and sharded construction
pipeline. Future work could replace the fixed interpolation coefficient
\(\alpha\) with an adaptive weighting mechanism that determines the memory
contribution for each input based on the context or model confidence. It could
also explore joint or staged training of the memory and backbone to improve
their coordination while preserving modular deployment. Such training may
improve integration but would increase training cost and could weaken
cross-backbone transfer.

\section*{Acknowledgments}

This work is sponsored by the National Natural Science Foundation of China (NSFC) grant (No. 62576211) and the National Key Research and Development Program of China (No. 2023ZD0121402). It is also the result of a collaborative project on novel language model architectures between Shanghai Jiao Tong University (SJTU) and the Shanghai Artificial Intelligence Laboratory. The computational resources required for pretraining the models were provided by the Shanghai AI Lab. This work is also supported by the Specialized Program on Fundamental Research from Science and Technology Commission of Shanghai Municipality (No. 2025SHZDZX025G09).

% ====================
% BIBLIOGRAPHY
% ====================
\nocite{yu2025vismem,guo2025deepsieve,du2025twinvoice}
\bibliography{references}

% ====================
% APPENDIX
% ====================
\clearpage
\appendix
\clearpage
\phantomsection
\label{sec:appendix}
\begin{center}
    {\color{darkblue}\Large\bfseries Appendix}
\end{center}
\vspace{0.5em}

\section{Training Datasets}
\label{sec:appendix_training_datasets}

Table~\ref{tab:appendix-training-datasets} summarizes the corpora used to train
the memory modules. The general memory uses the deduplicated Pile directly as a
language modeling corpus. For domain memories, we first normalize each
domain dataset into plain continued pretraining text before tokenization.
BioInst is converted from biology instruction data into task-oriented training
text. DISC-Law-SFT is deduplicated and formatted as legal domain instruction
text, preserving supporting legal materials when available. FinTrain-unsup is
drawn from the unsupervised financial text subset and chunked as plain CPT data.
We append the Qwen3 EOS token to each training example and pack the resulting
corpus with a block size of 4096 tokens.

\begin{table}[H]
\centering
\caption{Training datasets used for memory pretraining.}
\label{tab:appendix-training-datasets}
\footnotesize
\setlength{\tabcolsep}{8pt}
\renewcommand{\arraystretch}{1.12}
\begin{tabular}{@{}lllr@{}}
\toprule
\textbf{Memory} & \textbf{Dataset} & \textbf{Source} & \textbf{Tokens} \\
\midrule
General &
The Pile deduplicated &
\href{https://huggingface.co/datasets/EleutherAI/the_pile_deduplicated}{EleutherAI/the\_pile\_deduplicated} &
207B \\
\midrule
Domain &
BioInst &
\href{https://github.com/hhnqqq/Biology-Instructions}{hhnqqq/Biology-Instructions} &
552M \\
Domain &
DISC-Law-SFT &
\href{https://huggingface.co/datasets/ShengbinYue/DISC-Law-SFT}{ShengbinYue/DISC-Law-SFT} &
125M \\
Domain &
FinTrain-unsup &
\href{https://huggingface.co/datasets/Salesforce/FinTrain}{Salesforce/FinTrain} &
4.4B \\
\bottomrule
\end{tabular}
\end{table}

\section{Implementation Details}
\label{sec:appendix_implementation_details}

All memory modules are trained separately from the frozen base backbone and are
combined with the backbone distribution only at inference time. The general
memory runs use GPT-NeoX configurations from the Pythia family and scale the
memory from 1.4B to 6.9B parameters. We report these settings in
Table~\ref{tab:appendix-general-implementation}. For domain memories, we
initialize from Qwen3-1.7B-Base and list the shared training hyperparameters in
Table~\ref{tab:appendix-specialist-training-settings}.

\begingroup
\small
\setlength{\tabcolsep}{5pt}
\renewcommand{\arraystretch}{1.03}
\setlength{\LTleft}{0pt plus 1fill}
\setlength{\LTright}{0pt plus 1fill}
\newcommand{\continuedtablehead}[1]{%
\multicolumn{4}{@{}l}{%
\hyperref[#1]{\textcolor{darkblue}{\textbf{\underline{Table~\ref*{#1}}}}}%
\textbf{ continued}}\\[-0.15em]}
\begin{longtable}{@{}lccc@{}}
\caption{Implementation details for general memories.}
\label{tab:appendix-general-implementation}\\
\toprule
\textbf{Configuration / Hyperparameter} &
\textbf{Memdec-1.4B} & \textbf{Memdec-2.8B} & \textbf{Memdec-6.9B} \\
\midrule
\endfirsthead
\continuedtablehead{tab:appendix-general-implementation}
\toprule
\textbf{Configuration / Hyperparameter} &
\textbf{Memdec-1.4B} & \textbf{Memdec-2.8B} & \textbf{Memdec-6.9B} \\
\midrule
\endhead
\midrule
\multicolumn{4}{r}{\textit{Continued on next page}}\\
\endfoot
\bottomrule
\endlastfoot
\multicolumn{4}{@{}l}{\textit{Model configuration}} \\
\midrule
Architecture & GPT-NeoX & GPT-NeoX & GPT-NeoX \\
Hidden size & 2048 & 2560 & 4096 \\
Intermediate size & 8192 & 10240 & 16384 \\
Number of hidden layers & 24 & 32 & 32 \\
Number of attention heads & 16 & 32 & 32 \\
Activation function & GELU & GELU & GELU \\
Vocabulary size & 50,277 & 50,304 & 50,432 \\
Maximum sequence length & 2048 & 2048 & 2048 \\
Maximum position embeddings & 2048 & 2048 & 2048 \\
BOS / EOS token id & 0 / 0 & 0 / 0 & 0 / 0 \\
Model dtype & float32 & bfloat16 & bfloat16 \\
\pagebreak
\midrule
\multicolumn{4}{@{}l}{\textit{Training hyperparameters}} \\
\midrule
Training data & The Pile deduplicated & The Pile deduplicated & The Pile deduplicated \\
Training tokens & 300B & 300B & 300B \\
Training steps & 148,000 & 148,000 & 148,000 \\
Global batch size & 2M tokens & 2M tokens & 2M tokens \\
Sequence length & 2048 & 2048 & 2048 \\
Optimizer & AdamW & AdamW & AdamW \\
Learning rate & \(3\times10^{-4}\) & \(2.5\times10^{-4}\) & \(2\times10^{-4}\) \\
Learning rate schedule & Cosine decay & Cosine decay & Cosine decay \\
Minimum learning rate & \(3\times10^{-5}\) & \(2.5\times10^{-5}\) & \(2\times10^{-5}\) \\
Warmup steps & 2,000 & 2,000 & 2,000 \\
Adam \(\beta_1\) & 0.9 & 0.9 & 0.9 \\
Adam \(\beta_2\) & 0.95 & 0.95 & 0.95 \\
Weight decay & 0.01 & 0.01 & 0.01 \\
Gradient clipping & 1.0 & 1.0 & 1.0 \\
Memory loss weight & 0.5 & 0.5 & 0.5 \\
Training precision & bf16 & bf16 & bf16 \\
Number of GPUs & 64 & 64 & 256 \\
\end{longtable}

\begin{longtable}{@{}>{\raggedright\arraybackslash}p{0.34\linewidth}
>{\centering\arraybackslash}p{0.17\linewidth}
>{\centering\arraybackslash}p{0.20\linewidth}
>{\centering\arraybackslash}p{0.17\linewidth}@{}}
\caption{Training hyperparameters for domain memories.}
\label{tab:appendix-specialist-training-settings}\\
\toprule
\textbf{Configuration / Hyperparameter} &
\textbf{BioInst} & \textbf{DISC-Law-SFT} & \textbf{FinTrain-unsup} \\
\midrule
\multicolumn{4}{@{}l}{\textit{Training hyperparameters}} \\
\midrule
Memory initialization & Qwen3-1.7B-Base & Qwen3-1.7B-Base & Qwen3-1.7B-Base \\
Sequence length & 4096 & 4096 & 4096 \\
Global batch size & 0.5M tokens & 0.5M tokens & 0.25M tokens \\
Optimizer & AdamW & AdamW & AdamW \\
Learning rate & \(3\times10^{-4}\) & \(3\times10^{-4}\) & \(3\times10^{-4}\) \\
Learning rate schedule & Cosine decay & Cosine decay & Cosine decay \\
Minimum learning rate & \(3\times10^{-5}\) & \(3\times10^{-5}\) & \(3\times10^{-5}\) \\
Warmup steps & 1,000 & 1,000 & 200 \\
Adam \(\beta_1\) & 0.9 & 0.9 & 0.9 \\
Adam \(\beta_2\) & 0.95 & 0.95 & 0.95 \\
Weight decay & 0.01 & 0.01 & 0.01 \\
Gradient clipping & 1.0 & 1.0 & 1.0 \\
Memory loss weight & 0.5 & 0.5 & 0.5 \\
Training precision & bf16 & bf16 & bf16 \\
\bottomrule
\end{longtable}
\endgroup

\clearpage

\section{Evaluation Details}
\label{sec:appendix_evaluation_details}

\subsection{General Evaluation}
\label{sec:appendix_general_evaluation}

We use the LM Evaluation
Harness\footnote{\label{fn:lm-evaluation-harness}%
\href{https://github.com/EleutherAI/lm-evaluation-harness}
{EleutherAI/lm-evaluation-harness}} and its accompanying evaluators for the
knowledge tasks. The primary results in
Table~\ref{tab:general-memory-main-scale} follow the prompt setting defined for
each benchmark. The rows receive no dynamically sampled examples.
In Section~\ref{sec:analysis_fewshot_knowledge}, all three AVG bars use the same
13 tasks listed in the main text. Bamboogle retains zero-shot prompting.
PopQA, TruthfulQA-MC, HaluEval, and GPQA-main are omitted from all three
averages. TruthfulQA retains its canonical prefix of six
question and answer pairs as part of the task definition rather than as sampled
context.

The reported \texttt{acc} tasks are ARC-Easy, ARC-Challenge, LAMBADA-OpenAI,
LogiQA, MMLU, PIQA, SciQ, WinoGrande, PopQA, and GPQA-main. NQ-Open, TriviaQA,
2WikiMultiHopQA, Bamboogle, and HotpotQA use exact match. We report
TruthfulQA-MC as the mean of MC1 and MC2. For HaluEval, a fixed seed of 42
selects either the hallucinated or reference candidate for each example. The
model then predicts \texttt{Yes} or \texttt{No}, and a response containing both
labels or neither is incorrect. The HaluEval score averages accuracy across QA,
dialogue, and summarization. GPQA-main averages results over ten independently
permuted answer orders.

\subsection{Domain Evaluation}
\label{sec:appendix_domain_evaluation}

BioInst and LawBench use
OpenCompass\footnote{\href{https://github.com/open-compass/opencompass}
{open-compass/opencompass}}, while FinEval uses the LM Evaluation
Harness\textsuperscript{\ref{fn:lm-evaluation-harness}}. All
methods receive plain base model prompts without tokenizer chat templates and
use greedy generation. Prompts, demonstrations, generation budgets, and
evaluators remain fixed across methods. By default, RAG uses the top-5 retrieved
passages as additional context.
Table~\ref{tab:appendix-domain-generation-budgets} summarizes the generation
budget for each task.

\begingroup
\fontsize{7.5pt}{8pt}\selectfont
\setlength{\tabcolsep}{2.5pt}
\renewcommand{\arraystretch}{0.74}
\begin{table}[H]
\centering
\captionsetup{skip=3pt}
\caption{Values of \texttt{max\_new\_tokens} for the domain evaluations.}
\label{tab:appendix-domain-generation-budgets}
\begin{tabularx}{\textwidth}{@{}l >{\raggedright\arraybackslash}X r@{}}
\toprule
Domain & Tasks & Value \\
\midrule
BioInst & All 21 tasks & 256 \\
\addlinespace
LawBench &
\texttt{knowledge\_question\_answering}, \texttt{dispute\_focus\_identification},
\texttt{issue\_topic\_identification}, \texttt{argument\_mining}, \texttt{case\_analysis}
& 256 \\
LawBench &
\texttt{marital\_disputes\_identification}, \texttt{named\_entity\_recognition},
\texttt{event\_detection}, \texttt{fact\_based\_article\_prediction},
\texttt{prison\_term\_prediction\_wo\_article}, and
\texttt{prison\_term\_prediction\_w\_article} & 384 \\
LawBench & \texttt{document\_proofreading} & 512 \\
LawBench & \texttt{article\_recitation} & 768 \\
LawBench &
\texttt{reading\_comprehension}, \texttt{opinion\_summarization},
\texttt{trigger\_word\_extraction}, \texttt{scene\_based\_article\_prediction},
\texttt{charge\_prediction}, \texttt{criminal\_damages\_calculation}, and
\texttt{consultation} & 1,024 \\
\addlinespace
FinEval & All tasks except the four listed below & 16 \\
FinEval & \texttt{Flare-TATQA} and \texttt{NER} & 128 \\
FinEval & \texttt{ECTSUM} and \texttt{EDTSUM} & 512 \\
\bottomrule
\end{tabularx}
\end{table}
\endgroup

\newpage

\paragraph{BioInst.}
\label{sec:appendix_bioinst_evaluation}
We evaluate the 21 \texttt{biodata\_task\_gen} tasks on each
\texttt{sample\_1k} split with a sequence length of 16,384. BioInst combines
heterogeneous benchmark metrics. We multiply each task metric by 100 and report the
unweighted mean over the 21 tasks. BioInst often expects structured answers
specific to each task, which can cause a
semantically correct base model prediction to be rejected solely for its
format. We therefore apply a deterministic relaxed answer extractor uniformly
to all methods before computing each benchmark metric. The references and
metric definitions remain unchanged, and no model judge is used.

Extraction starts after the final \texttt{</think>} tag when present. It then
considers the final complete \texttt{\textbackslash boxed\{...\}} span, the
text following the last explicit answer cue, and the final 500 characters in
that order. For binary tasks, the extractor recognizes common label pairs and
unambiguous natural language evidence. For five DNA classification tasks, the
extractor gives precedence to conservative negative cues when positive and
negative evidence coexist. For Protein-Solubility, it uses a dedicated soluble
versus insoluble parser. For numeric tasks, the extractor accepts signed values,
scientific notation, and percentages, with
percentages mapped to $[0,1]$ when appropriate. For structured tasks, the
extractor parses dictionaries or key and value expressions, including aliases for
DNA enhancer activity and explicit \texttt{OFF}, \texttt{ON}, and
\texttt{ON\_OFF} fields for RNA switches. Categorical labels are matched
without case or separator sensitivity, and EC numbers are normalized before
computing $F_{\max}$. A fixed fallback is used for each task family when no answer can
be extracted, with the same fallback applied to every model and example.

\paragraph{LawBench.}
\label{sec:appendix_lawbench_evaluation}
The primary evaluation covers all 20 tasks without demonstrations, using 500
examples per task. OpenCompass concatenates \texttt{instruction}, a newline,
and \texttt{question} without retrieving additional examples. We report the
unweighted mean across tasks. Results with one demonstration are included only
as supplementary evidence and are aggregated separately. In this one-shot
setting, the demonstration is embedded directly in \texttt{instruction}, and
the generation budgets remain unchanged. The seven tasks assigned 1,024 tokens
retain this longer limit because shorter budgets can truncate the response
before a scorable final answer is produced.

\paragraph{FinEval.}
\label{sec:appendix_fineval_evaluation}
We evaluate the 25 \texttt{fineval\_em} tasks on their test splits without
demonstrations. Each task uses \texttt{generate\_until} with greedy decoding and
a context length of 32,768. We report the unweighted mean over 20 exact match
tasks, two MCC tasks (CRA-CCF and CRA-Polish), and three ROUGE-1 tasks (ECTSUM,
EDTSUM, and NER). The deterministic evaluator extracts explicit answer labels
when available and otherwise uses the final answer marker or final nonempty
line. Nonnumeric answers are normalized for case, punctuation, and whitespace.
Numeric answers are compared after removing thousands separators, common
currency symbols, and a trailing percent sign, using an absolute tolerance of
$10^{-6}$. Unrecognized predictions on the two MCC tasks count as errors rather
than being discarded. The ROUGE tasks collapse whitespace in the full generated
continuation before computing stemmed ROUGE-1 F1.

\clearpage

\section{Additional General Memory Results}
\label{sec:appendix_general_memory_details}

\begin{figure}[H]
    \centering
    \includegraphics[width=\linewidth]{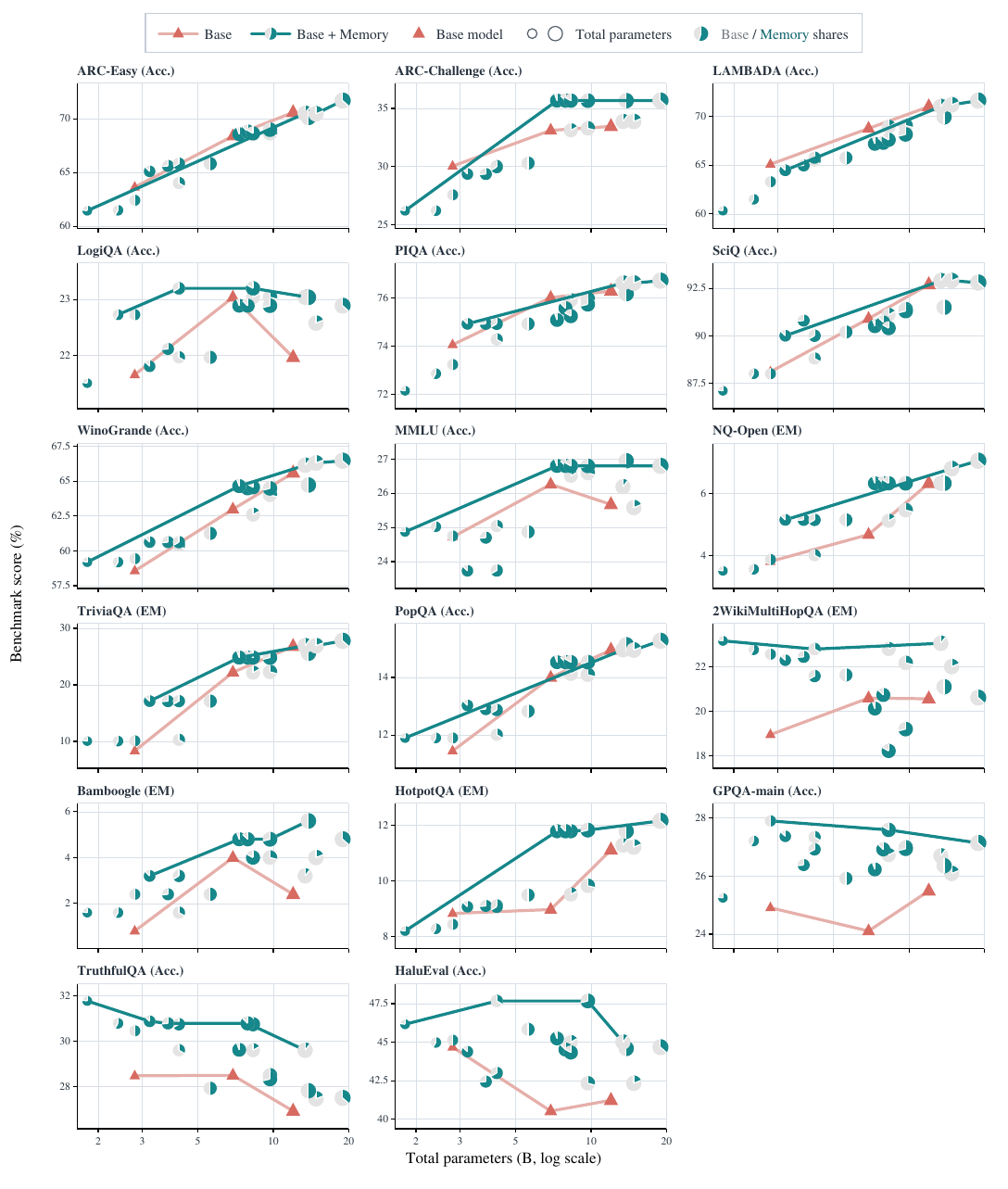}
    \caption{General memory transfer by task under matched parameter counts and
    training budgets. Marker area denotes total parameters. Triangles denote
    base models, while pie markers show the Base and Memory parameter shares.
    Lines trace Base and Base + Memory scaling.}
    \label{fig:appendix-general-memory-task-routes}
\end{figure}

Figure~\ref{fig:appendix-general-memory-task-routes} reports results for
individual tasks under the same total parameter view used in the main analysis.

\clearpage

\section{Additional Domain Memory Results}
\label{sec:appendix_specialist_memory_details}

\subsection{Results with 0.6B Domain Memory}
\label{sec:appendix_specialist_memory_domain_mem06}

Table~\ref{tab:specialist-memory-domain-mem06} reports the 0.6B domain
memory under the zero-shot domain setup used in the main results.
The average score improves for every Qwen3 backbone and both OLMo backbones.
This result shows that the gains are not limited to the larger 1.7B memory. The
OLMo rows use \(20\%\) of the memory training budget in
Table~\ref{tab:specialist-memory-qwen}.

\begin{table}[H]
\centering
\renewcommand{\arraystretch}{0.95}
\caption{Results for the 0.6B domain memory under the zero-shot
setting. OLMo memory rows use \(20\%\) of the training budget in
Table~\ref{tab:specialist-memory-qwen}. Metrics follow
Tables~\ref{tab:specialist-memory-qwen} and~\ref{tab:specialist-memory-olmo}.}
\label{tab:specialist-memory-domain-mem06}
\begin{tabular}{lcccc}
\spdomainhead
\spfamilyrow{Qwen3 Family}
\textbf{Qwen3-0.6B-Base}    & 5.78 & 17.87 & 23.31 & 15.65 \\
\textit{+CPT}               & 7.94\spgain{2.16} & 20.39\spgain{2.52} & 14.90\spdrop{8.41} & 14.41\avdrop{1.24} \\
\textit{+LoRA}              & 10.36\spgain{4.58} & 21.10\spgain{3.23} & 19.54\spdrop{3.77} & 17.00\avgain{1.35} \\
\textit{+RAG}               & 2.96\spdrop{2.82} & 17.82\spdrop{0.05} & 21.37\spdrop{1.94} & 14.05\avdrop{1.60} \\
\spmemrow\textit{+Mem-0.6B} & \textbf{16.53}\spgain{10.75} & \textbf{23.34}\spgain{5.47} & \textbf{29.42}\spgain{6.11} & \textbf{23.10}\avgain{7.44} \\
\midrule
\textbf{Qwen3-1.7B-Base}    & 5.39 & 26.86 & 27.26 & 19.84 \\
\textit{+CPT}               & 4.84\spdrop{0.55} & 28.86\spgain{2.00} & 29.02\spgain{1.76} & 20.91\avgain{1.07} \\
\textit{+LoRA}              & 9.01\spgain{3.62} & 28.53\spgain{1.67} & \textbf{38.74}\spgain{11.48} & 25.43\avgain{5.59} \\
\textit{+RAG}               & 8.07\spgain{2.68} & \textbf{29.61}\spgain{2.75} & 36.63\spgain{9.37} & 24.77\avgain{4.93} \\
\spmemrow\textit{+Mem-0.6B} & \textbf{18.82}\spgain{13.43} & 26.86\spgain{0.00} & 34.54\spgain{7.28} & \textbf{26.74}\avgain{6.90} \\
\midrule
\textbf{Qwen3-4B-Base}      & 5.02 & 27.47 & 37.43 & 23.31 \\
\textit{+CPT}               & 9.14\spgain{4.12} & \textbf{36.10}\spgain{8.63} & 30.22\spdrop{7.21} & 25.15\avgain{1.85} \\
\textit{+LoRA}              & 8.06\spgain{3.04} & 33.65\spgain{6.18} & 33.52\spdrop{3.91} & 25.08\avgain{1.77} \\
\textit{+RAG}               & 8.34\spgain{3.32} & 32.62\spgain{5.15} & 39.13\spgain{1.70} & 26.70\avgain{3.39} \\
\spmemrow\textit{+Mem-0.6B} & \textbf{18.73}\spgain{13.71} & 33.41\spgain{5.94} & \textbf{42.06}\spgain{4.63} & \textbf{31.40}\avgain{8.09} \\
\midrule
\textbf{Qwen3-8B-Base}      & 4.82 & 32.67 & 43.51 & 27.00 \\
\textit{+CPT}               & 9.79\spgain{4.97} & 39.24\spgain{6.57} & 40.00\spdrop{3.51} & 29.68\avgain{2.68} \\
\textit{+LoRA}              & 5.32\spgain{0.50} & 33.55\spgain{0.88} & 42.43\spdrop{1.08} & 27.10\avgain{0.10} \\
\textit{+RAG}               & 7.78\spgain{2.96} & \textbf{40.67}\spgain{8.00} & 46.97\spgain{3.46} & 31.81\avgain{4.81} \\
\spmemrow\textit{+Mem-0.6B} & \textbf{18.37}\spgain{13.55} & 39.59\spgain{6.92} & \textbf{47.36}\spgain{3.85} & \textbf{35.11}\avgain{8.11} \\
\midrule
\textbf{Qwen3-14B-Base}     & 4.01 & 35.45 & 44.25 & 27.90 \\
\textit{+RAG}               & 7.70\spgain{3.69} & \textbf{44.38}\spgain{8.93} & 42.81\spdrop{1.44} & 31.63\avgain{3.73} \\
\spmemrow\textit{+Mem-0.6B} & \textbf{18.42}\spgain{14.41} & 39.99\spgain{4.54} & \textbf{47.71}\spgain{3.46} & \textbf{35.37}\avgain{7.47} \\
\midrule
\spfamilyrow{OLMo Family (cross-vocabulary transfer, 20\% training budget)}
\textbf{OLMo-2-7B}          & 3.90 & 6.37 & 48.43 & 19.57 \\
\textit{+CPT}               & 9.51\spgain{5.61} & \textbf{22.95}\spgain{16.58} & 34.61\spdrop{13.82} & 22.36\avgain{2.79} \\
\textit{+LoRA}              & 9.44\spgain{5.54} & 13.21\spgain{6.84} & 32.34\spdrop{16.09} & 18.33\avdrop{1.24} \\
\textit{+RAG}               & 3.70\spdrop{0.20} & 12.11\spgain{5.74} & 40.88\spdrop{7.55} & 18.90\avdrop{0.67} \\
\spmemrow\textit{+Mem-0.6B} & \textbf{9.86}\spgain{5.96} & 16.78\spgain{10.41} & \textbf{50.15}\spgain{1.72} & \textbf{25.60}\avgain{6.03} \\
\midrule
\textbf{OLMo-3-7B}          & 6.39 & 13.19 & 36.43 & 18.67 \\
\textit{+CPT}               & \textbf{13.69}\spgain{7.30} & 20.29\spgain{7.10} & 38.89\spgain{2.46} & 24.29\avgain{5.62} \\
\textit{+LoRA}              & 6.95\spgain{0.56} & 16.40\spgain{3.21} & 28.21\spdrop{8.22} & 17.19\avdrop{1.48} \\
\textit{+RAG}               & 5.95\spdrop{0.44} & 21.85\spgain{8.66} & 33.10\spdrop{3.33} & 20.30\avgain{1.63} \\
\spmemrow\textit{+Mem-0.6B} & 9.28\spgain{2.89} & \textbf{22.34}\spgain{9.15} & \textbf{46.66}\spgain{10.23} & \textbf{26.09}\avgain{7.42} \\
\spdomaintail
\end{tabular}
\end{table}

\subsection{LawBench One-Shot Results}
\label{sec:appendix_specialist_memory_lawbench_oneshot}

Table~\ref{tab:specialist-memory-lawbench-oneshot} reports one-shot LawBench
results, with one demonstration included in each prompt. On Qwen3, +Mem-1.7B
achieves the highest score for all five backbones and outperforms RAG. On OLMo,
CPT achieves the highest score for both backbones, while both memory sizes still
improve over the frozen base using \(20\%\) of the memory training budget.

\providecommand{\splawfamilyrow}[1]{\rowcolor{gray!8}\multicolumn{7}{c}{\textit{#1}}\\}

\begin{table}[H]
\centering
\caption{One-shot LawBench results using the generation budgets of the primary
evaluation. OLMo memory columns use \(20\%\) of the training budget in
Table~\ref{tab:specialist-memory-qwen}. Bold marks the best score for each
backbone, and underlining marks the second best.}
\label{tab:specialist-memory-lawbench-oneshot}
\begin{tabular}{lcccccc}
\toprule
\textbf{Model} & \textbf{Base} & \textbf{CPT} & \textbf{LoRA} & \textbf{RAG} & \textbf{+Mem-0.6B} & \textbf{+Mem-1.7B} \\
\midrule
\splawfamilyrow{Qwen3 Family}
\textbf{Qwen3-0.6B-Base} & 17.19 & 21.09 & 22.37 & 18.79 & \underline{24.87} & \textbf{28.48} \\
\textbf{Qwen3-1.7B-Base} & 29.45 & 29.52 & 25.97 & \underline{33.25} & 32.20 & \textbf{34.58} \\
\textbf{Qwen3-4B-Base}   & 37.22 & 34.23 & 33.81 & \underline{40.77} & 39.56 & \textbf{41.64} \\
\textbf{Qwen3-8B-Base}   & 42.50 & 42.71 & 37.05 & 43.92 & \underline{44.13} & \textbf{45.70} \\
\textbf{Qwen3-14B-Base}  & 43.68 & -- & -- & 46.18 & \underline{46.46} & \textbf{48.68} \\
\midrule
\splawfamilyrow{OLMo Family (cross-vocabulary transfer, 20\% training budget)}
\textbf{OLMo-2-7B}  & 12.56 & \textbf{23.46} & 14.63 & 15.29 & \underline{18.64} & 17.99 \\
\textbf{OLMo-3-7B}  & 16.37 & \textbf{27.21} & 19.32 & \underline{25.92} & 22.33 & 24.07 \\
\bottomrule
\end{tabular}
\end{table}

\subsection{Effect of RAG Retrieval Depth}
\label{sec:appendix_specialist_memory_rag_topk}

Table~\ref{tab:specialist-memory-rag-topk} compares top-3 and top-5 retrieval
for RAG under the same evaluation settings as
Tables~\ref{tab:specialist-memory-qwen} and~\ref{tab:specialist-memory-olmo}.
Top-5 retrieval performs better in 11 of the 21 backbone and domain comparisons,
while top-3 performs better in the remaining 10. Neither retrieval depth
consistently dominates across backbones or domains.

\begin{table}[H]
\centering
\caption{Top-3 versus top-5 retrieval for the RAG baseline. Bold marks the
better retrieval depth for each backbone and metric.}
\label{tab:specialist-memory-rag-topk}
\begingroup
\small
\setlength{\tabcolsep}{6pt}
\renewcommand{\arraystretch}{0.95}
\begin{tabular}{@{}llcccc@{}}
\toprule
\textbf{Model} & \textbf{RAG} & \textbf{BioInst} & \textbf{LawBench} & \textbf{FinEval} & \textbf{Avg} \\
\midrule
\multirow{2}{*}{\textbf{Qwen3-0.6B-Base}}
& top-3 & \textbf{3.89} & \textbf{18.36} & \textbf{24.86} & \textbf{15.70} \\
& top-5 & 2.96 & 17.82 & 21.37 & 14.05 \\
\midrule
\multirow{2}{*}{\textbf{Qwen3-1.7B-Base}}
& top-3 & 7.33 & \textbf{29.81} & 36.60 & 24.58 \\
& top-5 & \textbf{8.07} & 29.61 & \textbf{36.63} & \textbf{24.77} \\
\midrule
\multirow{2}{*}{\textbf{Qwen3-4B-Base}}
& top-3 & 7.60 & \textbf{33.53} & \textbf{40.16} & \textbf{27.10} \\
& top-5 & \textbf{8.34} & 32.62 & 39.13 & 26.70 \\
\midrule
\multirow{2}{*}{\textbf{Qwen3-8B-Base}}
& top-3 & 7.18 & 40.46 & 45.78 & 31.14 \\
& top-5 & \textbf{7.78} & \textbf{40.67} & \textbf{46.97} & \textbf{31.81} \\
\midrule
\multirow{2}{*}{\textbf{Qwen3-14B-Base}}
& top-3 & 7.35 & 43.97 & \textbf{42.86} & 31.39 \\
& top-5 & \textbf{7.70} & \textbf{44.38} & 42.81 & \textbf{31.63} \\
\midrule
\multirow{2}{*}{\textbf{OLMo-2-7B}}
& top-3 & \textbf{4.16} & 11.88 & \textbf{45.91} & \textbf{20.65} \\
& top-5 & 3.70 & \textbf{12.11} & 40.88 & 18.90 \\
\midrule
\multirow{2}{*}{\textbf{OLMo-3-7B}}
& top-3 & 5.45 & \textbf{22.52} & 32.74 & 20.24 \\
& top-5 & \textbf{5.95} & 21.85 & \textbf{33.10} & \textbf{20.30} \\
\bottomrule
\end{tabular}
\endgroup
\end{table}

\clearpage

\section{Sensitivity to the Interpolation Coefficient}
\label{sec:appendix_interpolation_sensitivity}

At inference, we interpolate the output distributions of the frozen backbone
and memory as
\(
p_{\mathrm{final}}=(1-\alpha)p_{\mathrm{base}}
+\alpha p_{\mathrm{mem}}.
\)
When validation data is available, the interpolation coefficient \(\alpha\) is
tuned on the validation split and then fixed for test evaluation.
Table~\ref{tab:appendix-interpolation-coefficients} summarizes the
interpolation coefficients for the results in Table~1 that use a single
task-level coefficient.
We additionally evaluate
\(\alpha\in\{0.00,0.05,\ldots,1.00\}\) on six representative tasks spanning
general and knowledge-intensive benchmarks.

\begin{table}[htbp]
\centering
\caption{Interpolation coefficients for the results in Table~1 that use a
single task-level coefficient. Column headers indicate the parameter scale of
both the backbone and memory.}
\label{tab:appendix-interpolation-coefficients}
\begingroup
\small
\renewcommand{\arraystretch}{0.98}
\begin{tabular*}{0.82\textwidth}{@{\extracolsep{\fill}}lccc@{}}
\toprule
\textbf{Benchmark} & \textbf{1.4B} & \textbf{2.8B} & \textbf{6.9B} \\
\midrule
\rowcolor{gray!8}
\multicolumn{4}{c}{\textit{General Tasks}} \\
ARC-Easy      & 0.30 & 0.60 & 0.45 \\
ARC-Challenge & 0.25 & 0.75 & 1.00 \\
LAMBADA       & 0.50 & 0.60 & 0.45 \\
LogiQA        & 0.15 & 0.00 & 0.00 \\
PIQA          & 0.55 & 1.00 & 0.60 \\
SciQ          & 0.35 & 0.45 & 0.75 \\
WinoGrande    & 0.70 & 0.55 & 0.85 \\
\midrule
\rowcolor{gray!8}
\multicolumn{4}{c}{\textit{Knowledge}} \\
MMLU            & 0.95 & 0.10 & 0.60 \\
NQ-Open         & 0.80 & 1.00 & 1.00 \\
TriviaQA        & 0.90 & 1.00 & 0.80 \\
PopQA           & 1.00 & 0.85 & 0.55 \\
2WikiMultiHopQA & 0.95 & 0.90 & 0.30 \\
Bamboogle       & 0.40 & 0.60 & 0.55 \\
HotpotQA        & 0.80 & 0.30 & 1.00 \\
\midrule
\rowcolor{gray!8}
\multicolumn{4}{c}{\textit{Hallucination}} \\
TruthfulQA & 0.15 & 0.15 & 0.20 \\
\bottomrule
\end{tabular*}
\endgroup
\end{table}

The coefficient distributions in
Table~\ref{tab:appendix-interpolation-coefficients} provide a useful view of how
the memory contribution changes with evaluation requirements. For general
tasks, the coefficients concentrate in a moderate range, with a median of
\(0.55\) and an interquartile range of \(0.35\)--\(0.70\), reflecting substantial
contributions from both distributions. Knowledge-intensive tasks are more
heavily represented at larger coefficients, with a median of \(0.80\) and an
interquartile range of \(0.55\)--\(0.95\), consistent with stronger use of the
memory distribution when stored knowledge is central. The overlap between the
two ranges indicates a gradual change in emphasis rather than a sharp
separation.

Figure~\ref{fig:appendix-interpolation-sensitivity} reports performance gains
relative to the \(\alpha=0\) endpoint from the same sweep. This within-sweep
reference isolates the effect of interpolation from small differences between
separately executed base evaluations. Across the six tasks, 17 of the 18
task--scale curves stay above \(\alpha=0\) for all 20 nonzero coefficients.
WinoGrande, NQ-Open, TriviaQA, PopQA, and HotpotQA exhibit this behavior at all
three scales, showing that their improvements do not rely on a single isolated
coefficient. For ARC-Easy, the 1.4B curve remains positive over the central
portion of the sweep, while both larger configurations remain positive
throughout. Taken together, the curves show that the improvements persist over
broad coefficient intervals rather than arising from sharply localized peaks.

\clearpage

\begingroup
\definecolor{interpolationblue}{HTML}{0072B2}
\definecolor{interpolationamber}{HTML}{E69F00}
\definecolor{interpolationgreen}{HTML}{009E73}
\definecolor{interpolationgrid}{HTML}{E3E8EC}

\pgfplotsset{
  interpolation axis/.style={
    width=0.485\textwidth,
    height=5.40cm,
    xmin=0,
    xmax=1,
    xtick={0,0.25,0.5,0.75,1},
    xticklabels={0,0.25,0.5,0.75,1},
    tick label style={font=\footnotesize, text=black!82},
    label style={font=\small, text=black!88},
    title style={font=\small\bfseries, text=black!90, yshift=0pt},
    axis x line*=bottom,
    axis y line*=left,
    axis line style={draw=black!52, line width=0.55pt},
    tick style={draw=black!48, line width=0.45pt},
    grid=major,
    grid style={draw=interpolationgrid, line width=0.4pt},
    axis on top,
    enlarge y limits={abs=0.18},
    scaled y ticks=false,
  },
  interpolation 1.4B/.style={
    interpolationblue,
    solid,
    line width=1.25pt,
    mark=o,
    mark repeat=5,
    mark size=1.65pt,
    mark options={solid,fill=white,line width=0.85pt},
  },
  interpolation 2.8B/.style={
    interpolationamber,
    dash pattern=on 5pt off 2.1pt,
    line width=1.25pt,
    mark=square,
    mark repeat=5,
    mark size=1.55pt,
    mark options={solid,fill=white,line width=0.85pt},
  },
  interpolation 6.9B/.style={
    interpolationgreen,
    dash pattern=on 4.5pt off 1.6pt on 1.1pt off 1.6pt,
    line width=1.25pt,
    mark=triangle,
    mark repeat=5,
    mark size=1.80pt,
    mark options={solid,fill=white,line width=0.85pt},
  },
  interpolation zero/.style={black!62,densely dashed,line width=0.8pt},
}

\begin{figure}[t]
\centering
\begin{tikzpicture}
\begin{groupplot}[
  interpolation axis,
  group style={group size=2 by 3, horizontal sep=0.50cm, vertical sep=1.20cm},
  legend style={
    font=\footnotesize,
    draw=black!12,
    fill=black!1,
    rounded corners=1.5pt,
    inner xsep=5pt,
    inner ysep=2.5pt,
    legend columns=3,
    /tikz/every even column/.append style={column sep=11pt},
  },
]
\nextgroupplot[
  title={(a) ARC-Easy},
  ylabel={Performance gain},
  legend to name=interpolationlegend]
\addplot[interpolation 1.4B] table[x=alpha,y=d14] {figure/interpolation_arc_easy.dat};
\addlegendentry{1.4B + Mem-1.4B}
\addplot[interpolation 2.8B] table[x=alpha,y=d28] {figure/interpolation_arc_easy.dat};
\addlegendentry{2.8B + Mem-2.8B}
\addplot[interpolation 6.9B] table[x=alpha,y=d69] {figure/interpolation_arc_easy.dat};
\addlegendentry{6.9B + Mem-6.9B}
\addplot[interpolation zero,forget plot] coordinates {(0,0) (1,0)};

\nextgroupplot[title={(b) WinoGrande}]
\addplot[interpolation 1.4B] table[x=alpha,y=d14] {figure/interpolation_winogrande.dat};
\addplot[interpolation 2.8B] table[x=alpha,y=d28] {figure/interpolation_winogrande.dat};
\addplot[interpolation 6.9B] table[x=alpha,y=d69] {figure/interpolation_winogrande.dat};
\addplot[interpolation zero,forget plot] coordinates {(0,0) (1,0)};

\nextgroupplot[title={(c) NQ-Open}, ylabel={Performance gain}]
\addplot[interpolation 1.4B] table[x=alpha,y=d14] {figure/interpolation_nq_open.dat};
\addplot[interpolation 2.8B] table[x=alpha,y=d28] {figure/interpolation_nq_open.dat};
\addplot[interpolation 6.9B] table[x=alpha,y=d69] {figure/interpolation_nq_open.dat};
\addplot[interpolation zero,forget plot] coordinates {(0,0) (1,0)};

\nextgroupplot[title={(d) TriviaQA}]
\addplot[interpolation 1.4B] table[x=alpha,y=d14] {figure/interpolation_triviaqa.dat};
\addplot[interpolation 2.8B] table[x=alpha,y=d28] {figure/interpolation_triviaqa.dat};
\addplot[interpolation 6.9B] table[x=alpha,y=d69] {figure/interpolation_triviaqa.dat};
\addplot[interpolation zero,forget plot] coordinates {(0,0) (1,0)};

\nextgroupplot[
  title={(e) PopQA},
  xlabel={Interpolation coefficient \(\alpha\)},
  ylabel={Performance gain}]
\addplot[interpolation 1.4B] table[x=alpha,y=d14] {figure/interpolation_popqa.dat};
\addplot[interpolation 2.8B] table[x=alpha,y=d28] {figure/interpolation_popqa.dat};
\addplot[interpolation 6.9B] table[x=alpha,y=d69] {figure/interpolation_popqa.dat};
\addplot[interpolation zero,forget plot] coordinates {(0,0) (1,0)};

\nextgroupplot[
  title={(f) HotpotQA},
  xlabel={Interpolation coefficient \(\alpha\)}]
\addplot[interpolation 1.4B] table[x=alpha,y=d14] {figure/interpolation_hotpotqa.dat};
\addplot[interpolation 2.8B] table[x=alpha,y=d28] {figure/interpolation_hotpotqa.dat};
\addplot[interpolation 6.9B] table[x=alpha,y=d69] {figure/interpolation_hotpotqa.dat};
\addplot[interpolation zero,forget plot] coordinates {(0,0) (1,0)};
\end{groupplot}
\end{tikzpicture}

\vspace{2pt}
\pgfplotslegendfromname{interpolationlegend}
\caption{Sensitivity to the interpolation coefficient across six representative
tasks. Each panel reports the performance gain relative to the within-sweep
\(\alpha=0\) endpoint; positive values indicate an improvement over the
no-memory setting. Lines include all 21 coefficients at intervals of 0.05,
with markers shown every 0.25 for readability. Panels use task-specific
vertical scales.}
\label{fig:appendix-interpolation-sensitivity}
\end{figure}
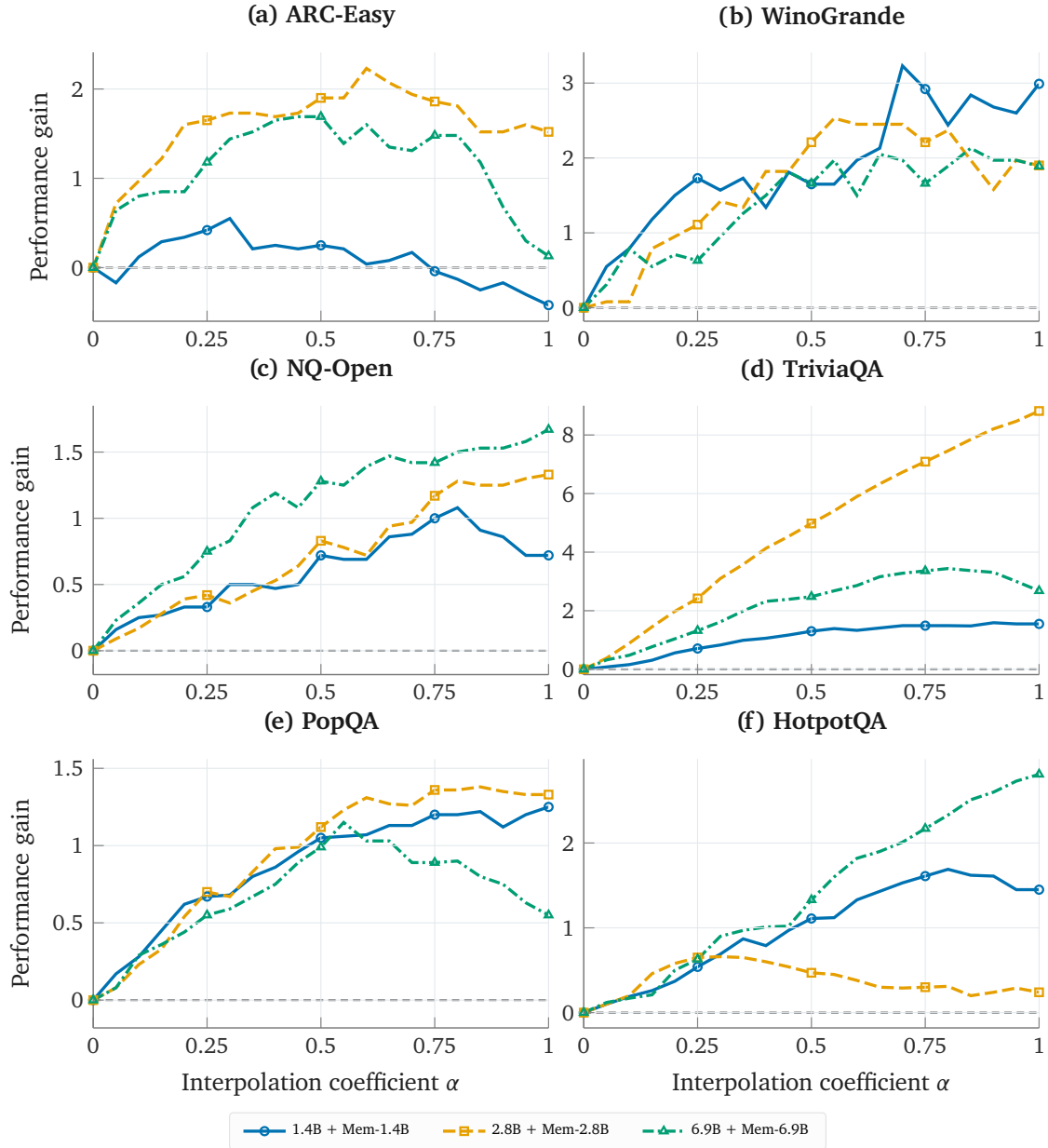
\endgroup

\clearpage

\section{Additional Case Study}
\label{sec:appendix_case_studies}

We provide additional case study examples in
Figures~\ref{fig:case-study-icc-knockout}--\ref{fig:case-study-gimme-shelter}.

\begingroup
\captionsetup{font=small,aboveskip=2pt,belowskip=2pt,hypcap=false}
\setlength{\abovecaptionskip}{2pt}
\setlength{\belowcaptionskip}{2pt}
\begin{center}
    \includegraphics[width=\linewidth]{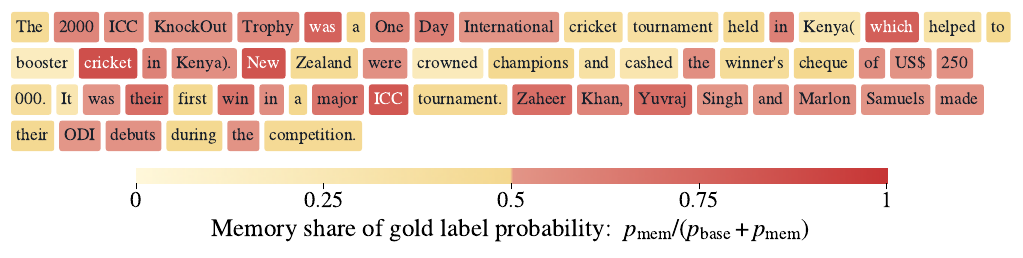}
    \captionof{figure}{Additional case study on a factual question about the
    \textit{2000 ICC KnockOut Trophy}.}
    \label{fig:case-study-icc-knockout}
    \vspace{0.35em}

    \includegraphics[width=\linewidth]{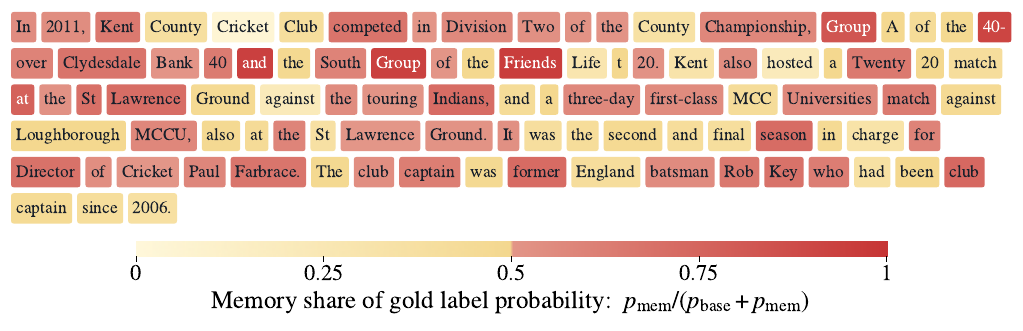}
    \captionof{figure}{Additional case study on a factual question about
    \textit{Kent County Cricket Club in 2011}.}
    \label{fig:case-study-kent-cricket}
    \vspace{0.35em}

    \includegraphics[width=\linewidth]{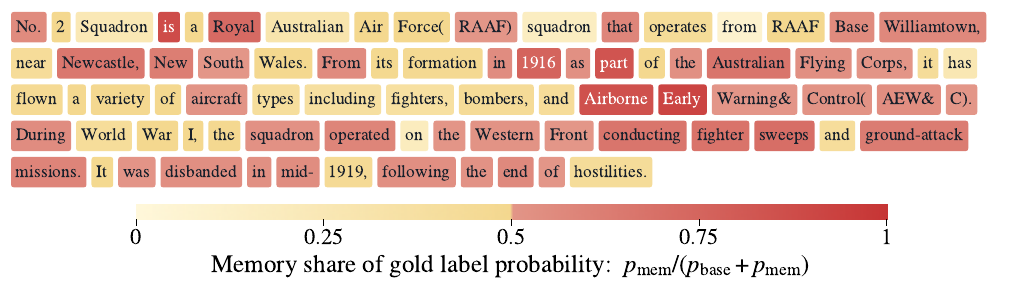}
    \captionof{figure}{Additional case study on a factual question about
    \textit{No. 2 Squadron RAAF}.}
    \label{fig:case-study-raaf}
    \vspace{0.35em}

    \includegraphics[width=\linewidth]{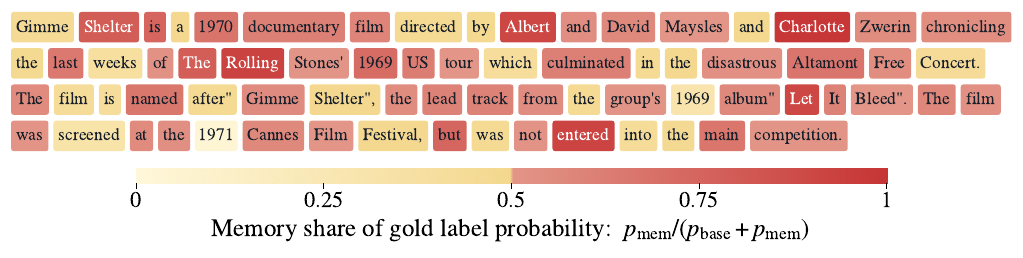}
    \captionof{figure}{Additional case study on a factual question about the
    documentary film \textit{Gimme Shelter}.}
    \label{fig:case-study-gimme-shelter}
\end{center}
\endgroup

\clearpage

\section{Extractable Memorization Evaluation Details}
\label{sec:appendix_extractable_memorization}

This appendix specifies the two BioInst traceability probes summarized in
Section~\ref{sec:extractable_memorization}. Both use examples from the raw
BioInst training corpus and compare the 1.7B memory model with a
Qwen3-1.7B-Base CPT model trained on the same corpus. Thus, the comparison
isolates the memory training mechanism from mere exposure to BioInst data.

\subsection{Metric and Decoding Protocol}

For a training output token sequence, let \(p_i^K\) be an output prefix of
\(K\) tokens and \(s_i^L\) the immediately following suffix of \(L\) tokens. The
model input combines the original task context with \(p_i^K\). The task context
is not counted in \(K\). With deterministic greedy decoding, we define
\[
    \mathrm{EM@}K,L
    = \frac{1}{N}\sum_{i=1}^{N}
      \mathbb{1}\!\left[
      \operatorname{Greedy}_L(M,c_i,p_i^K)=s_i^L
      \right].
\]
Equality is evaluated on token IDs, not normalized strings. We use one
deterministic query per example and do not estimate probabilistic extraction
under repeated stochastic queries \citep{hayes2025measuring}. We
also report \(k\)-eidetic EM, restricting the same rate to suffixes occurring
in at most \(k\) distinct BioInst training rows
\citep{carlini2021extracting}. For paired comparisons, the CPT only and memory
only columns count prompts solved by exactly one model.

\subsection{Verbatim Continuation}

We scan 15,537 \texttt{FunctionEC-FunctionEC} rows. Of these, 10,178 contain
an output prefix of eight tokens followed by a suffix of 16 tokens. Requiring each
selected prefix to have a unique observed continuation leaves 1,112 candidates,
from which we sample \(N=1{,}024\) with seed 41. Each model receives the full
protein and task context together with the eight output tokens, then greedily
generates 16 tokens. The full prompts have a median length of 208.5 tokens, a
mean of 225.4, and a range from 69 to 582.

\begin{table}[htbp]
    \centering
    \small
    \caption{Full results for FunctionEC verbatim continuation.}
    \label{tab:appendix-functionec-em}
    \setlength{\tabcolsep}{5pt}
    \begin{tabular}{@{}lrrrrrrr@{}}
        \toprule
        Metric & \(N\) & \makecell{CPT\\model} & \makecell{memory\\model} & \makecell{CPT\\only} & \makecell{memory\\only} & Both & Neither \\
        \midrule
        \(\mathrm{EM@8,16}\) & 1,024 & 434 (42.4\%) & 509 (49.7\%) & 56 & 131 & 378 & 459 \\
        \bottomrule
    \end{tabular}
\end{table}

Table~\ref{tab:appendix-functionec-rare} stratifies exact recovery by suffix
row frequency. The direction favors the memory model for all three thresholds.
However, these subsets are small, so we treat them as supporting rather than
standalone evidence.

\begin{table}[htbp]
    \centering
    \small
    \caption{FunctionEC continuation for suffixes with limited frequency. A
    suffix is included when it occurs in at most \(k\) BioInst training rows.}
    \label{tab:appendix-functionec-rare}
    \setlength{\tabcolsep}{6pt}
    \begin{tabular}{@{}lrrrrr@{}}
        \toprule
        Subset & \(N\) & \makecell{CPT\\model} & \makecell{memory\\model} & \makecell{CPT\\only} & \makecell{memory\\only} \\
        \midrule
        \(k\leq1\)  & 30  & 6 (20.0\%)  & 11 (36.7\%) & 3 & 8  \\
        \(k\leq5\)  & 76  & 13 (17.1\%) & 21 (27.6\%) & 7 & 15 \\
        \(k\leq10\) & 101 & 16 (15.8\%) & 24 (23.8\%) & 9 & 17 \\
        \bottomrule
    \end{tabular}
\end{table}

The FunctionEC example in Figure~\ref{fig:extractable-memorization-examples}
corresponds to training sample 52,021 (example
\texttt{bioinst\_em\_\allowbreak000616}). Its gold suffix occurs in one training row. The
memory model reproduces all 16 tokens, whereas the CPT model predicts the EC
number but terminates with a shorter continuation.

\subsection{Domain Anchor Verbatim Completion}

We scan 1,215,490 BioInst rows and retain the EMP tasks \texttt{emp-H3},
\texttt{emp-H4}, \texttt{emp-H3K4me1}, and \texttt{emp-H3K9ac}. Candidate
outputs contain a span of four words beginning with the domain anchors
\texttt{DNA}, \texttt{epigenetic}, or \texttt{marks}. The scan yields 10,321
candidates. After removing prompt leakage and capping overly frequent spans,
6,676 remain. We sample 1,024 examples with seed 41 and score whether greedy
generation begins with the exact hidden span. The full prompts have median
length 295 tokens, a mean of 295.6, and a range from 268 to 322. The output
prefix has a median of six words, and every target contains four words.

\begin{table}[htbp]
    \centering
    \small
    \caption{Full EMP domain anchor completion results.}
    \label{tab:appendix-emp-anchor}
    \setlength{\tabcolsep}{4.5pt}
    \begin{tabular}{@{}lrrrr@{}}
        \toprule
        Metric & \makecell{CPT\\model} & \makecell{memory\\model} & \makecell{CPT\\only} & \makecell{memory\\only} \\
        \midrule
        Starts with target span & 231 (22.6\%) & 579 (56.5\%) & 41 & 389 \\
        Target appears anywhere & 369 (36.0\%) & 660 (64.5\%) & 72 & 363 \\
        Generation occurs in source & 242 (23.6\%) & 587 (57.3\%) & 43 & 388 \\
        \bottomrule
    \end{tabular}
\end{table}

The advantage persists after restricting by target span frequency in the
candidate pool. For spans observed at most 50 times, the memory model recovers
194 of 312 targets (62.2\%), compared with 98 (31.4\%) for the CPT model. The
two models exclusively recover 111 and 15 prompts, respectively. The EMP example in
Figure~\ref{fig:extractable-memorization-examples} corresponds to training
sample 1,016,559 from \texttt{emp-H3}. The target span appears 35 times in the
extracted pool. The memory model restores ``epigenetic changes, successfully
confirmed,'' whereas the CPT model begins with the different continuation
``acetylation and methylation nucleosome occupancies.''

\subsection{Additional Qualitative Examples}

Together with Figure~\ref{fig:extractable-memorization-examples}, the eight
examples below complete two qualitative sets of five examples. Each prefix
omits the shared task context, and ellipses abbreviate text beyond the scored
span.

\clearpage
\begin{figure}[H]
    \centering
    \begin{minipage}[c][0.455\textheight][c]{\linewidth}
    \centering
    \begin{minipage}[t]{0.485\linewidth}
        \vspace{0pt}
        \begin{tcolorbox}[
            colback=violet!3,colframe=violet!65!black,boxrule=0.55pt,arc=1.5pt,
            height=40mm,left=4pt,right=4pt,top=3pt,bottom=3pt,
            before skip=0pt,after skip=0pt
        ]
        \scriptsize\raggedright
        \textbf{Domain anchor completion (EMP 1)}\\[-1pt]
        \textcolor{black!60}{Prefix}\quad
        \texttt{\ldots{} Output: \ldots{} our analysis of yeast}\\
        \textcolor{black!60}{Gold}\quad \texttt{DNA did not achieve}\\
        \textcolor{darkred}{\(\times\) CPT model}\quad
        \texttt{did not achieve this.}\\
        \textcolor{darkgreen}{\(\checkmark\) memory model}\quad
        \texttt{DNA did not achieve this.}
        \end{tcolorbox}
    \end{minipage}
    \hfill
    \begin{minipage}[t]{0.485\linewidth}
        \vspace{0pt}
        \begin{tcolorbox}[
            colback=violet!3,colframe=violet!65!black,boxrule=0.55pt,arc=1.5pt,
            height=40mm,left=4pt,right=4pt,top=3pt,bottom=3pt,
            before skip=0pt,after skip=0pt
        ]
        \scriptsize\raggedright
        \textbf{Domain anchor completion (EMP 2)}\\[-1pt]
        \textcolor{black!60}{Prefix}\quad
        \texttt{\ldots{} Output: \ldots{} epigenetic markers, and our yeast}\\
        \textcolor{black!60}{Gold}\quad \texttt{DNA analysis supports these}\\
        \textcolor{darkred}{\(\times\) CPT model}\quad
        \texttt{300 bp DNA sequence supports these predictions.}\\
        \textcolor{darkgreen}{\(\checkmark\) memory model}\quad
        \texttt{DNA analysis supports these predictions.}
        \end{tcolorbox}
    \end{minipage}

    \vspace{3pt}

    \begin{minipage}[t]{0.485\linewidth}
        \vspace{0pt}
        \begin{tcolorbox}[
            colback=violet!3,colframe=violet!65!black,boxrule=0.55pt,arc=1.5pt,
            height=40mm,left=4pt,right=4pt,top=3pt,bottom=3pt,
            before skip=0pt,after skip=0pt
        ]
        \scriptsize\raggedright
        \textbf{Domain anchor completion (EMP 3)}\\[-1pt]
        \textcolor{black!60}{Prefix}\quad
        \texttt{\ldots{} Output: The purpose of the EMP task is to predict}\\
        \textcolor{black!60}{Gold}\quad
        \texttt{epigenetic markers, and our}\\
        \textcolor{darkred}{\(\times\) CPT model}\quad
        \texttt{epigenetic modifications, and our \ldots{}}\\
        \textcolor{darkgreen}{\(\checkmark\) memory model}\quad
        \texttt{epigenetic markers, and our \ldots{}}
        \end{tcolorbox}
    \end{minipage}
    \hfill
    \begin{minipage}[t]{0.485\linewidth}
        \vspace{0pt}
        \begin{tcolorbox}[
            colback=violet!3,colframe=violet!65!black,boxrule=0.55pt,arc=1.5pt,
            height=40mm,left=4pt,right=4pt,top=3pt,bottom=3pt,
            before skip=0pt,after skip=0pt
        ]
        \scriptsize\raggedright
        \textbf{Domain anchor completion (EMP 4)}\\[-1pt]
        \textcolor{black!60}{Prefix}\quad
        \texttt{\ldots{} Output: \ldots{} could not confirm the presence of epigenetic}\\
        \textcolor{black!60}{Gold}\quad \texttt{marks in the DNA.}\\
        \textcolor{darkred}{\(\times\) CPT model}\quad
        \texttt{36me3 marks in the DNA.}\\
        \textcolor{darkgreen}{\(\checkmark\) memory model}\quad
        \texttt{marks in the DNA.}
        \end{tcolorbox}
    \end{minipage}
    \caption{Four additional domain anchor completions solved only by the
    memory model.}
    \label{fig:appendix-emp-examples}
    \end{minipage}
\end{figure}

\begin{figure}[H]
    \centering
    \begin{minipage}[c][0.455\textheight][c]{\linewidth}
    \centering
    \begin{minipage}[t]{0.485\linewidth}
        \vspace{0pt}
        \begin{tcolorbox}[
            colback=blue!3,colframe=sectionblue,boxrule=0.55pt,arc=1.5pt,
            height=40mm,left=4pt,right=4pt,top=3pt,bottom=3pt,
            before skip=0pt,after skip=0pt
        ]
        \scriptsize\raggedright
        \textbf{Verbatim continuation (FunctionEC 1)}\\[-1pt]
        \textcolor{black!60}{Prefix}\quad \texttt{\ldots{} Output: 5.3.1.9,}\\
        \textcolor{black!60}{Gold}\quad
        \texttt{EC5.3.1.- corresponds to the enzyme's function in this sequence}\\
        \textcolor{darkred}{\(\times\) CPT model}\quad
        \texttt{EC5.3.1.-.}\\
        \textcolor{darkgreen}{\(\checkmark\) memory model}\quad
        \texttt{EC5.3.1.- corresponds to the enzyme's function in this sequence}
        \end{tcolorbox}
    \end{minipage}
    \hfill
    \begin{minipage}[t]{0.485\linewidth}
        \vspace{0pt}
        \begin{tcolorbox}[
            colback=blue!3,colframe=sectionblue,boxrule=0.55pt,arc=1.5pt,
            height=40mm,left=4pt,right=4pt,top=3pt,bottom=3pt,
            before skip=0pt,after skip=0pt
        ]
        \scriptsize\raggedright
        \textbf{Verbatim continuation (FunctionEC 2)}\\[-1pt]
        \textcolor{black!60}{Prefix}\quad \texttt{\ldots{} Output: EC2.7.7.2}\\
        \textcolor{black!60}{Gold}\quad
        \texttt{4,EC2.7.7.- identifies the enzyme's function within it}\\
        \textcolor{darkred}{\(\times\) CPT model}\quad
        \texttt{4,EC2.7.7.- is the enzyme's function.}\\
        \textcolor{darkgreen}{\(\checkmark\) memory model}\quad
        \texttt{4,EC2.7.7.- identifies the enzyme's function within it}
        \end{tcolorbox}
    \end{minipage}

    \vspace{3pt}

    \begin{minipage}[t]{0.485\linewidth}
        \vspace{0pt}
        \begin{tcolorbox}[
            colback=blue!3,colframe=sectionblue,boxrule=0.55pt,arc=1.5pt,
            height=40mm,left=4pt,right=4pt,top=3pt,bottom=3pt,
            before skip=0pt,after skip=0pt
        ]
        \scriptsize\raggedright
        \textbf{Verbatim continuation (FunctionEC 3)}\\[-1pt]
        \textcolor{black!60}{Prefix}\quad \texttt{\ldots{} Output: EC3.3.1.1}\\
        \textcolor{black!60}{Gold}\quad
        \texttt{,EC3.3.1.- identifies the enzyme's function within it.}\\
        \textcolor{darkred}{\(\times\) CPT model}\quad
        \texttt{,EC3.3.1.- corresponds to the enzyme's function in this \ldots{}}\\
        \textcolor{darkgreen}{\(\checkmark\) memory model}\quad
        \texttt{,EC3.3.1.- identifies the enzyme's function within it.}
        \end{tcolorbox}
    \end{minipage}
    \hfill
    \begin{minipage}[t]{0.485\linewidth}
        \vspace{0pt}
        \begin{tcolorbox}[
            colback=blue!3,colframe=sectionblue,boxrule=0.55pt,arc=1.5pt,
            height=40mm,left=4pt,right=4pt,top=3pt,bottom=3pt,
            before skip=0pt,after skip=0pt
        ]
        \scriptsize\raggedright
        \textbf{Verbatim continuation (FunctionEC 4)}\\[-1pt]
        \textcolor{black!60}{Prefix}\quad
        \texttt{\ldots{} Output: number EC4.3.3.}\\
        \textcolor{black!60}{Gold}\quad
        \texttt{7,EC4.3.3.- identifies the enzyme's function within it}\\
        \textcolor{darkred}{\(\times\) CPT model}\quad
        \texttt{7,EC4.3.3.-4 indicates the presence of protein modifications}\\
        \textcolor{darkgreen}{\(\checkmark\) memory model}\quad
        \texttt{7,EC4.3.3.- identifies the enzyme's function within it}
        \end{tcolorbox}
    \end{minipage}
    \caption{Four additional FunctionEC continuations solved only by the memory
    model. All target suffixes occur in one BioInst training row.}
    \label{fig:appendix-functionec-examples}
    \end{minipage}
\end{figure}

\clearpage

\section{Additional Details on Memory Fidelity to Retrieval Supervision}
\label{sec:appendix_knn_distribution}

This appendix expands the direct comparison between the retrieval target and
the trained 1.7B BioInst domain memory in
Section~\ref{sec:analysis_knn_distribution}. The aggregate metrics and examples
show how the memory learns its \(k\)NN supervision and contributes this knowledge
to the final prediction through interpolation with a frozen base model.

\subsection{Sampling Protocol}

We uniformly draw 1,024 samples without replacement from the complete BioInst
training datastore. Each sample consists of a context and its corresponding
next token. For each sample, we compare the stored \(k\)NN target \(P_i\) with the
distribution \(Q_i\) produced by the same trained 1.7B BioInst domain memory
used in the main results.

\subsection{Metrics}

Let \(v_i^\star=\operatorname*{arg\,max}_v P_i(v)\). Top token match is the sample
mean of \(\mathbb{1}[v_i^\star=\operatorname*{arg\,max}_v Q_i(v)]\). The two
distribution distances are
\[
    D_{\mathrm{KL}}(P_i\Vert Q_i)
    = \sum_{v\in\mathcal{V}} P_i(v)
      \log\frac{P_i(v)}{Q_i(v)},
    \qquad
    \mathrm{TV}(P_i,Q_i)
    = \frac{1}{2}\sum_{v\in\mathcal{V}}\lvert P_i(v)-Q_i(v)\rvert.
\]
KL emphasizes cases where the memory assigns too little mass to a likely \(k\)NN
token. Total variation is bounded between zero and one and equals the amount
of probability mass that must be reassigned to make the distributions equal.
Finally, Pearson correlation is computed across the 1,024 paired values
\((P_i(v_i^\star),Q_i(v_i^\star))\). Unlike top token match, this statistic tests
whether confidence on the same target token varies consistently across
samples.

\subsection{Target Distribution Structure and Memory Fidelity}

\begin{table}[H]
    \centering
    \small
\setlength{\tabcolsep}{5.5pt}
    \renewcommand{\arraystretch}{1.04}
    \caption{Distribution reproduction for targets supported by one or multiple
    tokens. All rows use the same random set of samples. KL is measured in nats.
    Pearson is reported only for all samples.}
    \label{tab:appendix-knn-distribution-segments}
    \begin{tabular}{@{}lrrrrrr@{}}
        \toprule
        Target type & \(N\) & Share & Top token match & Mean KL & Mean TV & Pearson \(r\) \\
        \midrule
        All samples      & 1,024 & 100.00\% & 86.62\% & 0.1820 & 0.0823 & 0.9174 \\
        Single token     &   599 &  58.50\% & 99.50\% & 0.0206 & 0.0050 & - \\
        Multiple tokens  &   425 &  41.50\% & 68.47\% & 0.4095 & 0.1913 & - \\
        \bottomrule
    \end{tabular}
\end{table}

To examine how fidelity varies with target distribution structure,
Table~\ref{tab:appendix-knn-distribution-segments} separates targets supported
by one token from those that distribute probability across several tokens.
Single token targets reach 99.50\% top token agreement and mean TV 0.0050.

Targets with multiple support tokens are harder, with 68.47\% top token
agreement, mean KL 0.4095, and mean TV 0.1913. Total variation exceeds 0.5 for
63 of the 1,024 samples, corresponding to 6.15\%. Thus, the aggregate metrics
combine near exact reproduction of many concentrated targets with a smaller set
of more ambiguous samples. The strong overall agreement shows that the memory
learns both concentrated targets and targets that divide probability among
several plausible tokens, although the latter remain more difficult.

To assess the stability of the aggregate estimates,
Table~\ref{tab:appendix-knn-distribution-uncertainty} reports 95\% bootstrap
confidence intervals based on 10,000 resamples. The narrow intervals across all
four metrics show that the estimates remain stable under resampling.

\begin{table}[H]
    \centering
    \small
    \setlength{\tabcolsep}{8pt}
    \renewcommand{\arraystretch}{1.04}
    \caption{Bootstrap confidence intervals over evaluation samples, based on
    10,000 resamples.}
    \label{tab:appendix-knn-distribution-uncertainty}
    \begin{tabular}{@{}lrr@{}}
        \toprule
        Metric & Estimate & 95\% interval \\
        \midrule
        Top token match & 86.62\% & [84.47\%, 88.67\%] \\
        Mean KL & 0.1820 & [0.1395, 0.2273] \\
        Mean TV & 0.0823 & [0.0708, 0.0943] \\
        Pearson \(r\) & 0.9174 & [0.8954, 0.9383] \\
        \bottomrule
    \end{tabular}
\end{table}

\subsection{How Memory Contributes to Model Predictions}

The distribution comparison shows how the BioInst memory contributes to the
main results. The \(k\)NN targets encode which continuations are supported by the
BioInst datastore and how probability is divided among plausible tokens. The
high top token agreement, strong probability correlation, and small distances
show that the trained 1.7B memory has absorbed this knowledge into its
parameters. The examples provide the same evidence for individual samples,
where the memory recovers both the preferred token and the relative weights of
its main alternatives.

At inference time, the memory processes the same context as the frozen base
model. Their distributions are linearly interpolated for each next token prediction,
so the retrieval signal learned from BioInst contributes directly to the final
prediction without an online datastore lookup. This mechanism connects
the training objective to the strong gains in the main BioInst results. The
memory signal is mixed directly with the base distribution, allowing
continuations favored by BioInst retrieval to influence each token prediction
while the base model remains frozen. The analysis therefore connects the
retrieval signal learned by the memory to each next token prediction.

\subsection{Additional Examples}

The following cases use the same format as
Figure~\ref{fig:knn-distribution-example}. They complement the protein example
with DNA and RNA samples from the fixed random evaluation set. These
figures are qualitative illustrations. All quantitative conclusions use the
complete random sample.

In the DNA example, \emph{enh} is the target continuation and the mode of both
distributions. Their probability allocations remain close across the displayed
candidates, with KL 0.0020 and total variation 0.0249.

In the RNA example, \emph{does} is the target continuation and is ranked second by
both distributions, while \emph{RNA} is their shared mode. The close
probabilities, KL 0.0019, and total variation 0.0235 show that the trained
memory reproduces the \(k\)NN target. This distinction is why the example
is included here rather than used as the main illustration.

\begin{figure}[H]
    \centering
    \includegraphics[width=0.92\linewidth]{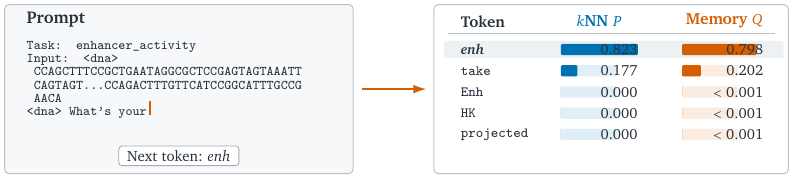}
    \vspace{0.5em}

    \includegraphics[width=0.92\linewidth]{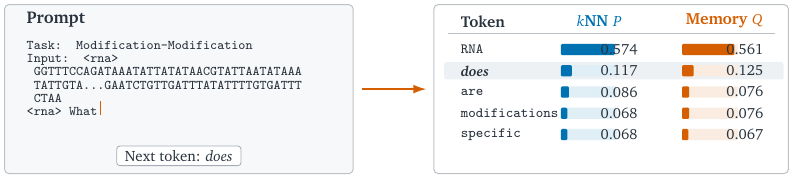}
    \caption{Additional examples in the same format as the main example. The
    DNA continuation \emph{enh} is the shared mode. The RNA continuation
    \emph{does} is ranked second behind \emph{RNA}.}
    \label{fig:appendix-knn-distribution-examples}
\end{figure}

\clearpage

\section{Per-Task Domain Results}
\label{sec:appendix_specialist_memory_task_details}

Tables~\ref{tab:specialist-memory-bioinst-qwen3-0p6b} through
\ref{tab:specialist-memory-fineval-olmo3-7b} expand the domain results from
domain averages to individual tasks. All evaluations use zero-shot prompting,
and RAG retrieves the five highest-ranked passages. We group the tables by
domain, with one compact table for each base backbone. Each table compares Base,
CPT, LoRA, RAG, and the two memory sizes. In the six OLMo tables, both memory
columns use \(20\%\) of the memory training budget in
Table~\ref{tab:specialist-memory-qwen}.

\providecommand{\spdomainmem}[1]{#1}
\providecommand{\spdomainbest}[1]{\textbf{#1}}
\providecommand{\spdomainmembest}[1]{\textbf{#1}}

\subsection{BioInst}

\Needspace{0.42\textheight}
\begingroup
\footnotesize
\setlength{\tabcolsep}{2.6pt}
\renewcommand{\arraystretch}{1.04}
\setlength{\LTpre}{3pt}
\setlength{\LTpost}{6pt}
\begin{longtable}{@{}>{\raggedright\arraybackslash}p{0.365\linewidth}>{\centering\arraybackslash}p{0.085\linewidth}rrrrrr@{}}
\caption{Per-task domain memory results on BioInst for the Qwen3-0.6B-Base backbone.}\label{tab:specialist-memory-bioinst-qwen3-0p6b}\\
\toprule
\textbf{Task} & \textbf{Metric} & \textbf{Base} & \textbf{CPT} & \textbf{LoRA} & \textbf{RAG} & \textbf{+Mem 0.6B} & \textbf{+Mem 1.7B} \\
\midrule
\endfirsthead
\multicolumn{8}{@{}l}{\textbf{\hyperref[tab:specialist-memory-bioinst-qwen3-0p6b]{\textcolor{darkblue}{Table~\ref*{tab:specialist-memory-bioinst-qwen3-0p6b}}} continued}}\\
\toprule
\textbf{Task} & \textbf{Metric} & \textbf{Base} & \textbf{CPT} & \textbf{LoRA} & \textbf{RAG} & \textbf{+Mem 0.6B} & \textbf{+Mem 1.7B} \\
\midrule
\endhead
\bottomrule
\endfoot
DNA cpd & MCC & 2.96 & 21.27 & 32.83 & -0.17 & \spdomainmembest{34.95} & \spdomainmem{31.68} \\
DNA emp & MCC & 1.40 & 7.08 & -4.28 & -3.22 & \spdomainmembest{10.19} & \spdomainmem{7.06} \\
DNA pd & MCC & -2.03 & 10.79 & 18.17 & -3.12 & \spdomainmem{29.96} & \spdomainmembest{40.24} \\
DNA tf h & MCC & 0.54 & 6.71 & -1.07 & -5.92 & \spdomainmem{21.06} & \spdomainmembest{21.45} \\
DNA tf m & MCC & 1.15 & 2.75 & 8.51 & -4.67 & \spdomainmembest{23.89} & \spdomainmem{20.55} \\
Antibody antigen & MCC & 1.21 & 2.47 & -0.37 & -2.83 & \spdomainmembest{4.53} & \spdomainmembest{4.53} \\
Promoter enhancer & MCC & 1.22 & 0.04 & 0.00 & -4.00 & \spdomainmembest{4.54} & \spdomainmem{3.95} \\
RNA protein interaction & MCC & 0.46 & -2.99 & 0.00 & 3.16 & \spdomainmem{4.23} & \spdomainmembest{12.73} \\
\end{longtable}
\endgroup

\Needspace{0.42\textheight}
\begingroup
\footnotesize
\setlength{\tabcolsep}{2.6pt}
\renewcommand{\arraystretch}{1.04}
\setlength{\LTpre}{3pt}
\setlength{\LTpost}{6pt}
\begin{longtable}{@{}>{\raggedright\arraybackslash}p{0.365\linewidth}>{\centering\arraybackslash}p{0.085\linewidth}rrrrrr@{}}
\caption{Per-task domain memory results on BioInst for the Qwen3-1.7B-Base backbone.}\label{tab:specialist-memory-bioinst-qwen3-1p7b}\\
\toprule
\textbf{Task} & \textbf{Metric} & \textbf{Base} & \textbf{CPT} & \textbf{LoRA} & \textbf{RAG} & \textbf{+Mem 0.6B} & \textbf{+Mem 1.7B} \\
\midrule
\endfirsthead
\multicolumn{8}{@{}l}{\textbf{\hyperref[tab:specialist-memory-bioinst-qwen3-1p7b]{\textcolor{darkblue}{Table~\ref*{tab:specialist-memory-bioinst-qwen3-1p7b}}} continued}}\\
\toprule
\textbf{Task} & \textbf{Metric} & \textbf{Base} & \textbf{CPT} & \textbf{LoRA} & \textbf{RAG} & \textbf{+Mem 0.6B} & \textbf{+Mem 1.7B} \\
\midrule
\endhead
\bottomrule
\endfoot
DNA cpd & MCC & 4.09 & 0.00 & 0.00 & -1.90 & \spdomainmembest{35.86} & \spdomainmem{31.04} \\
DNA emp & MCC & 1.26 & 0.00 & 0.00 & 3.92 & \spdomainmembest{10.90} & \spdomainmem{6.91} \\
DNA pd & MCC & -0.77 & 0.00 & 0.00 & 5.11 & \spdomainmem{55.83} & \spdomainmembest{58.31} \\
DNA tf h & MCC & 5.48 & 0.00 & 0.00 & 6.04 & \spdomainmem{22.43} & \spdomainmembest{25.96} \\
DNA tf m & MCC & -2.83 & 0.00 & 0.00 & -2.72 & \spdomainmembest{34.90} & \spdomainmem{34.02} \\
Antibody antigen & MCC & 1.42 & -3.43 & 4.14 & -3.31 & \spdomainmembest{5.26} & \spdomainmem{3.97} \\
Promoter enhancer & MCC & 2.98 & 0.00 & 0.00 & -1.81 & \spdomainmem{5.69} & \spdomainmembest{7.63} \\
RNA protein interaction & MCC & -0.04 & 0.00 & 0.00 & 6.64 & \spdomainmem{0.07} & \spdomainmembest{14.53} \\
\end{longtable}
\endgroup

\Needspace{0.42\textheight}
\begingroup
\footnotesize
\setlength{\tabcolsep}{2.6pt}
\renewcommand{\arraystretch}{1.04}
\setlength{\LTpre}{3pt}
\setlength{\LTpost}{6pt}
\begin{longtable}{@{}>{\raggedright\arraybackslash}p{0.365\linewidth}>{\centering\arraybackslash}p{0.085\linewidth}rrrrrr@{}}
\caption{Per-task domain memory results on BioInst for the Qwen3-4B-Base backbone.}\label{tab:specialist-memory-bioinst-qwen3-4b}\\
\toprule
\textbf{Task} & \textbf{Metric} & \textbf{Base} & \textbf{CPT} & \textbf{LoRA} & \textbf{RAG} & \textbf{+Mem 0.6B} & \textbf{+Mem 1.7B} \\
\midrule
\endfirsthead
\multicolumn{8}{@{}l}{\textbf{\hyperref[tab:specialist-memory-bioinst-qwen3-4b]{\textcolor{darkblue}{Table~\ref*{tab:specialist-memory-bioinst-qwen3-4b}}} continued}}\\
\toprule
\textbf{Task} & \textbf{Metric} & \textbf{Base} & \textbf{CPT} & \textbf{LoRA} & \textbf{RAG} & \textbf{+Mem 0.6B} & \textbf{+Mem 1.7B} \\
\midrule
\endhead
\bottomrule
\endfoot
DNA cpd & MCC & -3.82 & 28.93 & 3.65 & -1.47 & \spdomainmembest{36.80} & \spdomainmem{35.40} \\
DNA emp & MCC & -2.96 & 3.81 & 2.65 & 2.56 & \spdomainmembest{12.03} & \spdomainmem{6.33} \\
DNA pd & MCC & -2.98 & 2.96 & -2.39 & -0.24 & \spdomainmem{49.59} & \spdomainmembest{65.53} \\
DNA tf h & MCC & 1.88 & -0.51 & -1.02 & 0.62 & \spdomainmembest{29.23} & \spdomainmem{26.00} \\
DNA tf m & MCC & -5.77 & -0.55 & -1.01 & -0.34 & \spdomainmembest{34.02} & \spdomainmem{31.67} \\
Antibody antigen & MCC & 4.22 & 4.59 & 2.92 & \spdomainbest{5.70} & \spdomainmem{2.28} & \spdomainmem{3.80} \\
Promoter enhancer & MCC & \spdomainbest{4.02} & -3.16 & 1.67 & 0.10 & \spdomainmem{3.17} & \spdomainmem{3.36} \\
RNA protein interaction & MCC & -2.57 & 0.00 & 0.00 & 12.51 & \spdomainmem{4.74} & \spdomainmembest{14.53} \\
\end{longtable}
\endgroup

\Needspace{0.42\textheight}
\begingroup
\footnotesize
\setlength{\tabcolsep}{2.6pt}
\renewcommand{\arraystretch}{1.04}
\setlength{\LTpre}{3pt}
\setlength{\LTpost}{6pt}
\begin{longtable}{@{}>{\raggedright\arraybackslash}p{0.365\linewidth}>{\centering\arraybackslash}p{0.085\linewidth}rrrrrr@{}}
\caption{Per-task domain memory results on BioInst for the Qwen3-8B-Base backbone.}\label{tab:specialist-memory-bioinst-qwen3-8b}\\
\toprule
\textbf{Task} & \textbf{Metric} & \textbf{Base} & \textbf{CPT} & \textbf{LoRA} & \textbf{RAG} & \textbf{+Mem 0.6B} & \textbf{+Mem 1.7B} \\
\midrule
\endfirsthead
\multicolumn{8}{@{}l}{\textbf{\hyperref[tab:specialist-memory-bioinst-qwen3-8b]{\textcolor{darkblue}{Table~\ref*{tab:specialist-memory-bioinst-qwen3-8b}}} continued}}\\
\toprule
\textbf{Task} & \textbf{Metric} & \textbf{Base} & \textbf{CPT} & \textbf{LoRA} & \textbf{RAG} & \textbf{+Mem 0.6B} & \textbf{+Mem 1.7B} \\
\midrule
\endhead
\bottomrule
\endfoot
DNA cpd & MCC & -2.33 & 17.33 & 6.93 & 3.35 & \spdomainmembest{36.29} & \spdomainmem{29.57} \\
DNA emp & MCC & -0.89 & -0.01 & -2.11 & 2.72 & \spdomainmembest{10.06} & \spdomainmem{7.19} \\
DNA pd & MCC & -7.98 & \spdomainbest{43.17} & 6.06 & 4.87 & \spdomainmem{36.03} & \spdomainmem{38.62} \\
DNA tf h & MCC & -2.27 & 3.91 & 0.00 & 1.82 & \spdomainmembest{27.21} & \spdomainmem{22.19} \\
DNA tf m & MCC & 0.24 & -5.11 & 0.00 & 0.18 & \spdomainmembest{31.41} & \spdomainmem{25.94} \\
Antibody antigen & MCC & 3.16 & \spdomainbest{4.62} & -1.27 & 0.27 & \spdomainmem{4.12} & \spdomainmem{1.38} \\
Promoter enhancer & MCC & 2.83 & 0.00 & 0.00 & -2.29 & \spdomainmembest{7.10} & \spdomainmem{6.32} \\
RNA protein interaction & MCC & 2.01 & 0.00 & 0.00 & 6.36 & \spdomainmem{1.45} & \spdomainmembest{16.26} \\
\end{longtable}
\endgroup

\Needspace{0.42\textheight}
\begingroup
\footnotesize
\setlength{\tabcolsep}{2.6pt}
\renewcommand{\arraystretch}{1.04}
\setlength{\LTpre}{3pt}
\setlength{\LTpost}{6pt}
\begin{longtable}{@{}>{\raggedright\arraybackslash}p{0.365\linewidth}>{\centering\arraybackslash}p{0.085\linewidth}rrrrrr@{}}
\caption{Per-task domain memory results on BioInst for the Qwen3-14B-Base backbone.}\label{tab:specialist-memory-bioinst-qwen3-14b}\\
\toprule
\textbf{Task} & \textbf{Metric} & \textbf{Base} & \textbf{CPT} & \textbf{LoRA} & \textbf{RAG} & \textbf{+Mem 0.6B} & \textbf{+Mem 1.7B} \\
\midrule
\endfirsthead
\multicolumn{8}{@{}l}{\textbf{\hyperref[tab:specialist-memory-bioinst-qwen3-14b]{\textcolor{darkblue}{Table~\ref*{tab:specialist-memory-bioinst-qwen3-14b}}} continued}}\\
\toprule
\textbf{Task} & \textbf{Metric} & \textbf{Base} & \textbf{CPT} & \textbf{LoRA} & \textbf{RAG} & \textbf{+Mem 0.6B} & \textbf{+Mem 1.7B} \\
\midrule
\endhead
\bottomrule
\endfoot
DNA cpd & MCC & 0.09 & -- & -- & 8.45 & \spdomainmembest{35.29} & \spdomainmem{31.63} \\
DNA emp & MCC & -6.01 & -- & -- & -1.72 & \spdomainmembest{10.32} & \spdomainmem{9.78} \\
DNA pd & MCC & 1.72 & -- & -- & -0.27 & \spdomainmem{32.60} & \spdomainmembest{42.82} \\
DNA tf h & MCC & 1.29 & -- & -- & 2.17 & \spdomainmembest{27.32} & \spdomainmem{17.49} \\
DNA tf m & MCC & -3.08 & -- & -- & 4.34 & \spdomainmembest{29.08} & \spdomainmem{19.05} \\
Antibody antigen & MCC & 0.50 & -- & -- & 0.10 & \spdomainmembest{9.27} & \spdomainmem{4.02} \\
Promoter enhancer & MCC & -2.59 & -- & -- & 1.82 & \spdomainmembest{5.55} & \spdomainmem{4.37} \\
RNA protein interaction & MCC & -7.13 & -- & -- & 1.07 & \spdomainmem{4.67} & \spdomainmembest{12.28} \\
\end{longtable}
\endgroup

\Needspace{0.42\textheight}
\begingroup
\footnotesize
\setlength{\tabcolsep}{2.6pt}
\renewcommand{\arraystretch}{1.04}
\setlength{\LTpre}{3pt}
\setlength{\LTpost}{6pt}
\begin{longtable}{@{}>{\raggedright\arraybackslash}p{0.365\linewidth}>{\centering\arraybackslash}p{0.085\linewidth}rrrrrr@{}}
\caption{Per-task domain memory results on BioInst for the OLMo-2-7B backbone.}\label{tab:specialist-memory-bioinst-olmo2-7b}\\
\toprule
\textbf{Task} & \textbf{Metric} & \textbf{Base} & \textbf{CPT} & \textbf{LoRA} & \textbf{RAG} & \textbf{+Mem 0.6B} & \textbf{+Mem 1.7B} \\
\midrule
\endfirsthead
\multicolumn{8}{@{}l}{\textbf{\hyperref[tab:specialist-memory-bioinst-olmo2-7b]{\textcolor{darkblue}{Table~\ref*{tab:specialist-memory-bioinst-olmo2-7b}}} continued}}\\
\toprule
\textbf{Task} & \textbf{Metric} & \textbf{Base} & \textbf{CPT} & \textbf{LoRA} & \textbf{RAG} & \textbf{+Mem 0.6B} & \textbf{+Mem 1.7B} \\
\midrule
\endhead
\bottomrule
\endfoot
DNA cpd & MCC & 0.00 & 13.34 & 0.95 & 0.00 & \spdomainmembest{15.64} & \spdomainmem{5.72} \\
DNA emp & MCC & -6.14 & 5.77 & 5.02 & 0.00 & \spdomainmembest{6.72} & \spdomainmem{4.11} \\
DNA pd & MCC & 0.00 & -5.31 & \spdomainbest{18.49} & 0.00 & \spdomainmem{5.39} & \spdomainmem{8.07} \\
DNA tf h & MCC & 0.00 & -8.91 & -1.84 & -3.11 & \spdomainmembest{10.63} & \spdomainmem{5.96} \\
DNA tf m & MCC & 0.00 & 3.37 & -0.84 & -0.13 & \spdomainmem{5.04} & \spdomainmembest{6.15} \\
Antibody antigen & MCC & 0.00 & 3.75 & -4.38 & 0.00 & \spdomainmembest{4.53} & \spdomainmem{4.31} \\
Promoter enhancer & MCC & 0.00 & 0.00 & -0.64 & -3.16 & \spdomainmem{4.35} & \spdomainmembest{4.49} \\
RNA protein interaction & MCC & 0.00 & -4.22 & 9.06 & 0.00 & \spdomainmem{13.96} & \spdomainmembest{14.82} \\
\end{longtable}
\endgroup

\Needspace{0.42\textheight}
\begingroup
\footnotesize
\setlength{\tabcolsep}{2.6pt}
\renewcommand{\arraystretch}{1.04}
\setlength{\LTpre}{3pt}
\setlength{\LTpost}{6pt}
\begin{longtable}{@{}>{\raggedright\arraybackslash}p{0.365\linewidth}>{\centering\arraybackslash}p{0.085\linewidth}rrrrrr@{}}
\caption{Per-task domain memory results on BioInst for the OLMo-3-7B backbone.}\label{tab:specialist-memory-bioinst-olmo3-7b}\\
\toprule
\textbf{Task} & \textbf{Metric} & \textbf{Base} & \textbf{CPT} & \textbf{LoRA} & \textbf{RAG} & \textbf{+Mem 0.6B} & \textbf{+Mem 1.7B} \\
\midrule
\endfirsthead
\multicolumn{8}{@{}l}{\textbf{\hyperref[tab:specialist-memory-bioinst-olmo3-7b]{\textcolor{darkblue}{Table~\ref*{tab:specialist-memory-bioinst-olmo3-7b}}} continued}}\\
\toprule
\textbf{Task} & \textbf{Metric} & \textbf{Base} & \textbf{CPT} & \textbf{LoRA} & \textbf{RAG} & \textbf{+Mem 0.6B} & \textbf{+Mem 1.7B} \\
\midrule
\endhead
\bottomrule
\endfoot
DNA cpd & MCC & 1.50 & \spdomainbest{47.01} & -4.29 & 3.96 & \spdomainmem{14.90} & \spdomainmem{5.24} \\
DNA emp & MCC & -0.69 & \spdomainbest{5.84} & 0.00 & -0.01 & \spdomainmem{2.78} & \spdomainmem{5.31} \\
DNA pd & MCC & 7.04 & \spdomainbest{20.87} & 0.68 & 0.11 & \spdomainmem{6.30} & \spdomainmem{8.10} \\
DNA tf h & MCC & 2.47 & 3.73 & -5.51 & 0.95 & \spdomainmembest{6.66} & \spdomainmem{5.55} \\
DNA tf m & MCC & 6.11 & -1.17 & \spdomainbest{11.00} & 4.53 & \spdomainmem{5.36} & \spdomainmem{7.25} \\
Antibody antigen & MCC & -1.88 & 0.00 & 2.42 & 2.42 & \spdomainmembest{4.62} & \spdomainmem{2.56} \\
Promoter enhancer & MCC & -0.36 & 0.00 & 0.00 & 1.36 & \spdomainmem{2.67} & \spdomainmembest{4.49} \\
RNA protein interaction & MCC & -0.35 & 0.00 & 3.35 & 2.95 & \spdomainmembest{15.48} & \spdomainmem{4.67} \\
\end{longtable}
\endgroup

\clearpage

\subsection{LawBench}

\Needspace{0.42\textheight}
\begingroup
\footnotesize
\setlength{\tabcolsep}{2.6pt}
\renewcommand{\arraystretch}{1.04}
\setlength{\LTpre}{3pt}
\setlength{\LTpost}{6pt}
\begin{longtable}{@{}>{\raggedright\arraybackslash}p{0.445\linewidth}rrrrrr@{}}
\caption{Per-task domain memory results on LawBench for the Qwen3-0.6B-Base backbone.}\label{tab:specialist-memory-lawbench-qwen3-0p6b}\\
\toprule
\textbf{Task} & \textbf{Base} & \textbf{CPT} & \textbf{LoRA} & \textbf{RAG} & \textbf{+Mem 0.6B} & \textbf{+Mem 1.7B} \\
\midrule
\endfirsthead
\multicolumn{7}{@{}l}{\textbf{\hyperref[tab:specialist-memory-lawbench-qwen3-0p6b]{\textcolor{darkblue}{Table~\ref*{tab:specialist-memory-lawbench-qwen3-0p6b}}} continued}}\\
\toprule
\textbf{Task} & \textbf{Base} & \textbf{CPT} & \textbf{LoRA} & \textbf{RAG} & \textbf{+Mem 0.6B} & \textbf{+Mem 1.7B} \\
\midrule
\endhead
\bottomrule
\endfoot
1.1 Article recitation & 14.72 & 17.40 & \spdomainbest{17.55} & 1.73 & \spdomainmem{7.23} & \spdomainmem{8.54} \\
1.2 Knowledge question answering & 23.40 & 8.00 & 6.40 & \spdomainbest{30.00} & \spdomainmem{23.80} & \spdomainmem{29.60} \\
2.1 Document proofreading & \spdomainbest{2.88} & 1.48 & 1.12 & 2.09 & \spdomainmem{2.69} & \spdomainmem{2.78} \\
2.2 Dispute focus identification & 5.80 & 3.60 & 4.40 & 3.40 & \spdomainmem{5.00} & \spdomainmembest{6.60} \\
2.3 Marital disputes identification & 15.24 & 23.41 & 10.75 & \spdomainbest{38.32} & \spdomainmem{30.10} & \spdomainmem{37.41} \\
2.4 Issue topic identification & 10.60 & \spdomainbest{28.40} & 28.20 & 21.20 & \spdomainmem{17.80} & \spdomainmem{20.60} \\
2.5 Reading comprehension & \spdomainbest{33.86} & 24.96 & 25.00 & 3.96 & \spdomainmem{28.20} & \spdomainmem{29.97} \\
2.6 Named entity recognition & \spdomainbest{18.36} & 2.00 & 2.20 & 6.15 & \spdomainmem{12.72} & \spdomainmem{12.72} \\
\end{longtable}
\endgroup

\Needspace{0.42\textheight}
\begingroup
\footnotesize
\setlength{\tabcolsep}{2.6pt}
\renewcommand{\arraystretch}{1.04}
\setlength{\LTpre}{3pt}
\setlength{\LTpost}{6pt}
\begin{longtable}{@{}>{\raggedright\arraybackslash}p{0.445\linewidth}rrrrrr@{}}
\caption{Per-task domain memory results on LawBench for the Qwen3-1.7B-Base backbone.}\label{tab:specialist-memory-lawbench-qwen3-1p7b}\\
\toprule
\textbf{Task} & \textbf{Base} & \textbf{CPT} & \textbf{LoRA} & \textbf{RAG} & \textbf{+Mem 0.6B} & \textbf{+Mem 1.7B} \\
\midrule
\endfirsthead
\multicolumn{7}{@{}l}{\textbf{\hyperref[tab:specialist-memory-lawbench-qwen3-1p7b]{\textcolor{darkblue}{Table~\ref*{tab:specialist-memory-lawbench-qwen3-1p7b}}} continued}}\\
\toprule
\textbf{Task} & \textbf{Base} & \textbf{CPT} & \textbf{LoRA} & \textbf{RAG} & \textbf{+Mem 0.6B} & \textbf{+Mem 1.7B} \\
\midrule
\endhead
\bottomrule
\endfoot
1.1 Article recitation & 10.37 & 24.38 & \spdomainbest{26.27} & 14.53 & \spdomainmem{17.23} & \spdomainmem{20.76} \\
1.2 Knowledge question answering & 41.40 & 21.40 & 23.40 & \spdomainbest{72.40} & \spdomainmem{27.80} & \spdomainmem{37.00} \\
2.1 Document proofreading & 4.21 & 3.00 & 2.14 & \spdomainbest{4.53} & \spdomainmem{3.80} & \spdomainmem{3.62} \\
2.2 Dispute focus identification & \spdomainbest{15.40} & 5.40 & 5.00 & 8.40 & \spdomainmem{9.80} & \spdomainmem{10.00} \\
2.3 Marital disputes identification & 27.16 & 43.91 & 44.60 & \spdomainbest{46.82} & \spdomainmem{33.49} & \spdomainmem{38.85} \\
2.4 Issue topic identification & 23.20 & \spdomainbest{34.60} & 29.80 & 22.40 & \spdomainmem{22.20} & \spdomainmem{23.20} \\
2.5 Reading comprehension & \spdomainbest{50.82} & 29.75 & 29.71 & 20.03 & \spdomainmem{36.66} & \spdomainmem{42.47} \\
2.6 Named entity recognition & 6.91 & 4.55 & 2.28 & \spdomainbest{6.94} & \spdomainmem{3.77} & \spdomainmem{3.79} \\
\end{longtable}
\endgroup

\Needspace{0.42\textheight}
\begingroup
\footnotesize
\setlength{\tabcolsep}{2.6pt}
\renewcommand{\arraystretch}{1.04}
\setlength{\LTpre}{3pt}
\setlength{\LTpost}{6pt}
\begin{longtable}{@{}>{\raggedright\arraybackslash}p{0.445\linewidth}rrrrrr@{}}
\caption{Per-task domain memory results on LawBench for the Qwen3-4B-Base backbone.}\label{tab:specialist-memory-lawbench-qwen3-4b}\\
\toprule
\textbf{Task} & \textbf{Base} & \textbf{CPT} & \textbf{LoRA} & \textbf{RAG} & \textbf{+Mem 0.6B} & \textbf{+Mem 1.7B} \\
\midrule
\endfirsthead
\multicolumn{7}{@{}l}{\textbf{\hyperref[tab:specialist-memory-lawbench-qwen3-4b]{\textcolor{darkblue}{Table~\ref*{tab:specialist-memory-lawbench-qwen3-4b}}} continued}}\\
\toprule
\textbf{Task} & \textbf{Base} & \textbf{CPT} & \textbf{LoRA} & \textbf{RAG} & \textbf{+Mem 0.6B} & \textbf{+Mem 1.7B} \\
\midrule
\endhead
\bottomrule
\endfoot
1.1 Article recitation & 20.11 & 10.83 & 23.83 & \spdomainbest{25.47} & \spdomainmem{19.20} & \spdomainmem{21.78} \\
1.2 Knowledge question answering & 21.60 & 28.00 & 32.00 & \spdomainbest{83.20} & \spdomainmem{52.40} & \spdomainmem{62.60} \\
2.1 Document proofreading & \spdomainbest{6.01} & 5.08 & 3.79 & 3.54 & \spdomainmem{5.87} & \spdomainmem{5.88} \\
2.2 Dispute focus identification & \spdomainbest{20.20} & 4.20 & 9.20 & 14.20 & \spdomainmem{9.00} & \spdomainmem{12.60} \\
2.3 Marital disputes identification & 27.12 & \spdomainbest{53.55} & 26.28 & 52.03 & \spdomainmem{33.41} & \spdomainmem{40.44} \\
2.4 Issue topic identification & 31.40 & \spdomainbest{33.60} & 27.40 & 4.80 & \spdomainmem{31.40} & \spdomainmem{32.00} \\
2.5 Reading comprehension & 3.20 & \spdomainbest{30.25} & 30.18 & 21.73 & \spdomainmem{18.60} & \spdomainmem{22.10} \\
2.6 Named entity recognition & 10.81 & 21.73 & \spdomainbest{32.40} & 18.96 & \spdomainmem{9.69} & \spdomainmem{10.65} \\
\end{longtable}
\endgroup

\Needspace{0.42\textheight}
\begingroup
\footnotesize
\setlength{\tabcolsep}{2.6pt}
\renewcommand{\arraystretch}{1.04}
\setlength{\LTpre}{3pt}
\setlength{\LTpost}{6pt}
\begin{longtable}{@{}>{\raggedright\arraybackslash}p{0.445\linewidth}rrrrrr@{}}
\caption{Per-task domain memory results on LawBench for the Qwen3-8B-Base backbone.}\label{tab:specialist-memory-lawbench-qwen3-8b}\\
\toprule
\textbf{Task} & \textbf{Base} & \textbf{CPT} & \textbf{LoRA} & \textbf{RAG} & \textbf{+Mem 0.6B} & \textbf{+Mem 1.7B} \\
\midrule
\endfirsthead
\multicolumn{7}{@{}l}{\textbf{\hyperref[tab:specialist-memory-lawbench-qwen3-8b]{\textcolor{darkblue}{Table~\ref*{tab:specialist-memory-lawbench-qwen3-8b}}} continued}}\\
\toprule
\textbf{Task} & \textbf{Base} & \textbf{CPT} & \textbf{LoRA} & \textbf{RAG} & \textbf{+Mem 0.6B} & \textbf{+Mem 1.7B} \\
\midrule
\endhead
\bottomrule
\endfoot
1.1 Article recitation & 24.34 & \spdomainbest{48.07} & 42.17 & 27.46 & \spdomainmem{25.82} & \spdomainmem{24.91} \\
1.2 Knowledge question answering & 37.20 & 31.40 & 38.40 & \spdomainbest{92.00} & \spdomainmem{66.20} & \spdomainmem{72.80} \\
2.1 Document proofreading & \spdomainbest{8.54} & 6.31 & 5.47 & 6.33 & \spdomainmem{7.82} & \spdomainmem{8.14} \\
2.2 Dispute focus identification & 13.40 & 10.60 & 5.20 & 13.80 & \spdomainmem{12.80} & \spdomainmembest{14.00} \\
2.3 Marital disputes identification & 25.97 & 60.11 & 33.16 & \spdomainbest{63.28} & \spdomainmem{36.77} & \spdomainmem{41.08} \\
2.4 Issue topic identification & \spdomainbest{32.40} & 31.40 & 30.20 & 20.60 & \spdomainmem{31.60} & \spdomainmem{31.60} \\
2.5 Reading comprehension & 17.53 & 33.81 & \spdomainbest{35.84} & 20.10 & \spdomainmem{29.36} & \spdomainmem{30.72} \\
2.6 Named entity recognition & 42.87 & 13.04 & 30.79 & 20.92 & \spdomainmembest{46.00} & \spdomainmem{45.20} \\
\end{longtable}
\endgroup

\Needspace{0.42\textheight}
\begingroup
\footnotesize
\setlength{\tabcolsep}{2.6pt}
\renewcommand{\arraystretch}{1.04}
\setlength{\LTpre}{3pt}
\setlength{\LTpost}{6pt}
\begin{longtable}{@{}>{\raggedright\arraybackslash}p{0.445\linewidth}rrrrrr@{}}
\caption{Per-task domain memory results on LawBench for the Qwen3-14B-Base backbone.}\label{tab:specialist-memory-lawbench-qwen3-14b}\\
\toprule
\textbf{Task} & \textbf{Base} & \textbf{CPT} & \textbf{LoRA} & \textbf{RAG} & \textbf{+Mem 0.6B} & \textbf{+Mem 1.7B} \\
\midrule
\endfirsthead
\multicolumn{7}{@{}l}{\textbf{\hyperref[tab:specialist-memory-lawbench-qwen3-14b]{\textcolor{darkblue}{Table~\ref*{tab:specialist-memory-lawbench-qwen3-14b}}} continued}}\\
\toprule
\textbf{Task} & \textbf{Base} & \textbf{CPT} & \textbf{LoRA} & \textbf{RAG} & \textbf{+Mem 0.6B} & \textbf{+Mem 1.7B} \\
\midrule
\endhead
\bottomrule
\endfoot
1.1 Article recitation & 31.50 & -- & -- & 30.95 & \spdomainmem{30.29} & \spdomainmembest{31.59} \\
1.2 Knowledge question answering & 17.20 & -- & -- & \spdomainbest{97.40} & \spdomainmem{51.40} & \spdomainmem{70.20} \\
2.1 Document proofreading & 8.76 & -- & -- & \spdomainbest{9.57} & \spdomainmem{8.52} & \spdomainmem{8.52} \\
2.2 Dispute focus identification & \spdomainbest{29.40} & -- & -- & 25.60 & \spdomainmembest{29.40} & \spdomainmembest{29.40} \\
2.3 Marital disputes identification & 21.00 & -- & -- & \spdomainbest{69.57} & \spdomainmem{42.56} & \spdomainmem{39.77} \\
2.4 Issue topic identification & 33.20 & -- & -- & \spdomainbest{36.60} & \spdomainmem{33.20} & \spdomainmem{33.20} \\
2.5 Reading comprehension & 3.79 & -- & -- & 12.94 & \spdomainmem{26.91} & \spdomainmembest{28.93} \\
2.6 Named entity recognition & 35.51 & -- & -- & 23.73 & \spdomainmembest{37.86} & \spdomainmem{37.08} \\
\end{longtable}
\endgroup

\Needspace{0.42\textheight}
\begingroup
\footnotesize
\setlength{\tabcolsep}{2.6pt}
\renewcommand{\arraystretch}{1.04}
\setlength{\LTpre}{3pt}
\setlength{\LTpost}{6pt}
\begin{longtable}{@{}>{\raggedright\arraybackslash}p{0.445\linewidth}rrrrrr@{}}
\caption{Per-task domain memory results on LawBench for the OLMo-2-7B backbone.}\label{tab:specialist-memory-lawbench-olmo2-7b}\\
\toprule
\textbf{Task} & \textbf{Base} & \textbf{CPT} & \textbf{LoRA} & \textbf{RAG} & \textbf{+Mem 0.6B} & \textbf{+Mem 1.7B} \\
\midrule
\endfirsthead
\multicolumn{7}{@{}l}{\textbf{\hyperref[tab:specialist-memory-lawbench-olmo2-7b]{\textcolor{darkblue}{Table~\ref*{tab:specialist-memory-lawbench-olmo2-7b}}} continued}}\\
\toprule
\textbf{Task} & \textbf{Base} & \textbf{CPT} & \textbf{LoRA} & \textbf{RAG} & \textbf{+Mem 0.6B} & \textbf{+Mem 1.7B} \\
\midrule
\endhead
\bottomrule
\endfoot
1.1 Article recitation & 0.00 & \spdomainbest{15.80} & 4.15 & 0.38 & \spdomainmem{3.98} & \spdomainmem{7.10} \\
1.2 Knowledge question answering & 0.00 & 7.20 & 6.00 & 10.20 & \spdomainmembest{20.20} & \spdomainmembest{20.20} \\
2.1 Document proofreading & 0.56 & \spdomainbest{1.34} & 1.05 & 0.94 & \spdomainmem{0.90} & \spdomainmem{1.05} \\
2.2 Dispute focus identification & 3.60 & 3.60 & \spdomainbest{4.20} & 2.80 & \spdomainmembest{4.20} & \spdomainmem{2.80} \\
2.3 Marital disputes identification & 7.12 & \spdomainbest{40.12} & 8.13 & 20.61 & \spdomainmem{2.59} & \spdomainmem{2.40} \\
2.4 Issue topic identification & 6.80 & \spdomainbest{27.60} & 10.60 & 19.60 & \spdomainmem{7.60} & \spdomainmem{7.60} \\
2.5 Reading comprehension & 2.03 & \spdomainbest{27.11} & 26.85 & 2.95 & \spdomainmem{16.46} & \spdomainmem{18.52} \\
2.6 Named entity recognition & 2.36 & 2.00 & \spdomainbest{3.25} & 2.34 & \spdomainmem{2.42} & \spdomainmem{2.46} \\
\end{longtable}
\endgroup

\Needspace{0.42\textheight}
\begingroup
\footnotesize
\setlength{\tabcolsep}{2.6pt}
\renewcommand{\arraystretch}{1.04}
\setlength{\LTpre}{3pt}
\setlength{\LTpost}{6pt}
\begin{longtable}{@{}>{\raggedright\arraybackslash}p{0.445\linewidth}rrrrrr@{}}
\caption{Per-task domain memory results on LawBench for the OLMo-3-7B backbone.}\label{tab:specialist-memory-lawbench-olmo3-7b}\\
\toprule
\textbf{Task} & \textbf{Base} & \textbf{CPT} & \textbf{LoRA} & \textbf{RAG} & \textbf{+Mem 0.6B} & \textbf{+Mem 1.7B} \\
\midrule
\endfirsthead
\multicolumn{7}{@{}l}{\textbf{\hyperref[tab:specialist-memory-lawbench-olmo3-7b]{\textcolor{darkblue}{Table~\ref*{tab:specialist-memory-lawbench-olmo3-7b}}} continued}}\\
\toprule
\textbf{Task} & \textbf{Base} & \textbf{CPT} & \textbf{LoRA} & \textbf{RAG} & \textbf{+Mem 0.6B} & \textbf{+Mem 1.7B} \\
\midrule
\endhead
\bottomrule
\endfoot
1.1 Article recitation & 3.52 & \spdomainbest{18.26} & 5.88 & 6.43 & \spdomainmem{8.68} & \spdomainmem{11.52} \\
1.2 Knowledge question answering & 9.40 & 14.60 & 6.40 & \spdomainbest{34.20} & \spdomainmem{28.20} & \spdomainmem{27.80} \\
2.1 Document proofreading & 2.07 & \spdomainbest{2.69} & 2.04 & 1.71 & \spdomainmem{2.07} & \spdomainmem{2.07} \\
2.2 Dispute focus identification & 7.80 & 4.80 & 5.00 & 5.00 & \spdomainmembest{8.20} & \spdomainmem{7.80} \\
2.3 Marital disputes identification & 12.26 & 20.55 & 6.73 & \spdomainbest{54.61} & \spdomainmem{3.25} & \spdomainmem{12.26} \\
2.4 Issue topic identification & 17.20 & \spdomainbest{34.60} & 0.60 & 27.00 & \spdomainmem{17.20} & \spdomainmem{17.20} \\
2.5 Reading comprehension & 2.32 & 28.19 & 4.54 & 3.07 & \spdomainmembest{42.72} & \spdomainmembest{42.72} \\
2.6 Named entity recognition & \spdomainbest{16.08} & 3.43 & 2.01 & 7.48 & \spdomainmem{16.02} & \spdomainmem{16.02} \\
\end{longtable}
\endgroup

\clearpage

\subsection{FinEval}

All FinEval subtasks use \texttt{generate\_until}. The metric column lists the
primary lm-evaluation-harness metric for each task after multiplying by 100.
EM denotes exact match after answer normalization, MCC denotes Matthews
correlation coefficient over extracted labels, and ROUGE-1 denotes the unigram
F-measure for generated text. AVG is the macro average over the 25 task scores.

\Needspace{0.42\textheight}
\begingroup
\footnotesize
\setlength{\tabcolsep}{2.6pt}
\renewcommand{\arraystretch}{1.04}
\setlength{\LTpre}{3pt}
\setlength{\LTpost}{6pt}
\begin{longtable}{@{}>{\raggedright\arraybackslash}p{0.365\linewidth}>{\centering\arraybackslash}p{0.085\linewidth}rrrrrr@{}}
\caption{Per-task domain memory results on FinEval for the Qwen3-0.6B-Base backbone.}\label{tab:specialist-memory-fineval-qwen3-0p6b}\\
\toprule
\textbf{Task} & \textbf{Metric} & \textbf{Base} & \textbf{CPT} & \textbf{LoRA} & \textbf{RAG} & \textbf{+Mem 0.6B} & \textbf{+Mem 1.7B} \\
\midrule
\endfirsthead
\multicolumn{8}{@{}l}{\textbf{\hyperref[tab:specialist-memory-fineval-qwen3-0p6b]{\textcolor{darkblue}{Table~\ref*{tab:specialist-memory-fineval-qwen3-0p6b}}} continued}}\\
\toprule
\textbf{Task} & \textbf{Metric} & \textbf{Base} & \textbf{CPT} & \textbf{LoRA} & \textbf{RAG} & \textbf{+Mem 0.6B} & \textbf{+Mem 1.7B} \\
\midrule
\endhead
\bottomrule
\endfoot
CFA-Challenge & EM & 0.00 & 4.44 & 5.56 & 1.11 & \spdomainmem{11.11} & \spdomainmembest{25.56} \\
CFA-Easy & EM & 22.67 & 32.27 & 21.12 & 25.10 & \spdomainmembest{39.34} & \spdomainmem{29.84} \\
CRA-Bigdata & EM & 30.57 & \spdomainbest{54.96} & 43.21 & 38.25 & \spdomainmem{30.84} & \spdomainmem{54.14} \\
CRA-CCF & MCC & 0.00 & 2.61 & 0.00 & -1.37 & \spdomainmem{0.58} & \spdomainmembest{4.37} \\
CRA-CCFraud & EM & 6.01 & 0.00 & 12.35 & \spdomainbest{38.04} & \spdomainmem{9.01} & \spdomainmem{7.05} \\
CRA-LendingClub & EM & 78.93 & 16.16 & 59.35 & 23.86 & \spdomainmembest{79.52} & \spdomainmem{79.23} \\
CRA-Polish & MCC & 0.36 & 0.00 & \spdomainbest{1.12} & 0.00 & \spdomainmem{0.54} & \spdomainmem{0.88} \\
CRA-ProtoSeguro & EM & 3.02 & 3.02 & \spdomainbest{63.10} & 11.38 & \spdomainmem{3.02} & \spdomainmem{4.28} \\
\end{longtable}
\endgroup

\Needspace{0.42\textheight}
\begingroup
\footnotesize
\setlength{\tabcolsep}{2.6pt}
\renewcommand{\arraystretch}{1.04}
\setlength{\LTpre}{3pt}
\setlength{\LTpost}{6pt}
\begin{longtable}{@{}>{\raggedright\arraybackslash}p{0.365\linewidth}>{\centering\arraybackslash}p{0.085\linewidth}rrrrrr@{}}
\caption{Per-task domain memory results on FinEval for the Qwen3-1.7B-Base backbone.}\label{tab:specialist-memory-fineval-qwen3-1p7b}\\
\toprule
\textbf{Task} & \textbf{Metric} & \textbf{Base} & \textbf{CPT} & \textbf{LoRA} & \textbf{RAG} & \textbf{+Mem 0.6B} & \textbf{+Mem 1.7B} \\
\midrule
\endfirsthead
\multicolumn{8}{@{}l}{\textbf{\hyperref[tab:specialist-memory-fineval-qwen3-1p7b]{\textcolor{darkblue}{Table~\ref*{tab:specialist-memory-fineval-qwen3-1p7b}}} continued}}\\
\toprule
\textbf{Task} & \textbf{Metric} & \textbf{Base} & \textbf{CPT} & \textbf{LoRA} & \textbf{RAG} & \textbf{+Mem 0.6B} & \textbf{+Mem 1.7B} \\
\midrule
\endhead
\bottomrule
\endfoot
CFA-Challenge & EM & 0.00 & 4.44 & 11.11 & 1.11 & \spdomainmem{12.22} & \spdomainmembest{24.44} \\
CFA-Easy & EM & 3.29 & 47.58 & \spdomainbest{55.52} & 34.21 & \spdomainmem{40.41} & \spdomainmem{25.10} \\
CRA-Bigdata & EM & 30.64 & 43.48 & 46.60 & 40.69 & \spdomainmem{31.86} & \spdomainmembest{52.99} \\
CRA-CCF & MCC & -0.60 & -1.59 & 0.00 & -1.05 & \spdomainmembest{0.87} & \spdomainmem{0.04} \\
CRA-CCFraud & EM & 55.00 & 5.86 & 77.31 & \spdomainbest{90.37} & \spdomainmem{56.20} & \spdomainmem{56.20} \\
CRA-LendingClub & EM & 39.20 & 75.96 & \spdomainbest{79.19} & 75.66 & \spdomainmem{52.21} & \spdomainmem{54.14} \\
CRA-Polish & MCC & -1.37 & 0.79 & 0.00 & -3.71 & \spdomainmem{-1.37} & \spdomainmembest{0.92} \\
CRA-ProtoSeguro & EM & 28.59 & 37.57 & \spdomainbest{96.98} & 80.27 & \spdomainmem{31.11} & \spdomainmem{31.11} \\
\end{longtable}
\endgroup

\Needspace{0.42\textheight}
\begingroup
\footnotesize
\setlength{\tabcolsep}{2.6pt}
\renewcommand{\arraystretch}{1.04}
\setlength{\LTpre}{3pt}
\setlength{\LTpost}{6pt}
\begin{longtable}{@{}>{\raggedright\arraybackslash}p{0.365\linewidth}>{\centering\arraybackslash}p{0.085\linewidth}rrrrrr@{}}
\caption{Per-task domain memory results on FinEval for the Qwen3-4B-Base backbone.}\label{tab:specialist-memory-fineval-qwen3-4b}\\
\toprule
\textbf{Task} & \textbf{Metric} & \textbf{Base} & \textbf{CPT} & \textbf{LoRA} & \textbf{RAG} & \textbf{+Mem 0.6B} & \textbf{+Mem 1.7B} \\
\midrule
\endfirsthead
\multicolumn{8}{@{}l}{\textbf{\hyperref[tab:specialist-memory-fineval-qwen3-4b]{\textcolor{darkblue}{Table~\ref*{tab:specialist-memory-fineval-qwen3-4b}}} continued}}\\
\toprule
\textbf{Task} & \textbf{Metric} & \textbf{Base} & \textbf{CPT} & \textbf{LoRA} & \textbf{RAG} & \textbf{+Mem 0.6B} & \textbf{+Mem 1.7B} \\
\midrule
\endhead
\bottomrule
\endfoot
CFA-Challenge & EM & 0.00 & \spdomainbest{25.56} & 24.44 & 11.11 & \spdomainmem{12.22} & \spdomainmem{24.44} \\
CFA-Easy & EM & 49.03 & 62.98 & \spdomainbest{67.34} & 49.61 & \spdomainmem{63.08} & \spdomainmem{54.17} \\
CRA-Bigdata & EM & 54.01 & 40.42 & \spdomainbest{55.10} & 52.65 & \spdomainmem{54.55} & \spdomainmembest{55.10} \\
CRA-CCF & MCC & 0.00 & 1.38 & 0.00 & 0.08 & \spdomainmem{0.58} & \spdomainmembest{7.34} \\
CRA-CCFraud & EM & \spdomainbest{9.82} & 5.86 & 5.82 & 6.58 & \spdomainmem{8.58} & \spdomainmem{8.20} \\
CRA-LendingClub & EM & \spdomainbest{20.77} & 19.29 & 0.00 & \spdomainbest{20.77} & \spdomainmem{20.59} & \spdomainmem{20.62} \\
CRA-Polish & MCC & 15.25 & -1.10 & 12.98 & -2.24 & \spdomainmembest{15.56} & \spdomainmembest{15.56} \\
CRA-ProtoSeguro & EM & 86.73 & 1.51 & 0.00 & \spdomainbest{96.98} & \spdomainmem{84.47} & \spdomainmem{84.47} \\
\end{longtable}
\endgroup

\Needspace{0.42\textheight}
\begingroup
\footnotesize
\setlength{\tabcolsep}{2.6pt}
\renewcommand{\arraystretch}{1.04}
\setlength{\LTpre}{3pt}
\setlength{\LTpost}{6pt}
\begin{longtable}{@{}>{\raggedright\arraybackslash}p{0.365\linewidth}>{\centering\arraybackslash}p{0.085\linewidth}rrrrrr@{}}
\caption{Per-task domain memory results on FinEval for the Qwen3-8B-Base backbone.}\label{tab:specialist-memory-fineval-qwen3-8b}\\
\toprule
\textbf{Task} & \textbf{Metric} & \textbf{Base} & \textbf{CPT} & \textbf{LoRA} & \textbf{RAG} & \textbf{+Mem 0.6B} & \textbf{+Mem 1.7B} \\
\midrule
\endfirsthead
\multicolumn{8}{@{}l}{\textbf{\hyperref[tab:specialist-memory-fineval-qwen3-8b]{\textcolor{darkblue}{Table~\ref*{tab:specialist-memory-fineval-qwen3-8b}}} continued}}\\
\toprule
\textbf{Task} & \textbf{Metric} & \textbf{Base} & \textbf{CPT} & \textbf{LoRA} & \textbf{RAG} & \textbf{+Mem 0.6B} & \textbf{+Mem 1.7B} \\
\midrule
\endhead
\bottomrule
\endfoot
CFA-Challenge & EM & 4.44 & 8.89 & 0.00 & 3.33 & \spdomainmem{11.11} & \spdomainmembest{27.78} \\
CFA-Easy & EM & 36.63 & 47.00 & 37.79 & 31.30 & \spdomainmembest{62.02} & \spdomainmem{48.26} \\
CRA-Bigdata & EM & 52.65 & 54.14 & 53.94 & 50.54 & \spdomainmem{53.94} & \spdomainmembest{54.82} \\
CRA-CCF & MCC & 1.18 & 1.33 & \spdomainbest{6.66} & -0.20 & \spdomainmem{4.07} & \spdomainmem{4.07} \\
CRA-CCFraud & EM & 37.94 & 5.86 & 5.86 & 17.11 & \spdomainmem{36.51} & \spdomainmembest{38.08} \\
CRA-LendingClub & EM & 62.54 & 42.55 & \spdomainbest{67.19} & 58.60 & \spdomainmem{62.02} & \spdomainmem{62.02} \\
CRA-Polish & MCC & 13.64 & 2.34 & 11.17 & 11.42 & \spdomainmem{14.49} & \spdomainmembest{14.61} \\
CRA-ProtoSeguro & EM & \spdomainbest{96.98} & \spdomainbest{96.98} & \spdomainbest{96.98} & \spdomainbest{96.98} & \spdomainmembest{96.98} & \spdomainmembest{96.98} \\
\end{longtable}
\endgroup

\Needspace{0.42\textheight}
\begingroup
\footnotesize
\setlength{\tabcolsep}{2.6pt}
\renewcommand{\arraystretch}{1.04}
\setlength{\LTpre}{3pt}
\setlength{\LTpost}{6pt}
\begin{longtable}{@{}>{\raggedright\arraybackslash}p{0.365\linewidth}>{\centering\arraybackslash}p{0.085\linewidth}rrrrrr@{}}
\caption{Per-task domain memory results on FinEval for the Qwen3-14B-Base backbone.}\label{tab:specialist-memory-fineval-qwen3-14b}\\
\toprule
\textbf{Task} & \textbf{Metric} & \textbf{Base} & \textbf{CPT} & \textbf{LoRA} & \textbf{RAG} & \textbf{+Mem 0.6B} & \textbf{+Mem 1.7B} \\
\midrule
\endfirsthead
\multicolumn{8}{@{}l}{\textbf{\hyperref[tab:specialist-memory-fineval-qwen3-14b]{\textcolor{darkblue}{Table~\ref*{tab:specialist-memory-fineval-qwen3-14b}}} continued}}\\
\toprule
\textbf{Task} & \textbf{Metric} & \textbf{Base} & \textbf{CPT} & \textbf{LoRA} & \textbf{RAG} & \textbf{+Mem 0.6B} & \textbf{+Mem 1.7B} \\
\midrule
\endhead
\bottomrule
\endfoot
CFA-Challenge & EM & 3.33 & -- & -- & 8.89 & \spdomainmem{14.44} & \spdomainmembest{28.89} \\
CFA-Easy & EM & 8.04 & -- & -- & 18.80 & \spdomainmembest{50.29} & \spdomainmem{29.36} \\
CRA-Bigdata & EM & 48.10 & -- & -- & 44.23 & \spdomainmem{49.25} & \spdomainmembest{53.53} \\
CRA-CCF & MCC & 5.15 & -- & -- & -0.24 & \spdomainmem{5.09} & \spdomainmembest{6.32} \\
CRA-CCFraud & EM & 48.95 & -- & -- & 39.28 & \spdomainmembest{49.00} & \spdomainmembest{49.00} \\
CRA-LendingClub & EM & 62.17 & -- & -- & 49.39 & \spdomainmem{63.17} & \spdomainmembest{63.51} \\
CRA-Polish & MCC & \spdomainbest{13.32} & -- & -- & 11.87 & \spdomainmem{12.76} & \spdomainmem{12.77} \\
CRA-ProtoSeguro & EM & \spdomainbest{96.98} & -- & -- & \spdomainbest{96.98} & \spdomainmembest{96.98} & \spdomainmembest{96.98} \\
\end{longtable}
\endgroup

\Needspace{0.42\textheight}
\begingroup
\footnotesize
\setlength{\tabcolsep}{2.6pt}
\renewcommand{\arraystretch}{1.04}
\setlength{\LTpre}{3pt}
\setlength{\LTpost}{6pt}
\begin{longtable}{@{}>{\raggedright\arraybackslash}p{0.365\linewidth}>{\centering\arraybackslash}p{0.085\linewidth}rrrrrr@{}}
\caption{Per-task domain memory results on FinEval for the OLMo-2-7B backbone.}\label{tab:specialist-memory-fineval-olmo2-7b}\\
\toprule
\textbf{Task} & \textbf{Metric} & \textbf{Base} & \textbf{CPT} & \textbf{LoRA} & \textbf{RAG} & \textbf{+Mem 0.6B} & \textbf{+Mem 1.7B} \\
\midrule
\endfirsthead
\multicolumn{8}{@{}l}{\textbf{\hyperref[tab:specialist-memory-fineval-olmo2-7b]{\textcolor{darkblue}{Table~\ref*{tab:specialist-memory-fineval-olmo2-7b}}} continued}}\\
\toprule
\textbf{Task} & \textbf{Metric} & \textbf{Base} & \textbf{CPT} & \textbf{LoRA} & \textbf{RAG} & \textbf{+Mem 0.6B} & \textbf{+Mem 1.7B} \\
\midrule
\endhead
\bottomrule
\endfoot
CFA-Challenge & EM & 34.44 & 32.22 & 27.78 & 24.44 & \spdomainmem{34.44} & \spdomainmembest{35.56} \\
CFA-Easy & EM & 53.49 & 48.45 & 51.55 & \spdomainbest{56.69} & \spdomainmem{53.97} & \spdomainmem{54.17} \\
CRA-Bigdata & EM & 46.94 & 47.62 & 44.63 & 45.31 & \spdomainmembest{48.98} & \spdomainmem{47.55} \\
CRA-CCF & MCC & 0.00 & 0.11 & 0.13 & 0.20 & \spdomainmembest{2.28} & \spdomainmem{0.82} \\
CRA-CCFraud & EM & 83.41 & \spdomainbest{94.14} & 86.37 & 32.13 & \spdomainmem{84.22} & \spdomainmem{84.22} \\
CRA-LendingClub & EM & 19.77 & 52.62 & 19.47 & \spdomainbest{65.66} & \spdomainmem{28.61} & \spdomainmem{19.73} \\
CRA-Polish & MCC & \spdomainbest{9.73} & 0.75 & 0.00 & 0.00 & \spdomainmem{5.65} & \spdomainmem{5.36} \\
CRA-ProtoSeguro & EM & \spdomainbest{96.98} & \spdomainbest{96.98} & \spdomainbest{96.98} & \spdomainbest{96.98} & \spdomainmembest{96.98} & \spdomainmembest{96.98} \\
\end{longtable}
\endgroup

\Needspace{0.42\textheight}
\begingroup
\footnotesize
\setlength{\tabcolsep}{2.6pt}
\renewcommand{\arraystretch}{1.04}
\setlength{\LTpre}{3pt}
\setlength{\LTpost}{6pt}
\begin{longtable}{@{}>{\raggedright\arraybackslash}p{0.365\linewidth}>{\centering\arraybackslash}p{0.085\linewidth}rrrrrr@{}}
\caption{Per-task domain memory results on FinEval for the OLMo-3-7B backbone.}\label{tab:specialist-memory-fineval-olmo3-7b}\\
\toprule
\textbf{Task} & \textbf{Metric} & \textbf{Base} & \textbf{CPT} & \textbf{LoRA} & \textbf{RAG} & \textbf{+Mem 0.6B} & \textbf{+Mem 1.7B} \\
\midrule
\endfirsthead
\multicolumn{8}{@{}l}{\textbf{\hyperref[tab:specialist-memory-fineval-olmo3-7b]{\textcolor{darkblue}{Table~\ref*{tab:specialist-memory-fineval-olmo3-7b}}} continued}}\\
\toprule
\textbf{Task} & \textbf{Metric} & \textbf{Base} & \textbf{CPT} & \textbf{LoRA} & \textbf{RAG} & \textbf{+Mem 0.6B} & \textbf{+Mem 1.7B} \\
\midrule
\endhead
\bottomrule
\endfoot
CFA-Challenge & EM & 23.33 & \spdomainbest{26.67} & 20.00 & 17.78 & \spdomainmem{25.56} & \spdomainmem{25.56} \\
CFA-Easy & EM & 56.30 & 49.90 & 52.42 & 52.71 & \spdomainmembest{58.33} & \spdomainmem{56.78} \\
CRA-Bigdata & EM & 44.57 & 44.77 & 41.85 & 49.25 & \spdomainmembest{49.46} & \spdomainmem{47.89} \\
CRA-CCF & MCC & 0.00 & 0.00 & -0.45 & 0.00 & \spdomainmembest{11.22} & \spdomainmem{2.18} \\
CRA-CCFraud & EM & 10.77 & 14.68 & 10.53 & 23.02 & \spdomainmem{11.11} & \spdomainmembest{77.93} \\
CRA-LendingClub & EM & 45.00 & \spdomainbest{80.75} & 13.79 & 26.20 & \spdomainmem{44.93} & \spdomainmem{44.93} \\
CRA-Polish & MCC & 0.00 & -0.97 & 1.29 & \spdomainbest{3.79} & \spdomainmem{3.50} & \spdomainmem{1.50} \\
CRA-ProtoSeguro & EM & \spdomainbest{96.98} & 3.32 & 3.48 & 11.96 & \spdomainmembest{96.98} & \spdomainmembest{96.98} \\
\end{longtable}
\endgroup

\end{document}